\documentclass{article}

\usepackage{PRIMEarxiv}

\usepackage[utf8]{inputenc} 
\usepackage[T1]{fontenc}    
\usepackage{hyperref}       
\usepackage{url}            
\usepackage{booktabs}       
\usepackage{amsfonts}       
\usepackage{nicefrac}       
\usepackage{microtype}      
\usepackage{lipsum}
\usepackage{fancyhdr}       
\usepackage{graphicx}       
\graphicspath{{media/}}     
\usepackage{titlesec}
\usepackage{color}
\usepackage{graphicx}
\usepackage{subcaption}
\usepackage{booktabs}
\usepackage[utf8]{inputenc}
\usepackage[english]{babel}
\usepackage[T1]{fontenc} 
\usepackage{amsfonts} 

\usepackage{graphicx}
\graphicspath{{Images/}}
\usepackage{eso-pic} 
\usepackage{caption} 
\usepackage{transparent}

\usepackage{amsmath}
\usepackage{amsthm}
\usepackage{bm}
\usepackage[overload]{empheq}  

\usepackage{tabularx}
\usepackage{longtable} 
\usepackage{colortbl}

\usepackage{algorithm}
\usepackage{algpseudocode}

\usepackage{appendix}

\usepackage{enumitem}

\usepackage{amsthm,thmtools,xcolor} 
\usepackage{comment} 
\usepackage{fancyhdr} 
\usepackage{lipsum} 
\usepackage{tcolorbox} 
\usepackage{soul,xcolor}

\newcommand{\bea}{\begin{eqnarray}} 
\newcommand{\eea}{\end{eqnarray}}

\title{HypEMBER: Hypernetwork-based Ensemble for Robust Policy Learning of Parametrized Dynamical Systems}

\author{
  Nicolò Botteghi \\
  MOX, Department of Mathematics \\
  Politecnico di Milano \\
  Milano, Italy\\
  \texttt{nicolo.botteghi@polimi.it} \\
   \And
  Gabriele Pascali \\
  MOX, Department of Mathematics \\
  Politecnico di Milano \\
  Milano, Italy\\
   \AND
  Urban Fasel \\
  Department of Aeronautics \\
  Imperial College London \\
  London, United Kingdom \\
  \And
  Andrea Manzoni \\
  MOX, Department of Mathematics \\
  Politecnico di Milano \\
  Milano, Italy\\
}

\begin{document}
\maketitle

\begin{abstract}
In this work we investigate reinforcement learning (RL) as a framework for the robust control of parametrized dynamical systems in the presence of measurements and model uncertainties. High-dimensional state spaces, expensive numerical solvers, the partial knowledge of the governing equations, and the dependence on physical parameters that may be uncertain or difficult to estimate accurately, make the use of standard RL approaches computationally infeasible. Indeed, lack of robustness and poor generalization across parameter variations are further amplified in the presence of noisy or incomplete measurements, ultimately hampering control performance.\\ To address these challenges, we introduce \emph{HypEMBER}, a novel RL framework based on the combination of \emph{hypernetworks} and \emph{ensemble learning}. In the proposed approach, both the policy and value functions are represented through hypernetworks that generate the weights of the underlying models conditioned on the physical parameters of the system, thereby enabling parametric generalization across different dynamical regimes. In addition, an ensemble of policy and value approximators is employed to quantify epistemic uncertainty, leading to improved exploration strategies and enhanced robustness during and after training. \\ The performance of the proposed framework is assessed on two representative parametrized control problems: \emph{(i)} the one-dimensional Kuramoto–Sivashinsky equation and \emph{(ii)} a particle-navigation task in a two-dimensional time-dependent gyre flow, focusing on robustness with respect to measurement noise and parameter misspecification. Numerical results demonstrate that HypEMBER consistently improves training stability and sample efficiency, while achieving superior robustness to uncertainties affecting both the system dynamics and the available observations, in comparison with state-of-the-art RL methods.
\end{abstract}


\section{Introduction}
\label{sec:introduction}
Large-scale distributed dynamical systems arising from a wealth of applications, such as, e.g., mechanical processes, robotics, and autonomous vehicles,  are described in terms of high-dimensional systems of differential equations depending on a set of parameters to include a range of different operating conditions or physical scenarios. Controlling these systems requires the design of a control policy that drives the system toward the desired behavior for every scenario. This class of problems is traditionally addressed within the framework of optimal control, where a policy is obtained by minimizing a task-dependent cost functional subject to the differential equations governing the system dynamics \cite{manzoni2021optimal,kirk2004optimal,fleming2006controlled}. Optimal-control approaches rely on accurate mathematical models and repeated numerical simulations -- both forward and backward in time -- to compute optimal control laws. While well established, these methods become increasingly challenging when the parameters vary across a large set of possible values, when the governing physics are only partially known -- such as in many real-world problems -- and, above all, when the numerical solution of the underlying equations is computationally expensive. In such cases, the associated optimization problems quickly become intractable.

These limitations have motivated the investigation and the use of reinforcement learning (RL) as a general framework for control \cite{sutton2018reinforcement}. In RL, an agent interacts with an unknown environment, i.e., the dynamical system, and learns to select control actions that maximize a cumulative performance objective. From a control-theoretic perspective, RL can be interpreted as a data-driven realization of dynamic programming, where optimal policies are learned through interaction rather than by explicitly solving the underlying optimal control problem \cite{bertsekas1996neurodynamic,powell2011approximate}. More recently,  the integration of deep neural networks has led to deep reinforcement learning, enabling the representation of complex policies and value functions and allowing RL-based controllers to scale to high-dimensional problems \cite{arulkumaran2017deep,francois2018introduction}. As a result, (deep) RL has demonstrated remarkable success across a wide range of control applications, including games  \cite{mnih2015human, ye2020mastering, mnih2013playing, van2016deep}, simulated and real-world robotics \cite{lin1992reinforcement, kober2013reinforcement, zhang2015towards, gu2017deep, zhao2020sim, botteghi2021low}, and, more recently, systems governed by partial differential equations (PDEs), such as fluid dynamics and active flow control \cite{bucci2019control, fan2020reinforcement, rabault2019accelerating, xia2023active, peitz2023distributed, zolman2024sindy, botteghi2024parametric, Botteghi2025HypeMARL:Systems}. In most practical scenarios, RL agents are trained in simulated environments, where repeated interactions with numerical models are feasible and safe. However, despite the successes, the learned policies must ultimately operate in environments that differ -- sometimes significantly -- from the training conditions. A growing body of work has highlighted the sensitivity of RL algorithms to measurement noise, modeling inaccuracies, like misspecified physics, and numerical errors \cite{engstrom2020implementation, rajeswaran2017epopt, ChallengesRealWorldRL, agarwal2021deep, packer2018assessing}. These drawbacks are especially relevant in scientific computing, where the governing equations are usually approximated numerically and depend on uncertain physical parameters, further widening the simulation-to-reality gap \cite{salvato2021crossing, zhao2020sim, kober2013reinforcement, dulacarnold2021}.

Addressing these challenges requires learning frameworks that explicitly account for uncertainty and variability during both training and decision making. A key aspect in this context is the distinction between aleatoric and epistemic uncertainty \cite{hullermeier2021aleatoric}. Aleatoric uncertainty captures inherent randomness in the system and cannot be reduced through additional data, whereas epistemic uncertainty reflects lack of knowledge about the system and can, in principle, be reduced with more information. Standard RL methods typically do not disentangle these two sources, leading to suboptimal exploration strategies and reduced robustness.
Recent works on uncertainty-aware RL aim to mitigate these issues by incorporating uncertainty estimates into the learning process, improving both robustness and data efficiency \cite{MASURE, Masksembles, IVRL}.

In this work, we consider RL to solve generic optimal control problems of nonlinear, time-dependent, parametrized dynamical systems, potentially arising from the space–time discretization of PDEs or directly from a (large-scale) system of ordinary differential equations. The discrete-time system evolution is described by a relationship of the form
\begin{equation}
    \bm{s}_{t+1} = F(\bm{s}_t, \bm{a}_t;\boldsymbol{\mu}), \quad t=0, \ldots, N_t\, ,
    \label{eq:dynamics}
\end{equation}
where $\bm{s}_t \in \mathbb{R}^{N_s}$ denotes the system state at time $t$, $\bm{a}_t \in \mathbb{R}^{N_a}$ the control input, $\boldsymbol{\mu} \in \mathbb{R}^{N_\mu}$ a vector of physical, task-dependent, or environmental parameters, affecting, e.g., model coefficients, initial or boundary data, as well as the target of the control problem. $N_t$ denotes instead the number of discrete time steps of the control horizon. In the context of parametrized systems, variations in $\boldsymbol{\mu}$ induce potentially significant changes in the system behavior, posing additional challenges for control design and generalization. The mapping $F:\mathbb{R}^{N_s}\times\mathbb{R}^{N_a}\times\mathbb{R}^{N_{\mu}}\to\mathbb{R}^{N_s}$ represents the system dynamics induced by a suitable discretization scheme. The objective is to determine a feedback control law $\bm{a}_t=\pi(\bm{s}_t,\boldsymbol{\mu};\boldsymbol{\phi})$ -- parametrized by a neural network of learnable parameters $\boldsymbol{\phi}$ -- that maximizes a task-dependent reward function $R:\mathbb{R}^{N_s} \times \mathbb{R}^{N_a} \times \mathbb{R}^{N_{\mu}} \rightarrow \mathbb{R}$ of the type:
\begin{equation}
r_t =R(\bm{s}_t,\bm{a}_t;\boldsymbol{\mu})
=
-\frac{1}{2}\underbrace{\,\|\bm{s}_t-\bm{s}_{\mathrm{ref}}(\boldsymbol{\mu})\|_2^2}_{\text{state cost}}
-\frac{\beta}{2}\underbrace{\,\|\bm{a}_t(\boldsymbol{\mu})\|_2^2}_{\text{action cost}}\, , \quad t=0, \ldots, N_t\, ,
\label{eq:reward_function}
\end{equation}
where $\bm{s}_{\text{ref}}(\boldsymbol{\mu})$ denotes the $\boldsymbol{\mu}$-dependent reference state, and $\beta$ is a coefficient trading-off state and action costs.
\begin{figure}[h!]
    \centering
    \includegraphics[width=1.0\linewidth]{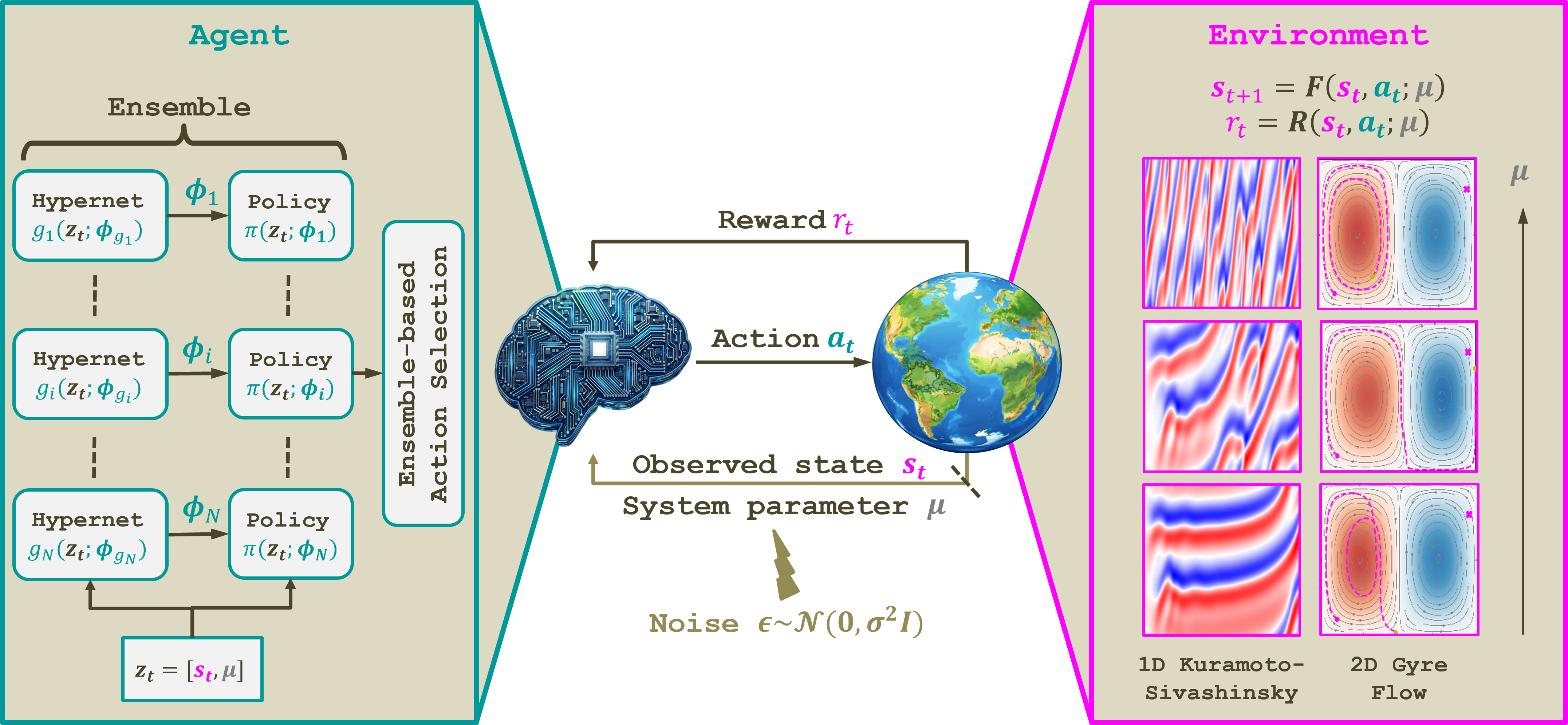}
    \caption{HypEMBER takes advantage of an ensemble of $N$ hypernetworks $\{g_i(\bm{z}_t;\boldsymbol{\phi}_{g_i})\}_{i=1}^N$ and $N$ policies $\{\pi(\bm{z}_t;\boldsymbol{\phi}_i)\}_{i=1}^N$ to learn robust control strategies for parametrized dynamical systems under perturbations $\boldsymbol{\epsilon} \sim \mathcal{N}(\bm{0}, \sigma^2 I)$ of the observed state $\bm{s}_t$ and system parameter $\boldsymbol{\mu}$.}
    \label{fig:1}
\end{figure}
In this context, we introduce a novel RL algorithm, HypEMBER, that builds upon two main lines of work (see Figure \ref{fig:1}): (i) the SUNRISE framework \cite{lee2021sunrise} and (ii) our hypernetwork-based parametrization introduced in HypeRL \cite{botteghi2025hyperl}. More specifically, 
\begin{itemize}
    \item[(i)] SUNRISE is a well-established ensemble-based RL approach designed to improve robustness and exploration by explicitly leveraging uncertainty estimates obtained from multiple value function and policy approximators. Epistemic uncertainty is estimated from the ensemble of critics via the empirical standard deviation of Q-function predictions and is directly used to re-weight Bellman updates, reducing the impact of highly uncertain transitions. Moreover, an ensemble of actors is used to guide exploration through an upper confidence bound strategy, where actions are selected by considering both their expected value and the uncertainty estimated across the ensemble, allowing the agent to balance exploration and exploitation during training. SUNRISE has been successfully applied to standard RL benchmarks, e.g., robotics tasks or games, with noisy rewards. However, in all these cases, the controlled dynamical systems were low dimensional with no parametric dependencies and the robustness of the learned policies was not tested against measurement and modeling uncertainties.
    \item[(ii)] HypeRL enhances standard RL algorithms with hypernetworks to improve data efficiency and generalization in the context of control of large-scale parametrized dynamical systems. In particular, HypeRL employs hypernetworks to generate the learnable parameters of policy and value networks, enabling an explicit conditioning of the policy and value function weights and biases on system and task-dependent parameters. Hypernetworks \cite{ha2016hypernetworks} are a class of neural network architectures in which one network is used to generate the weights and biases -- i.e., the learnable parameters of another network. From a modeling perspective, hypernetworks provide a flexible way to represent families of functions that depend on contextual information. Compared to standard neural networks with static parameters, hypernetwork-based models enable parameter sharing across different regimes while retaining the ability to specialize locally when needed. For this reason, hypernetworks have been successfully applied in a variety of settings, including meta-learning, continual learning, and parameter-conditioned control problems \cite{chauhan2023brief, oswald2020continual, beck2023hypernetworks, krueger2017bayesian}.  While the hypernetwork-based parametrization of HypeRL has been shown to improve the RL agents' performance in parametrized control problems, the robustness to uncertainties of HypeRL has yet to be assessed.
\end{itemize} 

With reference to Figure \ref{fig:1}, HypEMBER utilizes an ensemble of $N$ hypernetworks $\{g_i(\bm{z}_t;\boldsymbol{\phi}_{g_i})\}_{i=1}^N$ to learn the weights and biases of $N$ policies $\{\pi(\bm{z}_t;\boldsymbol{\phi}_i)\}_{i=1}^N$, where $N$ denotes the ensemble size, $\bm{z}_t = [\bm{s_t}, \boldsymbol{\mu}]$ is the augmented observed state by  the system parameter $\boldsymbol{\mu}$ to account for the parametric dependency and enable generalization, and $\bm{a}_t$ the action of the agent at time step $t$, chosen according to an ensemble-based action selection. HypEMBER can be viewed as a principled integration of a hypernetwork-based parametrization and an ensemble-based uncertainty estimation to learn robust control strategies for parametrized dynamical systems under perturbations $\boldsymbol{\epsilon} \sim \mathcal{N}(\bm{0}, \boldsymbol{\sigma}^2 I)$ of the observed state $\bm{s}_t$ and system parameter $\boldsymbol{\mu}$. \\
The proposed approach is tested on two challenging control problems, namely: (i) a one-dimensional parametrized Kuramoto--Sivashinsky equation, and (ii) a particle-navigation problem in a two-dimensional parametrized gyre flow, where we analyze the robustness of the several state-of-the-art RL algorithms with respect to uncertainties in the system dynamics, such as variations in physical parameters and emulating a potential simulation-to-reality gap. To evaluate robustness, agents are trained in idealized (noise-free and parameter-consistent) simulation environments and subsequently tested under perturbed conditions, including noisy observations and parameter misspecification.

The remainder of the paper is organized as follows: Section \ref{sec:preliminaries} introduces the RL framework and the main building blocks of HypEMBER. Section \ref{sec:methodology} presents the proposed method in detail. Section \ref{sec:numerical_results} reports and discusses the numerical results, and Section \ref{sec:conclusions} concludes the paper.


\section{Preliminaries}\label{sec:preliminaries}

In this section, we introduce the building blocks of HypEMBER, starting from a brief introduction to RL (Section \ref{subsec:Reinforcement_learning}), actor-critic algorithms (Section \ref{subsect:ActorCritic}), and concluding with a detailed description of SUNRISE (Section \ref{subsubsec:SUNRISE}) and HypeRL (Section \ref{subsec:hyperl}). 

\subsection{Reinforcement Learning}
\label{subsec:Reinforcement_learning}

Reinforcement learning (RL) provides a data-driven framework for sequential decision-making, where an agent learns to control a dynamical system through interaction \cite{sutton2018reinforcement}. At each time step $t$, the agent observes the current state $\bm{s}_t \in \mathcal{S} \subset \mathbb{R}^{N_s}$, selects an action $\bm{a}_t \in \mathcal{A} \subset \mathbb{R}^{N_a}$ according to a policy $\pi$, and receives a scalar reward $r_t \in \mathbb{R}$. The system then transitions to a new state $\bm{s}_{t+1}$.  The agent-environment interaction is formalized as a Markov Decision Process (MDP) \cite{puterman1994markov}
\begin{equation*}
\mathcal{M} = (\mathcal{S}, \mathcal{A}, T, R, \gamma),
\end{equation*}
where $T:\mathcal{S} \times \mathcal{S} \times \mathcal{A} \rightarrow [0,1]$ denotes the transition probability, $R:\mathcal{S} \times \mathcal{A}\rightarrow \mathbb{R}$ the reward function, and $\gamma \in (0,1)$ the discount factor. A key assumption of the MDP framework is the Markov property, which assumes that transitions depend on the current state and action only:
\begin{equation*}
T(\bm{s}',\bm{s},\bm{a}) = P(S_{t+1} = \bm{s}' \mid S_t = \bm{s}, A_t=\bm{a})\, ,
\end{equation*} 
where $S_{t+1}=\bm{s}'$, $S_t=\bm{s}$, and $A_t=\bm{a}$ denote the values $\bm{s}', \bm{s}, \bm{a}$ of random variables $S_{t+1}, S_t, A_t$ at time step $t$, and $P$ a generic probability distribution.  Note that in the case of a deterministic transition function, $T$ is equivalent to the function $F$ from Equation \ref{eq:dynamics}.
A policy $\pi(\bm{s})$ characterizes the agent's behavior and, in the most general case, can be defined as a conditional probability distribution over actions given the current state. In particular,
\begin{equation*}
\pi(\bm{a}|\bm{s}) = P(A_t = \bm{a} \mid S_t = \bm{s})\, ,
\end{equation*} 
where $A_t=\bm{a}$ and $S_t=\bm{s}$ denote the values of random variables (in capital letters) at time step $t$.
The performance of a policy is quantified through the return, defined as the discounted cumulative reward
\begin{equation*}
G_t = \sum_{k=0}^{N_t-1} \gamma^{k} r_{t+k}\, ,
\end{equation*}
and the goal of the agent is to learn the policy that maximizes the expected cumulative reward over a control horizon $N_t$.
Value functions play a central role in RL as they quantify the expected performance of a policy $\pi$. The state-value function associated with a policy $\pi$ is defined as the expected return when the agent starts from state $s$ and subsequently follows policy $\pi$,
\begin{equation*}
V^{\pi}(\bm{s}) = \mathbb{E}_{\pi} \left[ G_t \mid S_t = \bm{s} \right]\, ,
\end{equation*}
with $\mathbb{E}_{\pi}$ denotes the expected value under the policy $\pi$. 
Similarly, the action-value function represents the expected return obtained by taking action $a$ in state $s$ and then following policy $\pi$,
\begin{equation*}
Q^{\pi}(\bm{s},\bm{a}) = \mathbb{E}_{\pi}[G_t \mid S_t = \bm{s}, A_t = \bm{a}].
\end{equation*}
After estimating the optimal value function $Q^*(\bm{s}, \bm{a})$, the optimal (greedy) policy $\pi^\star$ can be obtained as:
\begin{equation}
\pi^\star(\bm{s}) \in \arg\max_{\bm{a} \in \mathcal{A}} Q^\star(\bm{s},\bm{a})\, ,
\label{eq:opt_policyQ}
\end{equation}
optimally solving the RL problem.

In contrast to classical optimal control, RL does not require explicit knowledge of $T$ or $R$, rather, it learns from sampled transitions $(\bm{s}_t,\bm{a}_t,r_t,\bm{s}_{t+1})$.  A central challenge is, therefore, induced by the exploration–exploitation trade-off, namely balancing the use of known high-reward actions with the acquisition of new information.


Within RL, it is common to distinguish between different classes of algorithms based on how the control policy is represented, how experience is used for learning, and whether an explicit model of the environment is available. A first important distinction is between \emph{model-based} and \emph{model-free} methods. Model-based approaches explicitly learn or assume a model of the environment dynamics and exploit it for planning or policy optimization \cite{bertsekas2012dynamic,schrittwieser2020mastering, kaiser2019model, m2023model}. In contrast, model-free methods learn policies or value functions directly from interaction data, without relying on an explicit representation of the system dynamics. The latter class of methods is particularly attractive when accurate system models are difficult to obtain or computationally expensive to use \cite{sutton2018reinforcement,arulkumaran2017deep}.

A second key distinction concerns the way the policy is represented and optimized. In \emph{value-based} methods, the policy is derived implicitly from a learned action-value function, typically by selecting actions that maximize the estimated value \cite{watkins1992q} (see Equation \eqref{eq:opt_policyQ}).  
\emph{Policy-based} methods, instead, explicitly parametrize the policy and optimize it directly. Eventually, \emph{actor-critic} methods learn a policy (actor) and a separate value function approximation (the critic) that is used to evaluate actions and guide policy updates. This separation allows for more flexible policy representations and has proven especially effective in continuous action spaces \cite{sutton2018reinforcement,konda2000actor}.

Another important classification relates to how experience is collected and reused during learning. In \emph{on-policy} algorithms, the policy is updated using data generated by the current policy itself \cite{sutton2018reinforcement}. \emph{Off-policy} algorithms, on the other hand, learn from data collected by a possibly different behavior policy, which enables the reuse of past experience. This is typically achieved through an experience replay buffer, leading to improved sample efficiency \cite{mnih2015human,fujimoto2018td3}.

Finally, RL algorithms can be categorized as \emph{online} or \emph{offline}, depending on whether learning and data collection occur simultaneously. In online reinforcement learning, the agent continuously interacts with the environment and updates its policy as new data become available. This contrasts with offline reinforcement learning, where policy optimization is performed from a fixed dataset without further interaction \cite{levine2020offline}.

In this work we focus on online learning of control policies using off-policy, actor-critic RL algorithms. Off-policy algorithms tend to have a higher sample efficiency than on-policy ones due to a more effective re-use of the interaction data. Actor-critic algorithms, instead, represent the state-of-the-art for continuous control tasks. Despite several advances achieved in the last decades, state-of-the-art RL algorithms still lack the robustness and the generalization capabilities that are required to solve control problems involving parametrized dynamical systems. To this end, we introduce in the following the two building blocks for robust and generalizable RL, respectively the ensemble-based soft actor-critic SUNRISE and HypeRL. 

\subsection{Actor-Critic Algorithms}\label{subsect:ActorCritic}

In this section, we introduce soft-actor critic (SAC)~\cite{haarnoja2018soft} and twin-delayed deep deterministic policy gradient (TD3)~\cite{fujimoto2018addressing} as the RL algorithms underlying SUNRISE and HypeRL, respectively.

\subsubsection{Soft Actor-Critic}\label{subsec:SAC}

SAC~\cite{haarnoja2018soft} aims to maximize not only the expected cumulative reward but also policy-entropy maximization to improve exploration. SAC learns a stochastic policy $\pi_{\boldsymbol{\phi}}(\bm{a}_t | \bm{s}_t)=\pi(\bm{a}_t | \bm{s}_t;\boldsymbol{\phi})$ and the action-value function $Q_{\boldsymbol{\theta}}(\bm{s}_t, \bm{a}_t)=Q(\bm{s}_t, \bm{a}_t;\boldsymbol{\theta})$, both parametrized by neural networks of parameters $\boldsymbol{\phi}$ and $\boldsymbol{\theta}$, respectively. The stochastic policy is modeled as a Gaussian distribution
\begin{equation*}
\pi_{\boldsymbol{\phi}}(\bm{a}_t | \bm{s}_t) =
\mathcal{N}\big(
m_{\boldsymbol{\phi}}(\bm{s}_t),d^2_{\boldsymbol{\phi}}(\bm{s}_t)
\big)\, ,
\end{equation*}
where the policy network outputs the mean $m_{\boldsymbol{\phi}}(\bm{s}_t)$ and standard deviation $d_{\boldsymbol{\phi}}(\bm{s}_t)$ of the action distribution. The policy is optimized by minimizing the entropy-regularized actor objective $\mathcal{L}_{\pi}$ using samples, collected during the interaction of the agent with the environment, from an experience-replay memory buffer $\mathcal{B}$
\begin{equation}
\mathcal{L}_{\pi}(\boldsymbol{\phi})
=
\mathbb{E}_{\bm{s}_t \sim \mathcal{B},\, \bm{a}_t \sim \pi_{\boldsymbol{\phi}}(\bm{a}_t | \bm{s}_t)}
\left[-
Q_{\boldsymbol{\theta}}(\bm{s}_t,\bm{a}_t) +
\alpha \log \pi_{\boldsymbol{\phi}}(\bm{a}_t | \bm{s}_t)
\right]\, ,
\label{eq:actor_loss_SAC}
\end{equation}
where $\alpha$ is a coefficient weighting the contribution of the entropy regularization term $\log \pi_{\boldsymbol{\phi}}(\bm{s}_t)$ over the whole loss function, and $\mathbb{E}_{\bm{s}_t \sim \mathcal{B},\, \bm{a}_t \sim \pi_{\boldsymbol{\phi}}(\bm{a}_t | \bm{s}_t)}$ denotes the expected value when $\bm{s}_t$ is sampled from the memory buffer $\mathcal{B}$ and  $\bm{a}_t$ is sampled from the stochastic policy $\pi_{\boldsymbol{\phi}}(\bm{a}_t | \bm{s}_t)$.
The actor objective encourages the learning of policies that achieve high expected return by ascending the gradient of the value function (in practice descending the gradient of the value computed with the opposite sign), while maintaining sufficient entropy for effective exploration. The entropy term prevents the policy distribution from collapsing to its mean, i.e., becoming deterministic, which promotes more consistent exploration and improves robustness to errors in the value function approximation.

Using the data stored in the memory buffer $\mathcal{B}$, the critic parameters are updated by minimizing the following objective:
\begin{equation*}
\mathcal{L}_{Q}(\boldsymbol{\theta}) =
\mathbb{E}_{(\bm{s}_t,\bm{a}_t,r_t,\bm{s}_{t+1}) \sim \mathcal{B}}
\left[
\big(Q_{\boldsymbol{\theta}}(\bm{s}_t,\bm{a}_t) - y_t \big)^2
\right]\, ,
\end{equation*}
where the target for the critic update $y_t$ is given by
\begin{equation*}
y_t =
r_t + \gamma\,
\mathbb{E}_{\bm{a}_{t+1}\sim \pi_{\boldsymbol{\phi}}(\bm{a}_{t+1} | \bm{s}_{t+1})}
\left[
Q_{\bar{\boldsymbol{\theta}}}(\bm{s}_{t+1}, \bm{a}_{t+1})
- \alpha \log \pi_{\boldsymbol{\phi}}(\bm{a}_{t+1} | \bm{s}_{t+1})
\right]\, ,
\end{equation*}
and $Q_{\bar{\boldsymbol{\theta}}}(\bm{s}_{t+1},\bm{a}_{t+1})$ denotes the target critic networks with parameters $\bar{\boldsymbol{\theta}}$. Off-policy algorithms tend to suffer from training instabilities as the regression target of the critic may be generated by the critic -- this technique is often referred to as bootstrapping and it is commonly employed in, for example, temporal-difference learning \cite{sutton_reinforcement_2018}. Therefore, a target critic network, that is not updated by the gradients of the loss function, is typically used to "fix" the regression target and improve training stability. The parameters of the target network are updated at a slower rate than the critic using a soft update rule
\begin{equation*}
\bar{\boldsymbol{\theta}} \leftarrow \tau \boldsymbol{\theta} + (1-\tau)\bar{\boldsymbol{\theta}},
\end{equation*}
with $0 < \tau \ll 1$ controlling the speed of variation of the parameters.

\subsubsection{Twin Delayed Deep Deterministic Policy Gradient}
\label{subsec:td3}
TD3 \cite{fujimoto2018addressing} is an actor-critic algorithm developed to improve the training stability of the deep deterministic policy gradient (DDPG) \cite{lillicrap2015continuous}. In DDPG, the actor is updated by ascending the gradient of the value function, estimated by the critic network -- according to the \emph{deterministic} policy gradient theorem \cite{silver2014deterministic}. As a result, errors in the critic approximation may propagate to the policy, often leading to systematic overestimation of the action values and unstable learning \cite{fujimoto2018addressing}. TD3 learns a deterministic policy 
\begin{equation*}
\bm{a}_t = \pi_{\boldsymbol{\varphi}}(\bm{s}_t) = \pi(\bm{s}_t;\boldsymbol{\varphi})\, ,
\end{equation*}
and two independent critics 
\begin{equation*}
    Q_{\boldsymbol{\vartheta}_j}(\bm{s}_t, \bm{a}_t) = Q(\bm{s}_t,\bm{a}_t;\boldsymbol{\vartheta}_{j}),\quad j\in\{1,2\}.
\end{equation*}
The actor and the critics have a target network associated with parameters $\bar{\boldsymbol{\varphi}}, \bar{\boldsymbol{\vartheta}}_{1}, \bar{\boldsymbol{\vartheta}}_{2}$, respectively. 

Similarly to DDPG, the policy parameters are obtained by maximizing the action-value function according to the deterministic policy gradient
\begin{equation*}
\mathcal{L}_{\pi}(\boldsymbol{\varphi})
=
\mathbb{E}_{\bm{s}_t\sim\mathcal{B}}
\left[
- Q_{\boldsymbol{\vartheta}_{1}}(\bm{s}_t,\pi_{\boldsymbol{\varphi}}(\bm{s}_t))
\right].
\end{equation*}
However, the actor is updated less frequently than the critics -- hence the term "delayed".
When it comes to the critic update, TD3 introduces several modifications aimed at reducing approximation errors derived by overestimating the values. In particular, TD3 constructs conservative regression targets $y_t$ for critic updates by taking the minimum of the two target critics prediction:
\begin{equation*}
y_t = r_t + \gamma \min_{j=1,2} \bar{Q}_{\bar{\boldsymbol{\vartheta}}_{j}}(\bm{s}_{t+1},\bar{\pi}_{\bar{\boldsymbol{\varphi}}}(\bm{s}_{t+1})),
\end{equation*}
where $\bar{\pi}_{\bar{\varphi}}$ denotes the target actor network. Each critic is trained by minimizing the mean-squared error between the network prediction and the target $y_t$:
\begin{equation*}
\mathcal{L}_{Q}(\boldsymbol{\vartheta}_{j}) =
\mathbb{E}_{(\bm{s}_t,\bm{a}_t,r_t,\bm{s}_{t+1}) \sim \mathcal{B}}
\big[
\big(Q_{\boldsymbol{\vartheta}_{j}}(\bm{s}_t,\bm{a}_t) - y_t \big)^2
\big]\, , \quad j=1,2.
\end{equation*}
Eventually, the parameters of target networks are updated using a soft update rule
\begin{equation}
\begin{split}
\bar{\boldsymbol{\varphi}} &\leftarrow \tau \boldsymbol{\varphi} + (1-\tau)\bar{\boldsymbol{\varphi}}\, , \\
\bar{\boldsymbol{\vartheta}}_{j} &\leftarrow \tau \boldsymbol{\vartheta}_{j} + (1-\tau)\bar{\boldsymbol{\vartheta}}_{j}\, , \quad j=1,2\, ,
\end{split}
\label{eq:TD3_slow_update_targets}
\end{equation}
with $0 < \tau \ll 1$ controlling the speed of the update.

\subsection{SUNRISE: Ensemble-based Soft Actor-Critic}\label{subsubsec:SUNRISE}

SUNRISE \cite{lee2021sunrise} is a unified framework that combines the SAC algorithm (see Section \ref{subsec:SAC}) with ensemble-learning methods  for enhancing the robustness of RL to uncertainties. In particular, SUNRISE employs an ensemble of $N$ critics $\{Q_{\boldsymbol{\theta}_i}(\bm{s}_t, \bm{a}_t)=Q(\bm{s}_t, \bm{a}_t;\boldsymbol{\theta}_i)\}_{i=1}^N$ and $N$ actors $\{\pi_{\boldsymbol{\phi}_i}(\bm{a}_t | \bm{s}_t)=\pi(\bm{a}_t | \bm{s}_t;\boldsymbol{\phi}_i)\}_{i=1}^N$, where $\boldsymbol{\theta}_i$ and $\boldsymbol{\phi}_i$ denote the parameters of the $i$-th action-value function and the $i$-th policy, respectively. Similarly to SAC, each critic has a unique target Q-function that we indicate with $\bar{Q}_{\bar{\boldsymbol{\theta}}_i}$ with parameters denoted by $\bar{\boldsymbol{\theta}}_i$. Each (stochastic) policy is modeled as a Gaussian distribution
\begin{equation}
\pi_{\boldsymbol{\phi}_i}(\bm{a}_t | \bm{s}_t) = \pi(\bm{a}_t | \bm{s}_t;\boldsymbol{\phi}_i) = 
\mathcal{N}\big(
m_{\boldsymbol{\phi}_i}(\bm{s}_t),d^2_{\boldsymbol{\phi}_i}(\bm{s}_t)
\big)\, ,\quad i=1, \ldots,N\, ,
\label{eq:stochastic_policy}
\end{equation}
where the policy network outputs the mean $m_{\boldsymbol{\phi}_i}(\bm{s}_t)=m(\bm{s}_t;\boldsymbol{\phi}_i)$ and standard deviation $d_{\boldsymbol{\phi}_i}(\bm{s}_t)=d(\bm{s}_t;\boldsymbol{\phi}_i)$ of the action distribution.
Each policy of the ensemble is updated by minimizing the SAC objective (see Equation \eqref{eq:actor_loss_SAC})
\begin{equation}
\mathcal{L}_{\pi}(\boldsymbol{\phi}_i)
=
\mathbb{E}_{\bm{s}_t \sim \mathcal{B},\, \bm{a}_t \sim \pi_{\boldsymbol{\phi}_i}(\bm{a}_t | \bm{s}_t)}
\left[ -Q_{\boldsymbol{\theta}_i}(\bm{s}_t,\bm{a}_t)+
 \alpha_i \log \pi_{\boldsymbol{\phi}_i}(\bm{a}_t | \bm{s}_t) 
\right]\, ,\quad i=1, \ldots,N\, ,
\label{eq:actor_loss_sunrise}
\end{equation}
where $\mathcal{B}$ denotes the memory buffer collecting the experience tuples $(\bm{s}_t,\bm{a}_t,r_t,\bm{s}_{t+1})$, and $\alpha_i$ a scaling (and potentially learnable) coefficient balancing the contribution of the two terms of the loss function. The policy objective encourages the learning of policies that achieve high expected return -- ascending the action-value function $Q$ -- while maintaining sufficient entropy for effective exploration -- policy-entropy term. 

A key difference between SAC and SUNRISE is the update rule of the action-value function. SUNRISE explicitly exploits the ensemble of actors and critics (i) to weight the action-value function updates, and  (ii) to define an upper-confidence bound (UCB) exploration strategy. 

The action-value function update is weighted by a term proportional to the standard deviation of the critic estimates. In particular, the uncertainty-weighted critic loss that is minimized by SUNRISE is defined as
\begin{equation}
\mathcal{L}_{wQ}(\boldsymbol{\theta}_i)
= \mathbb{E}_{(\bm{s}_t,\bm{a}_t,r_t,\bm{s}_{t+1}) \sim \mathcal{B}, \ \bm{a}_{t+1} \sim \pi_{\phi_i}(\bm{a}_{t+1} | \bm{s}_{t+1})}\Big[
w(\bm{s}_{t+1}, \bm{a}_{t+1})
\left(
Q_{\boldsymbol{\theta}_i}(\bm{s}_t, \bm{a}_t) - y_t \right)^2
\Big]\,,
\label{eq:weighted_bellman}
\end{equation}
where the confidence weight $w(\bm{s}_{t+1},\bm{a}_{t+1})$ is defined as
\begin{equation*}
w(\bm{s}_{t+1},\bm{a}_{t+1})
=
\texttt{sigmoid}\!\left(
- \bar{Q}_{\mathrm{std}}(\bm{s}_{t+1},\bm{a}_{t+1}) \cdot T
\right)
+ \frac{1}{2}\,,
\end{equation*}
with \texttt{sigmoid}$(\cdot)$ denoting the sigmoid function, $T > 0$ indicating a temperature parameter, and $\bar{Q}_{\mathrm{std}}(\bm{s}_{t+1},\bm{a}_{t+1})$ the empirical standard deviation of the ensemble of target critics $\{\bar{Q}_{\bar{\boldsymbol{\theta}}_i}\}_{i=1}^N$.
To improve training stability and provide a fixed regression target $y_t$ for the action-value function, SUNRISE relies on the target network $\bar{Q}_{\bar{\boldsymbol{\theta}}_i}$, resulting in:
\begin{equation*}
y_t =
r_t + \gamma\,
\mathbb{E}_{\bm{a}_{t+1}\sim \pi_{\boldsymbol{\phi}_i}(\bm{a}_{t+1} | \bm{s}_{t+1})}
\left[
\bar{Q}_{\bar{\boldsymbol{\theta}}_i}(\bm{s}_{t+1}, \bm{a}_{t+1})
- \alpha_i \log \pi_{\boldsymbol{\phi}_i}(\bm{a}_{t+1} | \bm{s}_{t+1})
\right]\, .
\end{equation*} 
Each target network $\bar{Q}_{\bar{\boldsymbol{\theta}}_i}(\bm{s},\bm{a})$ is updated at a slower rate than the critic using a soft update rule
\begin{equation*}
\bar{\boldsymbol{\theta}}_i \leftarrow \tau \boldsymbol{\theta}_i + (1-\tau)\bar{\boldsymbol{\theta}}_i\,, \quad i=1,\ldots, N\,,
\end{equation*}
with $0 < \tau \ll 1$. 
The weighted critic update represents the first critical component of the SUNRISE framework. By modulating the contribution of each transition according to the uncertainty estimated from the critic ensemble, the update reduces the influence of unreliable or noisy target values. This mechanism improves training stability by preventing high-variance targets from dominating the value function updates. 

The second key ingredient of  SUNRISE is the use of the ensemble for efficient exploration. At each iteration, the action is selected by maximizing an UCB criterion of the form
\begin{equation}
\bm{a}^*_t
=
\arg\max_{\bm{a}_t}
\left(
Q_{\mathrm{mean}}(\bm{s}_t, \bm{a}_t) +
\eta\, Q_{\mathrm{std}}(\bm{s}_t, \bm{a}_t)
\right),
\label{eq:ucb}
\end{equation}
where $Q_{\mathrm{mean}}(\bm{s}_t, \bm{a}_t)$ and $Q_{\mathrm{std}}(\bm{s}_t, \bm{a}_t)$ denote the mean and standard deviation of the ensemble Q-function predictions, respectively, and $\eta > 0$ is a hyperparameter that balances the trade-off between exploitation and exploration.
The UCB strategy explicitly favors actions that are either expected to yield high returns or are associated with high uncertainty, encouraging the agent to explore underrepresented regions of the state-action space.  However, in continuous action spaces, such as the ones treated in this work, Equation \eqref{eq:ucb} is expensive to compute. Thus, to mitigate this issue, SUNRISE chooses the action out of the policy ensemble $ \{\pi_{{\boldsymbol{\phi}}_i}\}_{i=1}^N$ that maximizes~\eqref{eq:ucb}. After training, the best action is computed by taking the mean of the actions predicted by the actors.

\subsection{HypeRL: Hypernetwork-based Reinforcement Learning for Parametrized Dynamical Systems}\label{subsec:hyperl}

HypeRL is a hypernetwork-based actor-critic framework tailored to the control of parametrized system dynamics that conditions the policy and value function on the parameter vector $\boldsymbol{\mu}$ \cite{botteghi2025hyperl}. Hypernetworks are neural network architectures in which the parameters of a main network -- namely weights and biases -- are generated by a separate neural network \cite{ha2016hypernetworks}.

Instead of learning a single fixed set of parameters, hypernetworks enable conditional generation of model weights as a function of auxiliary inputs, allowing the representation of entire families of networks within a unified framework. Let $f(\cdot;\boldsymbol{\vartheta}_f)$ denote a main neural network, whose parameters $\boldsymbol{\vartheta}_f$  are generated by a hypernetwork $h(\cdot;\boldsymbol{\vartheta}_h)$ with parameters $\boldsymbol{\vartheta}_h$. Given a conditioning variable $\bm{z}$, the parameters of the target network are obtained as
\begin{equation}
\boldsymbol{\vartheta}_f = h(\bm{z};\boldsymbol{\vartheta}_h),
\label{eq:hyper_theta}
\end{equation}
where $\bm{z}$ may encode task-specific information, physical parameters, or contextual descriptors. For a given input $\bm{x}$, the output of the target network is then computed as
\begin{equation}
\hat{\bm{y}} = f(\bm{x};\boldsymbol{\vartheta}_f),
\label{eq:hyper_output}
\end{equation}
highlighting the dependence of the prediction on both the input $\bm{x}$ and the conditioning variable $\bm{z}$ through the generated parameters. Learning is performed by optimizing the hypernetwork parameters $\boldsymbol{\vartheta}_h$ with respect to a dataset of input-output pairs $\{(\bm{x}^{(j)}, \bm{z}^{(j)}, \bm{y}^{(j)})\}_{j=1}^M$. A typical training objective is given by
\begin{equation}
\mathcal{L}(\boldsymbol{\vartheta}_h) 
=
\sum_{j=1}^{M} \left\| \bm{y}^{(j)} - \hat{\bm{y}}^{(j)} \right\|_2^2\,,
\label{eq:hyper_loss}
\end{equation}
where $\hat{\bm{y}}^{(j)}$ is defined as in Equation~\eqref{eq:hyper_output}, where the dependence of $\boldsymbol{\vartheta}_f$ on $\bm{z}$ is induced by the hypernetwork mapping in Equation \eqref{eq:hyper_theta}. Gradient-based optimization is then used to update $\boldsymbol{\vartheta}_h$, propagating gradients through both the main network and the hypernetwork. This formulation allows smooth variations in the conditioning variable $\bm{z}^{(j)}$ to induce smooth variations in the generated parameters $\boldsymbol{\vartheta}_f$, enabling effective parameter sharing and conditional model generation. In control and RL settings, hypernetworks can be employed to generate policies or value functions that adapt continuously across different dynamical regimes or parameter configurations \cite{osband2018randomized, botteghi2025hyperl, Botteghi2025HypeMARL:Systems}.

HypeRL is an RL algorithm tailored to the control of parametrized dynamical systems, whose dynamics is defined as in Equation \eqref{eq:dynamics}. HypeRL enhances the expressivity of TD3 by replacing the standard neural networks with hypernetworks to learn context-dependent weights and biases of actor and critics. The weights of the actor and critic networks are dynamically generated as functions of a context vector defined as
\begin{equation*}
\bm{z}_t = [\bm{s}_t,\boldsymbol{\mu}]\, ,
\end{equation*}
where $\boldsymbol{\mu}$ may represent the vector characterizing the parameters of the dynamical system and/or task-dependent information. The hypernetwork-based parametrization enables the policy and value function to adapt their parametrization to different states and values of the system parameters. 

Given the context vector $\bm{z}_t$, three hypernetworks produce the parameters of the main policy $\boldsymbol{\varphi}$ and the two critic networks $\boldsymbol{\vartheta}_1$ and $\boldsymbol{\vartheta}_2$ according to
\begin{equation*}
\begin{split}
    \boldsymbol{\varphi} &= g_{\boldsymbol{\varphi}_{g}}(\bm{z}_t) = g(\bm{z}_t;\boldsymbol{\varphi}_{g})\, ,
\qquad
\boldsymbol{\vartheta}_j = h_{\boldsymbol{\vartheta}_{h_j}}(\bm{z}_t) = h_j(\bm{z}_t;\boldsymbol{\vartheta}_{h_{j}}), \quad j=1,2\, , \\
\bm{a}_t &= \pi_{\boldsymbol{\varphi}}(\bm{z}_t) = \pi(\bm{z}_t;\boldsymbol{\varphi})\, , \qquad 
Q_{\boldsymbol{\vartheta}_j}(\bm{z}_t, \bm{a}_t) = Q(\bm{z}_t,\bm{a}_t;\boldsymbol{\vartheta}_{j}),\quad j=1,2\,.
\end{split}
\end{equation*}
Similarly to TD3, HypeRL makes use of target (hyper)networks that we indicate with $\bar{\boldsymbol{\varphi}}, \bar{\boldsymbol{\varphi}_{g}}, \bar{\boldsymbol{\vartheta}}_1, \bar{\boldsymbol{\vartheta}}_{h_1}, \bar{\boldsymbol{\vartheta}}_2, \bar{\boldsymbol{\vartheta}}_{h_2}$ with slowly updating parameters (see Equation \eqref{eq:TD3_slow_update_targets}).

The policy parameters are jointly updated with their corresponding hypernetwork by maximizing
the action-value function according to the deterministic policy gradient
\begin{equation*}
\mathcal{L}_{\pi}(\boldsymbol{\varphi}_{g})
=
\mathbb{E}_{\bm{z}_t\sim\mathcal{B}}
\left[
- Q_{\boldsymbol{\vartheta}_{1}}(\bm{z}_t,\pi_{\boldsymbol{\varphi}}(\bm{z}_t))
\right].
\end{equation*}
The parameters of the critic networks $\boldsymbol{\vartheta}_1, \boldsymbol{\vartheta}_2$ and of the corresponding hypernetworks $\boldsymbol{\vartheta}_{h_1}, \boldsymbol{\vartheta}_{h_2}$ are then trained by minimizing:
\begin{equation*}
\mathcal{L}_{Q}(\boldsymbol{\vartheta}_{h_{j}})
=
\mathbb{E}_{(\bm{z}_t,\bm{a}_t,r_t,\bm{z}_{t+1})\sim\mathcal{B}}
\big[
 \texttt{Huber}(
Q_{\boldsymbol{\vartheta}_{j}}(\bm{z}_t,\bm{a}_t) - y_t)\big]\, , \quad j=1,2.
\end{equation*}
where
\begin{equation}
    \texttt{Huber}(x) = 
    \begin{cases}
        \frac{1}{2}x^2\, , \quad \text{if} \ |x| < 1 \\
        |x| - \frac{1}{2}\, , \quad \text{otherwise}
    \end{cases}\, ,
\label{eq:huber_loss}
\end{equation}
and $\bm{z}_{t+1}=[\bm{s}_{t+1},\boldsymbol{\mu}]$.
The Huber loss function reduces the influence of outliers in the target values and further improves the stability of critic learning \cite{fujimoto2023sale}. The value target $y_t$  is constructed using two target critic networks, and the target policy network as
\begin{equation*}
y_t = r_t + \gamma \min_{j=1,2} \bar{Q}_{\bar{\boldsymbol{\vartheta}}_{j}}\left(\bm{z}_{t+1}, \bar{\pi}_{\bar{\varphi}}(\bm{z}_{t+1})\right)\,.
\end{equation*}
The  target networks are updated using a soft update rule
\begin{equation}
\begin{split}
\bar{\boldsymbol{\varphi}}_g &\leftarrow \tau \boldsymbol{\varphi}_g + (1-\tau)\bar{\boldsymbol{\varphi}}_g\, , \\
\bar{\boldsymbol{\vartheta}}_{h_j} &\leftarrow \tau \boldsymbol{\vartheta}_{h_j} + (1-\tau)\bar{\boldsymbol{\vartheta}}_{h_j}\, , \quad j=1,2\, ,
\end{split}
\end{equation}
with $0 < \tau \ll 1$ controlling the update rate.

\section{Methodology}\label{sec:methodology}

In this section, we present HypEMBER, a hypernetwork-based ensemble RL algorithm for robust control of parametrized dynamical systems described by Equation \eqref{eq:dynamics}. HypEMBER combines the generalization capabilities of hypernetworks with the uncertainty quantification capabilities provided by an ensemble of policy and value functions. The integration of these two techniques enables HypEMBER to improve stability and data efficiency during training in ideal, noise-less environment settings, while showing robustness to measurement noise and uncertainties of system parameters at deployment time. HypEMBER adopts the SUNRISE training and inference structure (see Section \ref{subsubsec:SUNRISE}), while replacing the neural network parametrization with the hypernetwork-based formulation proposed in HypeRL (see Section \ref{subsec:hyperl}). In addition, we propose a novel action-selection strategy based on the ensemble of actors and critics suitable for the deployment of the agent in evaluation environments with measurement noise and uncertainties over the system parameters.

HypEMBER maintains an ensemble of $N$ actors and $N$ critics, whose parameters $\{\boldsymbol{\phi}_{i}\}_{i=1}^N, \{\boldsymbol{\theta}_{i}\}_{i=1}^N$ are generated by hypernetworks:
\begin{equation*}
\{\boldsymbol{\phi}_{i} = g_{i}(\bm{z}_t; \boldsymbol{\phi}_{g_{i}})\}_{i=1}^N\, ,
\qquad
\{\boldsymbol{\theta}_{i} = h_{i}(\bm{z}_t; \boldsymbol{\theta}_{h_{i}})\}_{i=1}^N\, ,
\end{equation*}
where $h_{i}$ and $g_{i}$ denote the hypernetworks associated with the $i^{\text{th}}$ critic and the $i^{\text{th}}$ actor with training parameters $\boldsymbol{\theta}_{h_{i}}$ and $\boldsymbol{\phi}_{g_{_i}}$, respectively, and $\bm{z}_t = [\bm{s}_t, \boldsymbol{\mu}]$ denotes the concatenation of the system state and the parameter vector. After the dynamic generation of the parameters, the policy and action-value functions take the following forms
\begin{equation*}
\{\pi_{\boldsymbol{\phi}_{i}}(\bm{a}_t | \bm{z}_{t})=\pi(\bm{a}_t | \bm{z}_{t}; \boldsymbol{\phi}_{i})\}_{i=1}^N\, ,
\qquad
\{Q_{\boldsymbol{\theta}_i}(\bm{z}_t, \bm{a}_t)=Q(\bm{z}_t, \bm{a}_t; \boldsymbol{\theta}_{i})\}_{i=1}^N\, ,
\end{equation*}
where each policy $\pi$ is stochastic and defined as in Equation \eqref{eq:stochastic_policy}. 

The critics are updated using the uncertainty-weighted critic loss:
\begin{equation}
\mathcal{L}_{wQ}(\boldsymbol{\theta}_{h_i})
= \mathbb{E}_{(\bm{z}_t,\bm{a}_t,r_t,\bm{z}_{t+1}) \sim \mathcal{B}, \ \bm{a}_{t+1} \sim \pi_{\phi_i}(\bm{a}_{t+1} | \bm{z}_{t+1})}\Big[
w(\bm{z}_{t+1}, \bm{a}_{t+1}) \
\texttt{Huber}\left(
Q_{\boldsymbol{\theta}_i}(\bm{z}_t, \bm{a}_t)
-
y_t
\right)\Big]\,, \quad i=1,\ldots,N\, ,
\label{eq:weighted_bellman_hyper}
\end{equation}
where $\bm{z}_{t+1}=[\bm{s}_{t+1}, \boldsymbol{\mu}]$. The parameters of the hypernetworks are jointly optimized with the critics through the minimization of Equation \eqref{eq:weighted_bellman_hyper}. Differently from Equation \eqref{eq:weighted_bellman}, we rely on the Huber loss \eqref{eq:huber_loss} instead of the mean-squared error to stabilize the training of the hypernetworks as discussed in Section \ref{subsec:hyperl}. The target critic $y_t$ is defined as:
\begin{equation*}
   y_t = r_t
+
\gamma \big[\bar{Q}_{\bar{\boldsymbol{\theta}_i}}(\bm{z}_{t+1},\bm{a}_{t+1})- \alpha_i \log \pi_{\boldsymbol{\phi}_i}(\bm{a}_{t+1} | \bm{z}_{t+1})\big].
\end{equation*}
The confidence weight $w(\bm{z}_{t+1},\bm{a}_{t+1})$ is computed from the ensemble of target critics and it is defined as
\begin{equation*}
w(\bm{z}_{t+1},\bm{a}_{t+1}) = \texttt{sigmoid}\!\left(- \bar{Q}_{\mathrm{std}}(\bm{z}_{t+1},\bm{a}_{t+1}) \cdot T \right) + \frac{1}{2}
\end{equation*}
with 
\begin{equation*}
\bar{Q}_{\mathrm{std}}(\bm{z}_{t+1},\bm{a}_{t+1}) =\mathrm{Std}\Big(\{\bar{Q}_{\bar{\boldsymbol{\theta}_i}}(\bm{z}_{t+1}, \bm{a}_{t+1}) \}_{i=1}^N\Big).
\end{equation*}

Similarly to the critic case, the actors and their respective hypernetworks are jointly optimized through an entropy-regularized actor objective (see Equation \eqref{eq:actor_loss_sunrise}):
\begin{equation*}
\mathcal{L}_{\pi}(\boldsymbol{\phi}_{g_i})
=
\mathbb{E}_{\bm{z}_t \sim \mathcal{B},\, \bm{a}_t \sim \pi_{\boldsymbol{\phi}_i}(\bm{a}_t | \bm{z}_t)}
\left[
\alpha_i \log \pi_{\boldsymbol{\phi}_i}(\bm{a}_t | \bm{z}_t)
-
Q_{\boldsymbol{\theta}_i}(\bm{z}_t,\bm{a}_t)
\right]\, , \quad i=1,\ldots,N\, ,
\end{equation*}
Eventually, we exploit the UCB criterion (see Section \ref{subsubsec:SUNRISE}) to guide the action selection and improve exploration during training:
\begin{equation*}
\bm{a}^*_t
=
\arg\max_{\bm{a}_{t}}
\left(
Q_{\mathrm{mean}}(\bm{z}_t, \bm{a}_t) +
\eta\, Q_{\mathrm{std}}(\bm{z}_t, \bm{a}_t)
\right),
\end{equation*}
where $Q_{\mathrm{mean}}(\bm{z}_t, \bm{a}_t)$ and $Q_{\mathrm{std}}(\bm{z}_t, \bm{a}_t)$ denote the mean and standard deviation of the ensemble Q-function predictions 
\begin{equation*}
\begin{split}
    Q_{\mathrm{mean}}(\bm{z}_{t},\bm{a}_{t}) &=\mathrm{Mean}\Big(\{Q_{\boldsymbol{\theta}_i}(\bm{z}_{t}, \bm{a}_{t}) \}_{i=1}^N\Big)\\
    Q_{\mathrm{std}}(\bm{z}_{t},\bm{a}_{t}) &=\mathrm{Std}\Big(\{Q_{\boldsymbol{\theta}_i}(\bm{z}_{t}, \bm{a}_{t}) \}_{i=1}^N\Big)\\
\end{split}
\end{equation*}
and $\lambda > 0$ is a hyperparameter that balances the trade-off between exploitation and exploration. The UCB promotes the selection of good actions -- in terms of average Q-values predicted by the ensemble -- and exploratory actions reducing the uncertainties -- represented by the standard deviation of the Q-values predicted by the ensemble. 

\begin{figure}
    \centering
    \includegraphics[width=1.0\linewidth]{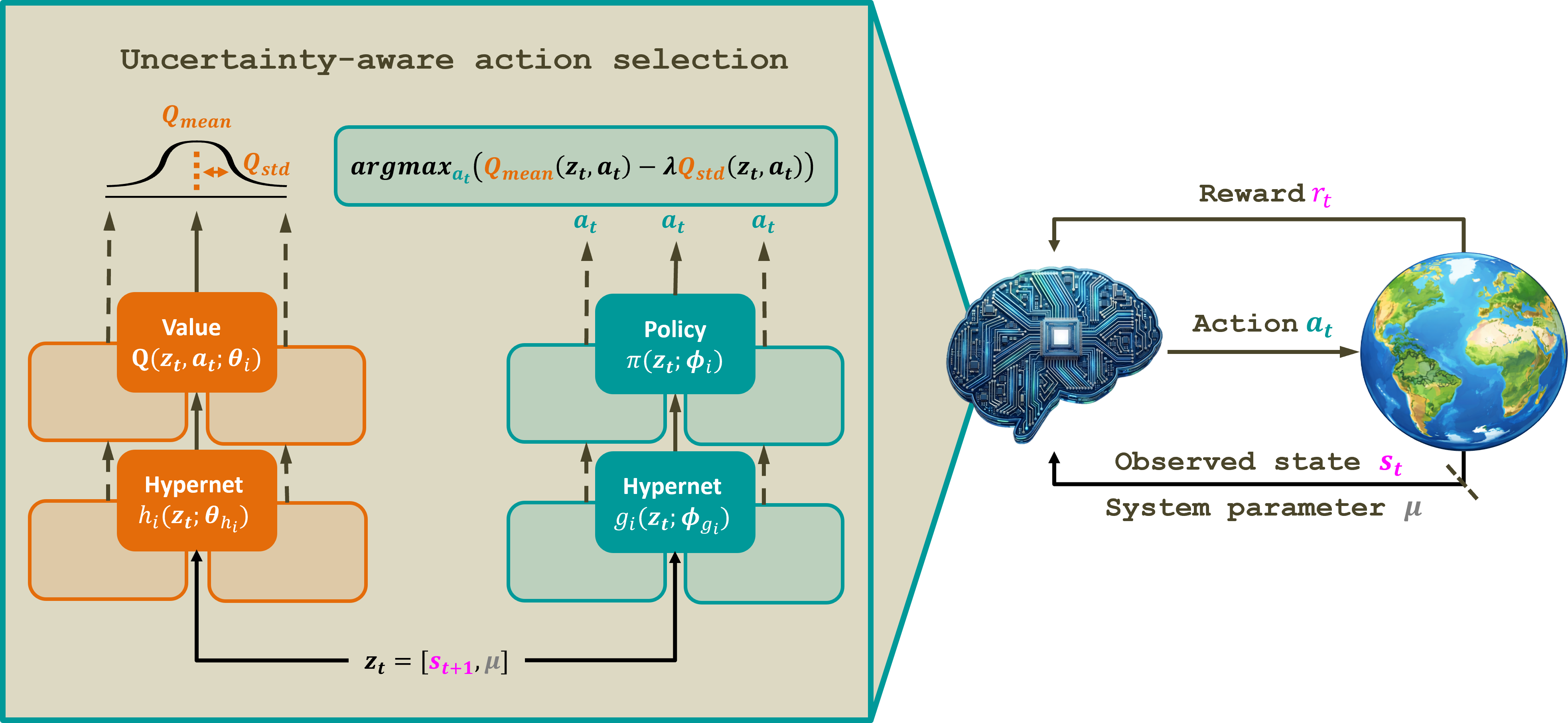}
    \caption{Uncertainty-aware action selection strategy. The ensemble of actors and critics is exploited to select optimal actions in terms of Q-values that, at the same time, reduce the uncertainties of the value estimates.}
    \label{fig:UA}
\end{figure}

In this work, we consider the problem of the mismatch between the training and evaluation environment when the latter is uncertain. When deploying agents in uncertain environment, simply selecting the average actions predicted by the ensemble, as SUNRISE does, may not guarantee the best performance. 
Inspired by the UCB, we develop an uncertainty-aware acting strategy that exploits the ensemble of actors and critics to account for uncertainties (see Figure \ref{fig:UA}). In particular, actions are selected to balance the maximization of the average Q-values, while at the same time minimizing the standard deviation of the Q-values:
\begin{equation*}
\bm{a}^*_t
=
\arg\max_{\bm{a}_{t}}
\left(
Q_{\mathrm{mean}}(\bm{z}_t, \bm{a}_t) -
\lambda\, Q_{\mathrm{std}}(\bm{z}_t, \bm{a}_t)
\right),
\end{equation*}
where $\lambda  \geq 0$ is a scalar factor weighting the contribution of the two terms.  

A detailed description of the proposed algorithm is provided in Algorithm~\ref{alg:hypEMBER} in Appendix~\ref{sec:pseudocode_HypeRMBER}, while the training hyperparameters are reported in Appendix~\ref{hyperparameters}. An ablation study on the value of $\lambda$ is shown in Appendix~\ref{app:additional_results_lambda}.

\section{Numerical Results}\label{sec:numerical_results}

Differently from the majority of the works in the literature, where RL algorithms are trained and evaluated in idealized simulation-based studies, toward assessing algorithm robustness, we compare five different RL agents -- trained in idealized (noise-free and parameter-consistent) simulation environments -- under perturbed conditions, including noisy observations and parameter misspecification. In practical scenarios, measurement noise may arise from sensor imperfections, numerical errors, or external disturbances. Moreover, additional sources of uncertainty are often associated with imperfect knowledge of the system dynamics, such as inaccuracies in physical parameters or modeling approximations.

In our numerical experiments, we consider three different scenarios: \emph{(i)} measurement noise, \emph{(ii)} modeling misspecification, and \emph{(iii)} combination of the two. We model uncertainties with Gaussian additive noise $\epsilon$ of the type:
\begin{equation*}
\boldsymbol{\epsilon} \sim \mathcal{N}(\bm{0}, \boldsymbol{\sigma}^2I)\, ,
\label{eq:gaussian_noise}
\end{equation*}
where $\boldsymbol{\sigma}=[\sigma_1, \cdots, \sigma_n]$ denotes the standard deviation of the noise and $n$ depends on the dimension of the measurements or the parameter vector. After training the agents without uncertainties, during the evaluation phase, robustness is assessed by progressively increasing the noise intensity to quantify performance degradation. 

Measurement noise is introduced by perturbing the observations available to the agent. Let $\bm{s}_i$ denote the $i$-th component of the measured state, with $i = 1, \ldots, N_s$ indexing the available sensors. The corresponding noisy observation is given by
\begin{equation}
\tilde{\bm{s}}_t = \bm{s}_t + \boldsymbol{\epsilon}_{\text{M}}\, ,
\label{eq:measurement_noise}
\end{equation}
where $\boldsymbol{\epsilon}_{\text{M}} \sim \mathcal{N}(\bm{0}, \boldsymbol{\sigma}^2_{\text{M}}I) \in \mathbb{R}^{N_s}$ is the noise vector used for the measurements with standard deviation $\boldsymbol{\sigma}_{\text{M}}$.
Model misspecification is introduced by perturbing the parameters governing the system dynamics. Denoting by $\boldsymbol{\mu}$ a generic system parameter, the perturbed parameter is defined as
\begin{equation}
\tilde{\boldsymbol{\mu}} = \boldsymbol{\mu} + \boldsymbol{\epsilon}_{\text{P}}\, ,
\label{eq:model_misspecification}
\end{equation}
where $\boldsymbol{\epsilon}_{\text{P}} \sim \mathcal{N}(\bm{0}, \boldsymbol{\sigma}^2_{\text{P}}I) \in \mathbb{R}^{N_{\mu}}$ is the noise vector used for perturbing the parameter vector $\boldsymbol{\mu}$  with standard deviation $\boldsymbol{\sigma}_{\text{P}}$.

In our numerical experiments, we compare HypEMBER with four different state-of-the-art algorithms, namely:
\begin{itemize}
    \item[-] \emph{SUNRISE} (see Section \ref{subsubsec:SUNRISE}),
    \item[-] \emph{HypeRL} (see Section \ref{subsec:hyperl}),
    \item[-] \emph{TD3} (see Section \ref{subsec:td3}), and
    \item[-] \emph{PolyL0-TD3} -- a variant of TD3 in which the neural network policy is replaced with a sparse polynomial policy (see Appendix \ref{polyl0} for more detail).
\end{itemize}
We compare the different algorithms not only in terms of reward accumulated over training, but also on their robustness when evaluated under uncertainties. The hyperparameters used in our experiments are reported in Appendix \ref{hyperparameters}.

\subsection{Stabilization of a Parametric Kuramoto-Sivashinsky Equation}
\label{subsec:kuramoto}

As first test case, we consider the stabilization of a Kuramoto–Sivashinsky (KS) equation, a nonlinear PDE that arises in a variety of physical contexts, including flame front propagation, reaction-diffusion systems, and thin film flows \cite{kudryashov1990exact}. For sufficiently large domain lengths, the uncontrolled KS equation exhibits chaotic behavior, making the stabilization task particularly challenging. As a consequence, the KS equation has been widely adopted as a reference problem for the control of nonlinear PDEs \cite{peitz2023distributed, Boucher2025EvidenceLearning, bucci2019control, botteghi2025hyperl}.

Assuming the state of the KS denoted by $y(x,t)$, where $x$ indicates the spatial variable and $t$ the time variable, we can write the KS equation with the addition of a parametric spatial cosine forcing term as:
\begin{equation}
\frac{\partial y(x,t)}{\partial t}
+
y(x,t)\frac{\partial y(x,t)}{\partial x}
+
\frac{\partial^2 y(x,t)}{\partial x^2}
+
\frac{\partial^4 y(x,t)}{\partial x^4}
+
\mu \cos\!\left(\frac{4\pi x}{L}\right)
=
\omega(x,t), \quad x \in (0, L), t \in (0, T)
\label{eq:ks_forced}
\end{equation}
where $\mu \in (-0.25,0.25)$ is our parameter of interest modulating the contribution of the spatial cosine forcing term, $\omega(x,t)$ is the control input function, $L=22$ with periodic boundary conditions, i.e., $y(0, t)=y(L,t)$, and $T=300$. The control function $\omega(x,t)$ is defined as a linear combination of spatially localized Gaussian actuators 
\begin{equation*}
\omega(x,t) = \sum_{i=1}^{N_a} a_i(t)\,\psi(x,m_i),
\label{eq:ks_control}
\end{equation*}
where $N_a = 8$ denotes the number of actuators, $a_i(t) \in (-1,1)$ are time-dependent and learnable control coefficients -- namely the RL policy outputs $\bm{a}_t=[a_{1,t}, \ldots,a_{N_a, t}]$ -- and $m_i$ represent equally spaced actuator locations. Each actuator shape is modeled by a Gaussian kernel of the form
\begin{equation*}
\psi(x,m_i)
=
\frac{1}{2}
\exp\!\left(
- \left(\frac{x - m_i}{\sigma}\right)^2
\right)\,,
\label{eq:ks_actuator}
\end{equation*}
where $\sigma = 0.8$ controls the spatial width of the actuation. To numerically solve  the state equation in \eqref{eq:ks_forced}, we employ a Fourier pseudo-spectral semi-implicit Crank–Nicolson Adams–Bashforth solver and we discretize the spatial domain in $N_x=64$ grid points and the time domain with a $\Delta t = 0.1$, leading to a state vector $\bm{y}_t \in \mathbb{R}^{N_x}$.
The agent state corresponds to the value of $N_s=8$ equally spaced sensors across the spatial domain, concatenated with the one-dimensional parameter $\mu$, namely $\bm{s}_t = [u_{1,t}, \ldots, u_{N_s,t},\mu]$. 

The control objective consists of stabilizing the system around the equilibrium solution $\bm{s}_{\mathrm{ref}}=\bm{0}$ for different values of the parameter $\mu$, while keeping the applied control effort limited (see \eqref{eq:reward_function}). The control is activated at $t = 100$, after the uncontrolled flow has developed. The state cost penalizes deviations of the current KS state from the desired equilibrium. Minimizing the state cost -- namely maximizing its negative -- therefore enforces stabilization by driving the solution toward $\bm{s}_{\mathrm{ref}}=\bm{0}$. On the other side, the action cost penalizes large control inputs and acts as a regularization term that discourages unnecessarily strong actuation.  In this work, we set $\beta=0.1$, prioritizing stabilization of the state while still preventing excessively large control actions. During training, the control policies are learned by randomly sampling the parameter $\mu$ at the beginning of each episode from the discrete set  of equally spaced value from $-0.225$ to $0.225$ with step $0.005$. To evaluate generalization capabilities, the trained policies are tested on previously unseen parameter values randomly sampled from the interval $\mu \in [-0.25,\,0.25]$.

In Table \ref{tab:training_rew_ks}, we show the cumulative reward over training collected by the five different agents when trained on 1000 episodes. 
\begin{table}[b!]
\centering
\begin{tabular}{l|c|c|c|c|c}
\hline
Training reward & HypeRL & HypEMBER & SUNRISE & PolyL0-TD3 & TD3 \\
\hline
\hline
Mean $\pm$ Std 
& $\mathbf{-70.4 \pm 6.2}$ 
& $-82.9 \pm 8.4$ 
& $-143.0 \pm 77.9$ 
& $-186.4 \pm 66.8$ 
& $-207.9 \pm 48.5$ \\
\hline
\end{tabular}
\vspace{2pt}
\caption{Training rewards collected by the different agents. We highlight in \textbf{bold} the best performing agent.}
\label{tab:training_rew_ks}
\end{table}
While HypeRL obtains the best training performance, HypEMBER is able to achieve similar rewards at the end of the training. TD3 instead achieves the worst performance, followed by PolyL0-TD3, and SUNRISE. A closer inspection of the training dynamics highlights that HypEMBER is characterized by improved stability compared to SUNRISE, and shows smoother learning trajectories and reduced variability across seeds, suggesting a more reliable and robust optimization process. 

\begin{figure}[h!]
    \centering
    \begin{minipage}{0.49\textwidth}
        \centering
        \includegraphics[width=\textwidth]{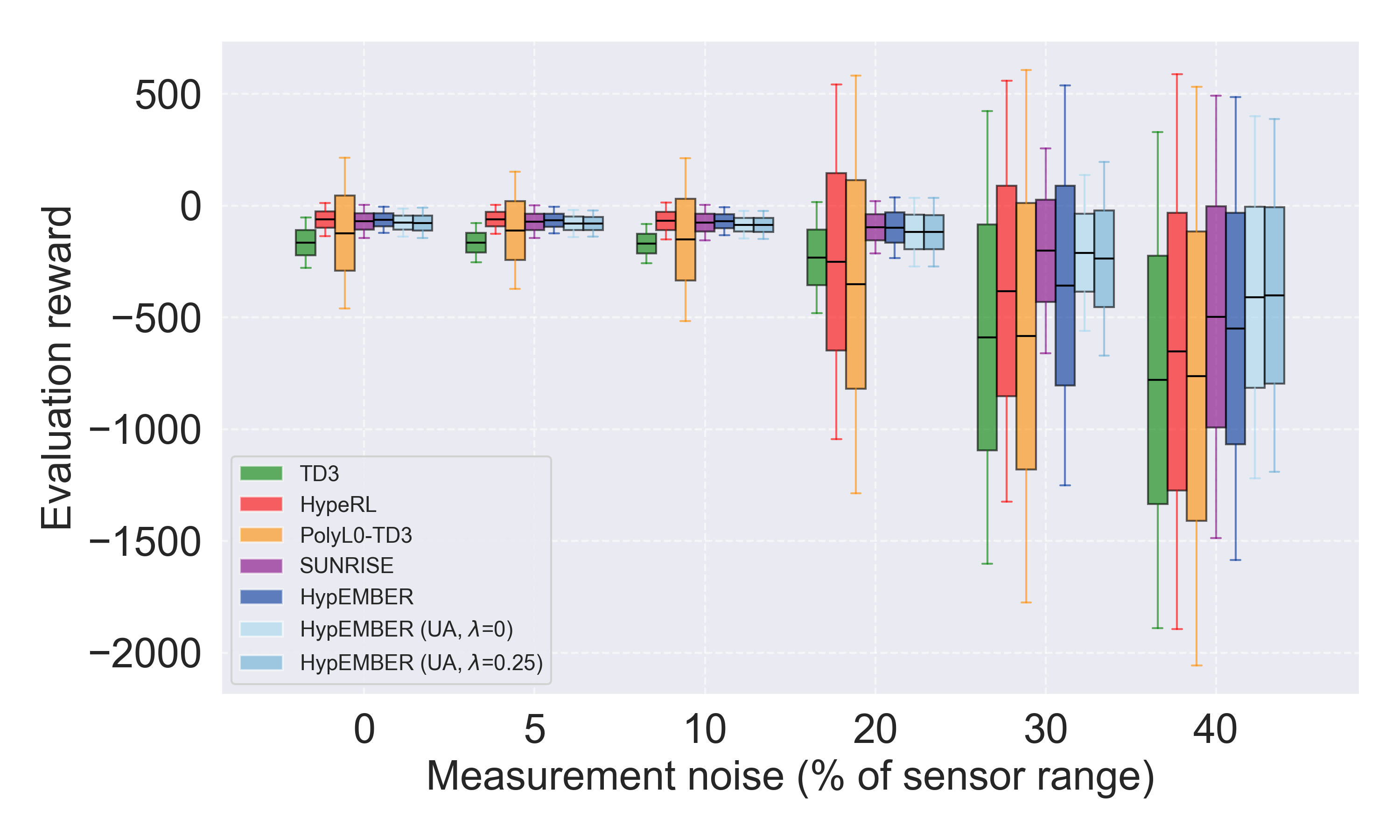}
        \caption{Performance of the different agents when evaluated with measurement noise.}
        \label{fig:measurment_noise_ks}
    \end{minipage}
    \hfill
    \begin{minipage}{0.49\textwidth}
        \centering
        \includegraphics[width=\textwidth]{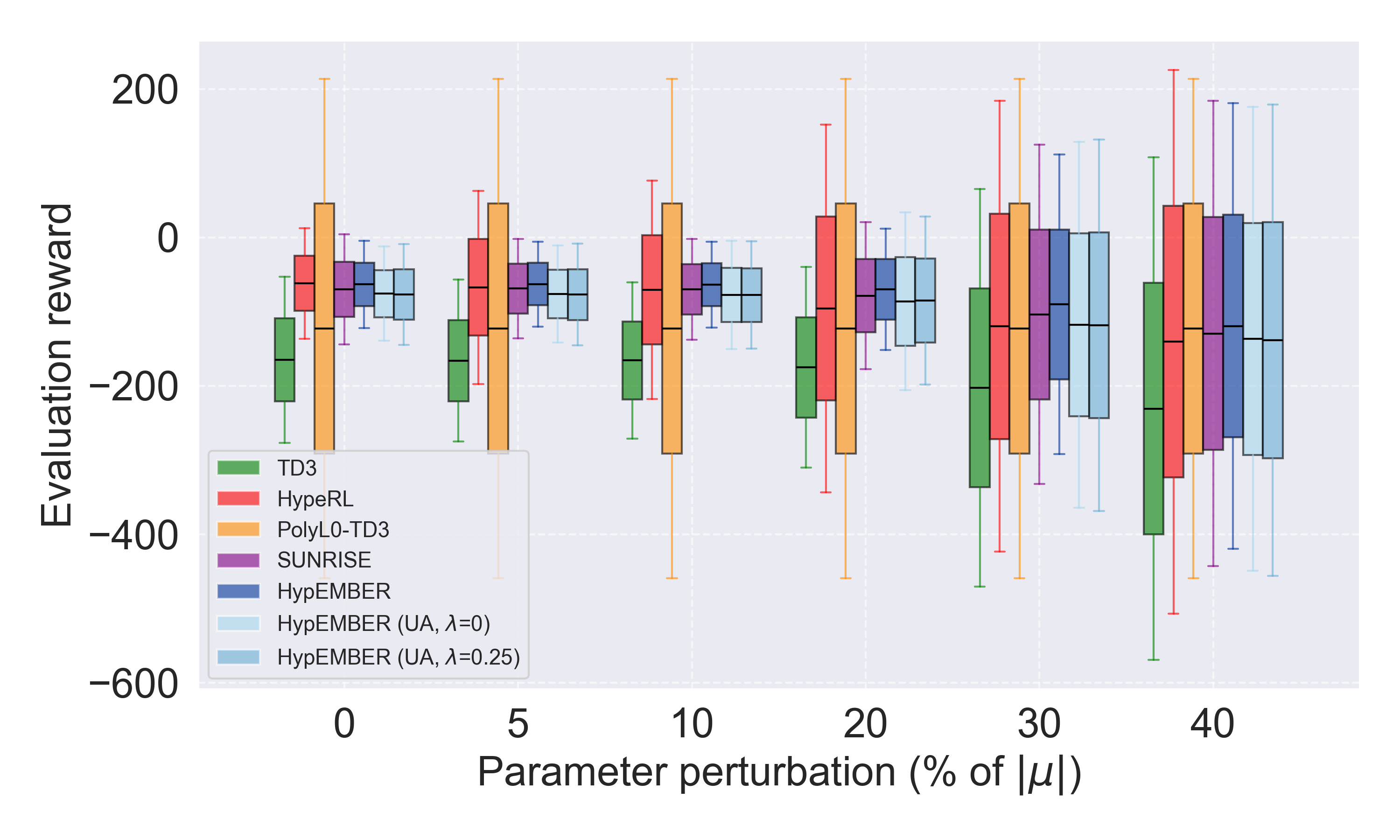}
        \caption{Performance of the different agents when evaluated with perturbation on the physical parameter.}
        \label{fig:misspecification_ks}
    \end{minipage}
        \hfill
    \begin{minipage}{0.49\textwidth}
        \centering
        \includegraphics[width=\textwidth]{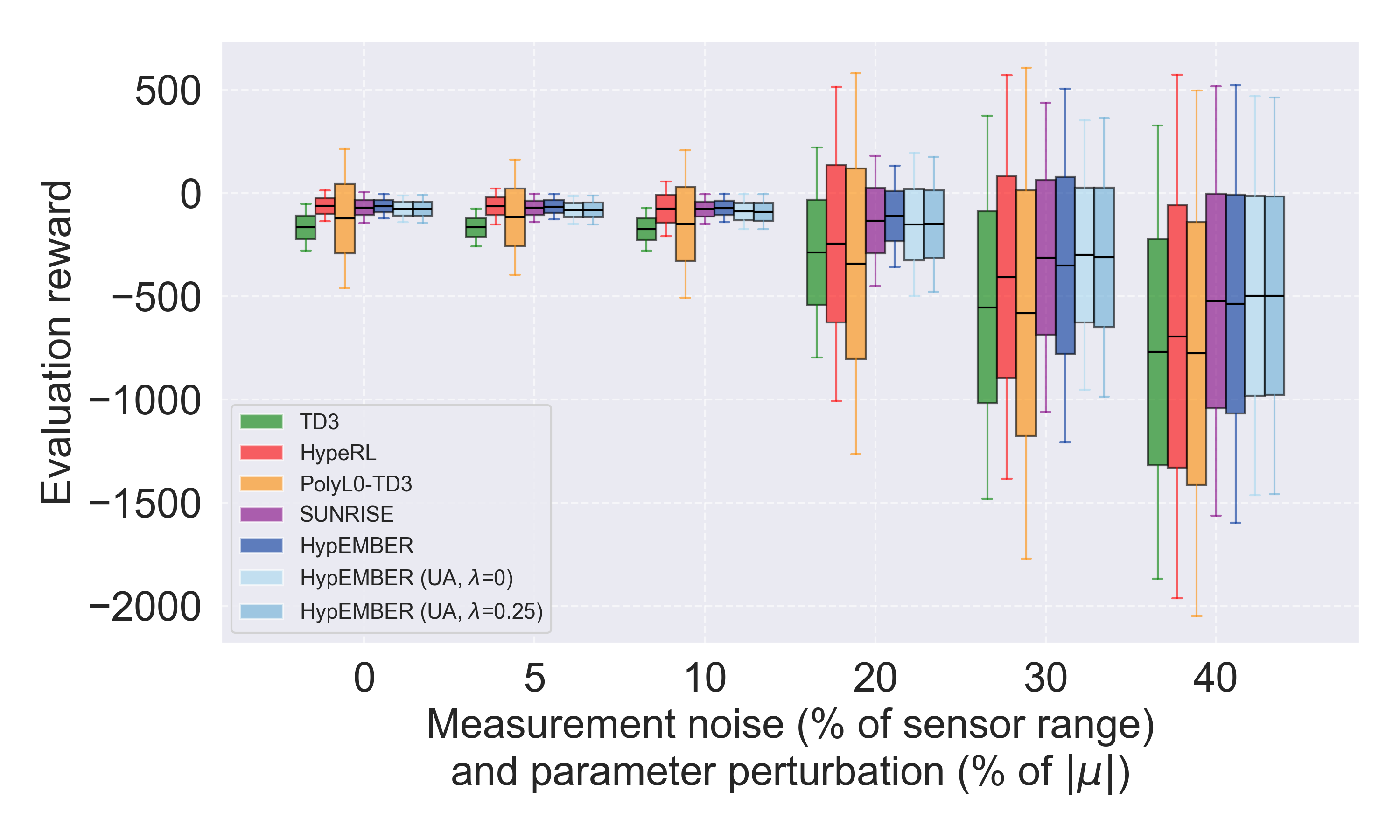}
        \caption{Performance of the different agents when evaluated with measurement noise and perturbation on the physical parameter.}
        \label{fig:combo_ks}
    \end{minipage}
\end{figure}

In our numerical experiments, we consider three different scenarios: \emph{(i)} measurement noise, \emph{(ii)} modeling misspecification, and \emph{(iii)} combination of the two.
The critical investigation of our study involves the evaluation of the agents in the presence of uncertainties (not seen during training). To do so, for each algorithm and seed, we assess the performance of the agents over 20 independent evaluation episodes with a randomly sampled $\mu$. The additive noise intensity -- namely $\boldsymbol{\sigma}_M$ and $\boldsymbol{\sigma}_P$ (see Equation \eqref{eq:measurement_noise} and \eqref{eq:model_misspecification}) -- is varied from $0\% $ (ideal testing scenario) to $40\%$ of the signals, allowing a systematic assessment of performance degradation as uncertainties increase. In Figures \ref{fig:measurment_noise_ks}, \ref{fig:misspecification_ks}, and \ref{fig:combo_ks} we report the results (mean and standard deviation of the rewards) obtained from the robustness analysis.
In general, the performance of all the agents degrades with the increment of the uncertainties and we can identify a large drop in the performance usually between $10\%-30\%$. However, not all the agents are affected by the uncertainties in the same way:
\begin{itemize}
    \item[-] TD3 shows consistent performance up to $20\%$ noise, with a quick degrade after.
    \item[-] PolyL0-TD3 exhibits the highest standard deviation across all the three different scenarios, but it seems the least affected by perturbations over the parameters (the reward standard deviation does not increase proportionally to the increment of the uncertainties, although very high from the beginning). An inspection of the learned coefficients reveals that those associated with the physical parameter $\mu$ are close to zero, effectively reducing the sensitivity of the policy to parameter variations. As a result, the agent is only weakly affected by changes in the parameter value, leading to a comparatively stable, although suboptimal, performance.
    \item[-] The two ensemble-based approaches, namely SUNRISE and HypEMBER, are capable of best coping with higher levels of noise and best mitigating the degradation due to uncertainties, with HypEMBER usually achieving higher rewards.
    \item[-] Depending on the perturbation intensity, HypeRL and HypEMBER alternately achieve the best performance overall. This result suggests that explicitly conditioning the policy on physical parameters using hypernetworks, as done by HypeRL, is advantageous for moderate levels of uncertainties (up to $10\%$), while for higher levels it is essential to employ the additional uncertainty-aware mechanisms enabled by the hypernetwork-ensemble learning used by HypEMBER.
    \item[-] The uncertainty-aware (UA) action selection appears beneficial in the presence of measurement uncertainties and in the case of combined uncertainties (measurements and parameters), while it does not improve performance in the case of modeling errors compared to the default action-selection strategy (mean action).
\end{itemize}

\begin{figure*}[h!]
    \centering

    \begin{minipage}{0.49\linewidth}
        \centering
        \subfloat{\includegraphics[width=\linewidth]{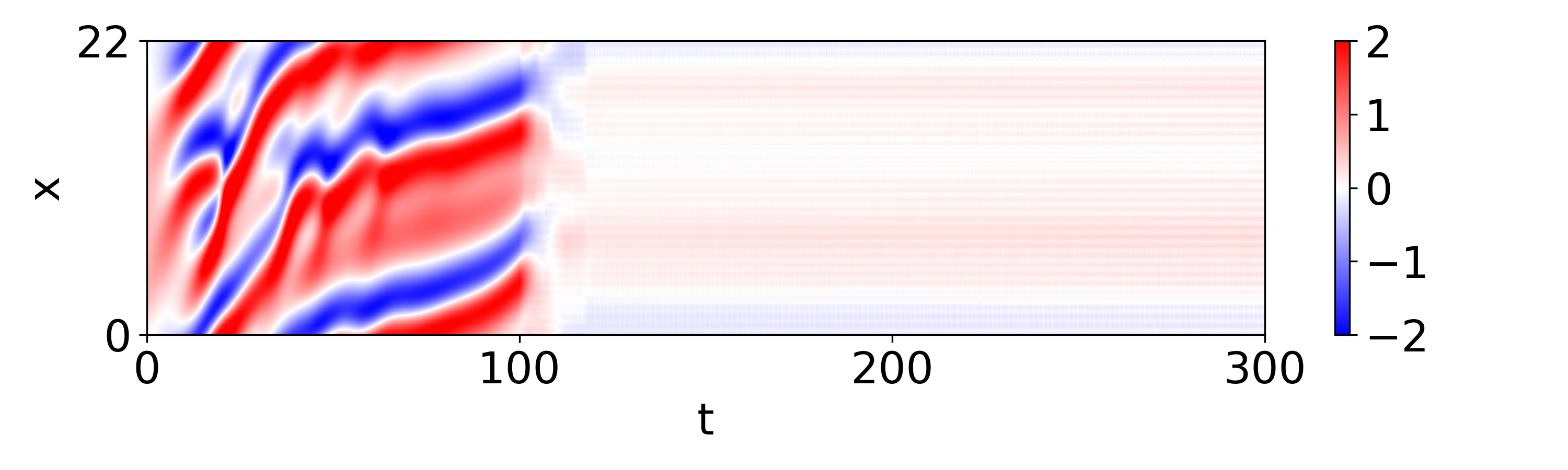}}\\[-0.1cm]
        \subfloat{\includegraphics[width=\linewidth]{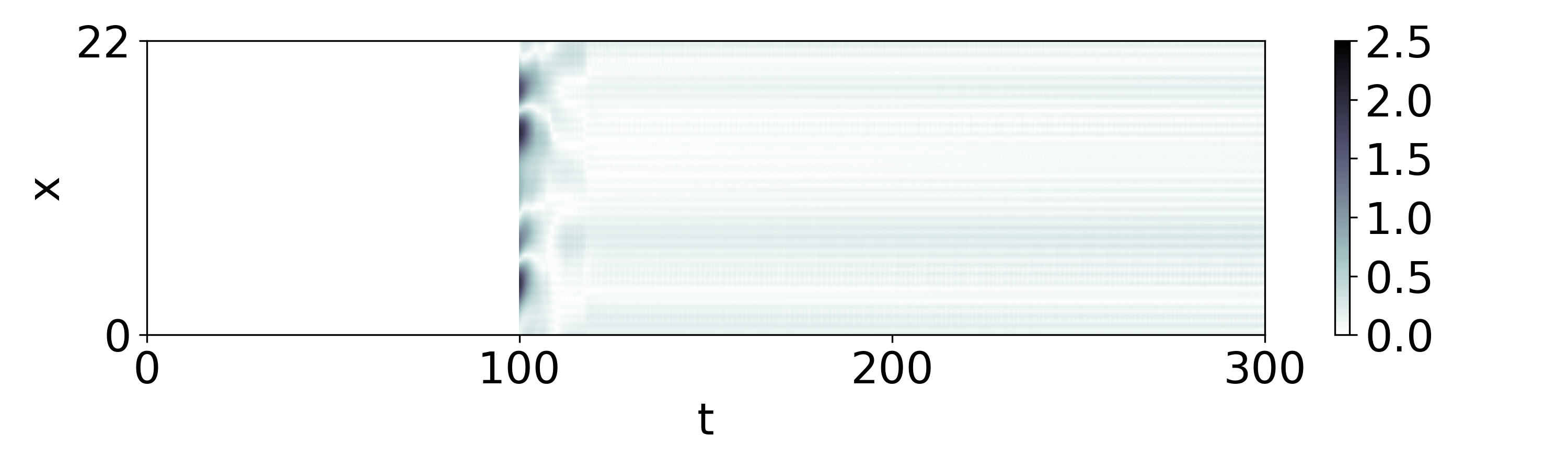}}
    \end{minipage}
    \hfill
    \begin{minipage}{0.49\linewidth}
        \centering
        \subfloat{\includegraphics[width=\linewidth]{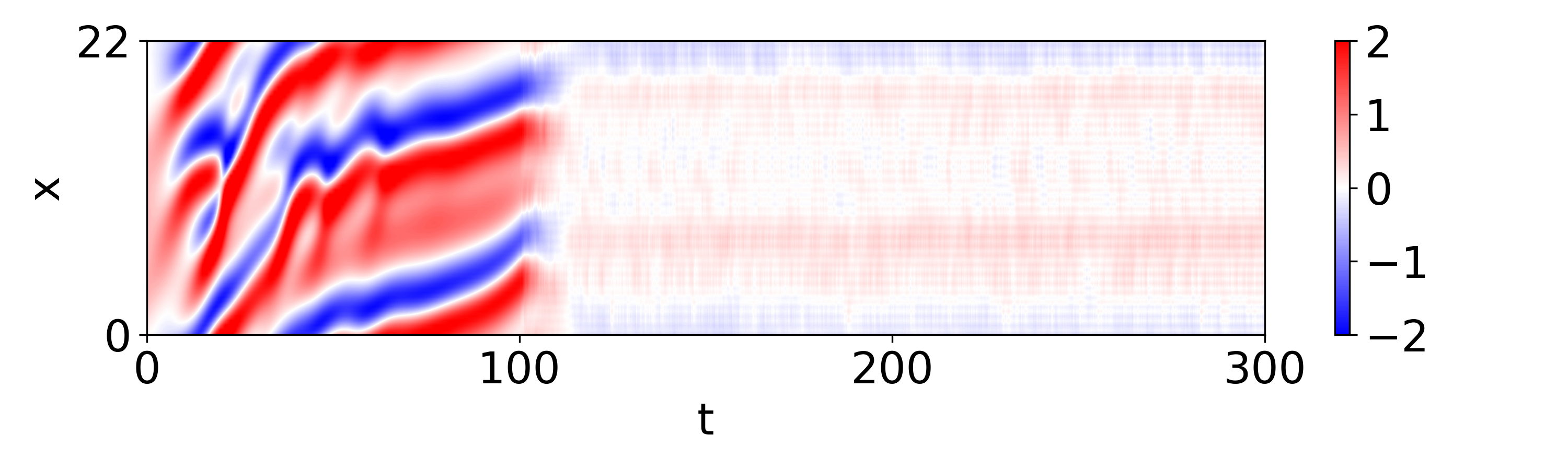}}\\[-0.1cm]
        \subfloat{\includegraphics[width=\linewidth]{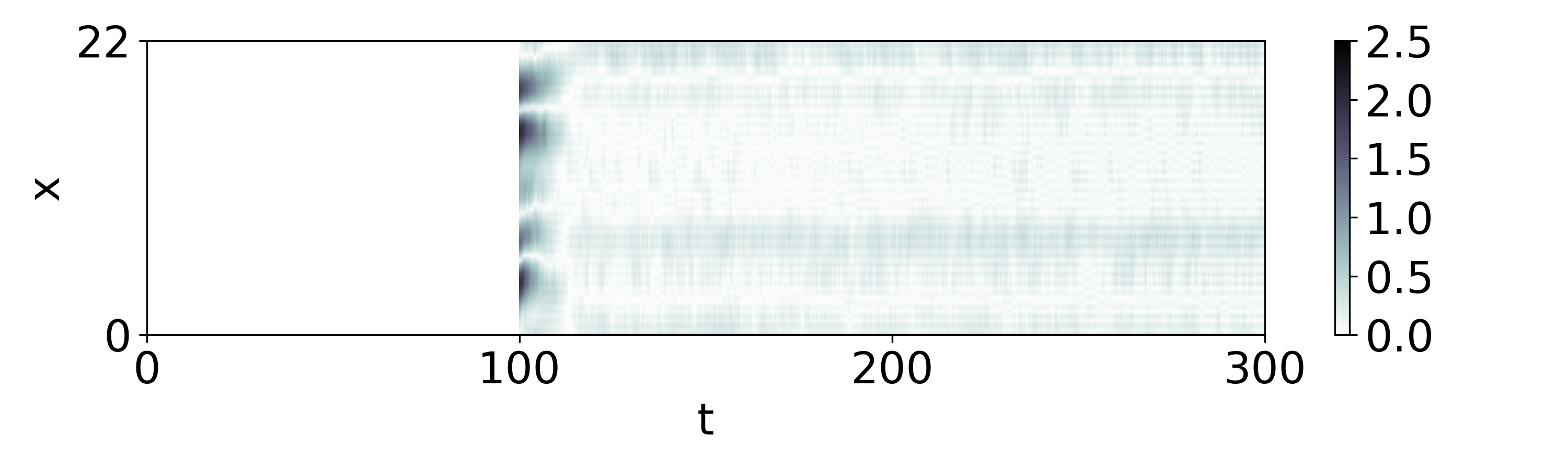}}
    \end{minipage}

    \caption{Controlled solutions for the HypEMBER (uncertainty-aware action selection with $\lambda = 0.25$) case. Left column: clean setting. Right column: combined setting. From top to bottom: controlled solution, and absolute state-tracking error.}
    \label{fig:hypEMBER_clean_noise}
\end{figure*}

\begin{figure*}[h!]
    \centering
    \begin{minipage}{0.49\linewidth}
        \centering
        \subfloat{\includegraphics[width=\linewidth]{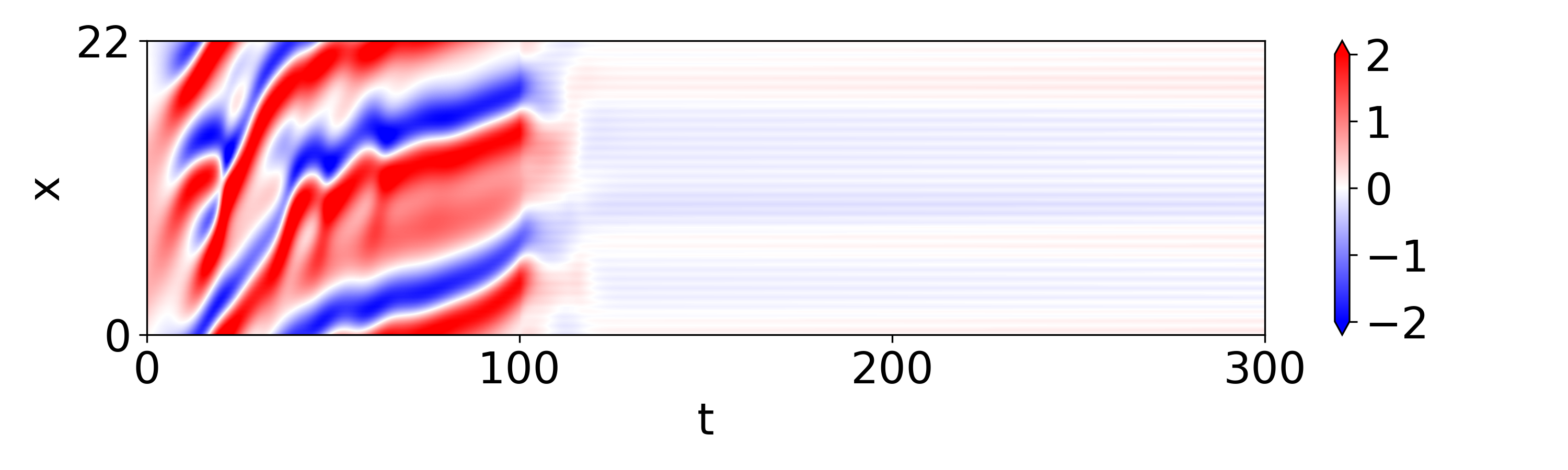}}\\[-0.1cm]
        \subfloat{\includegraphics[width=\linewidth]{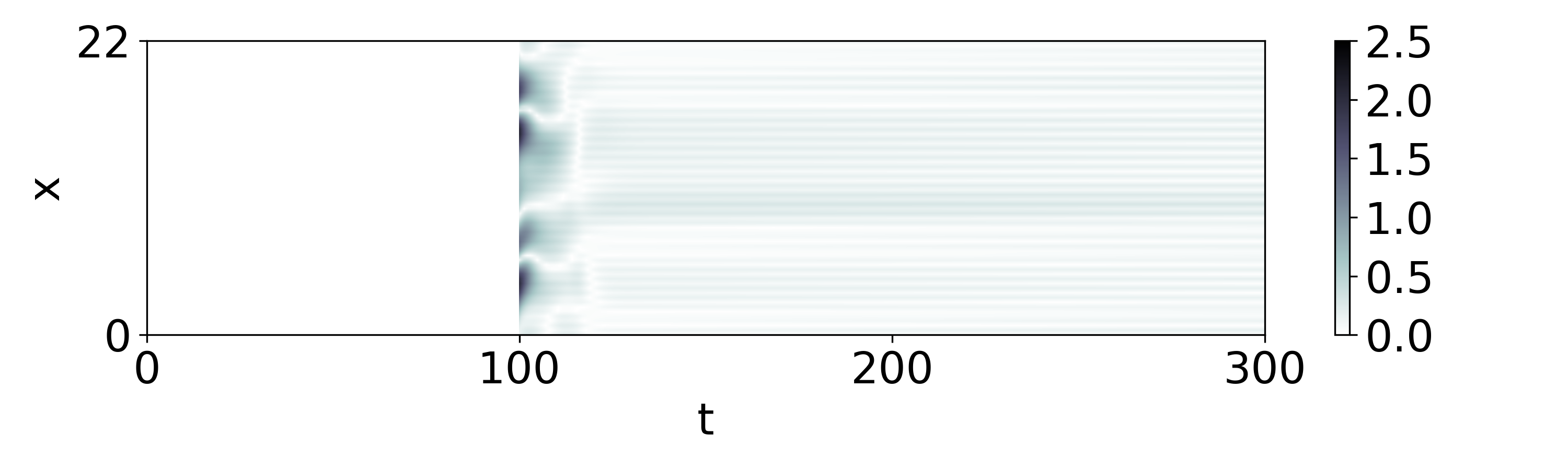}}
    \end{minipage}
    \hfill
    \begin{minipage}{0.49\linewidth}
        \centering
        \subfloat{\includegraphics[width=\linewidth]{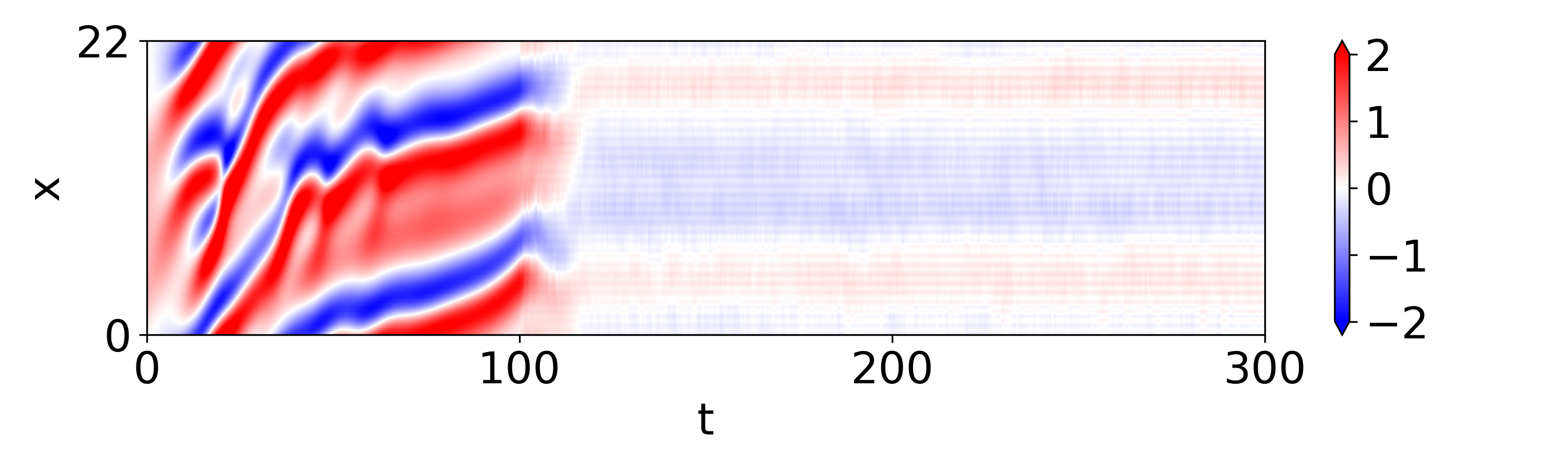}}\\[-0.1cm]
        \subfloat{\includegraphics[width=\linewidth]{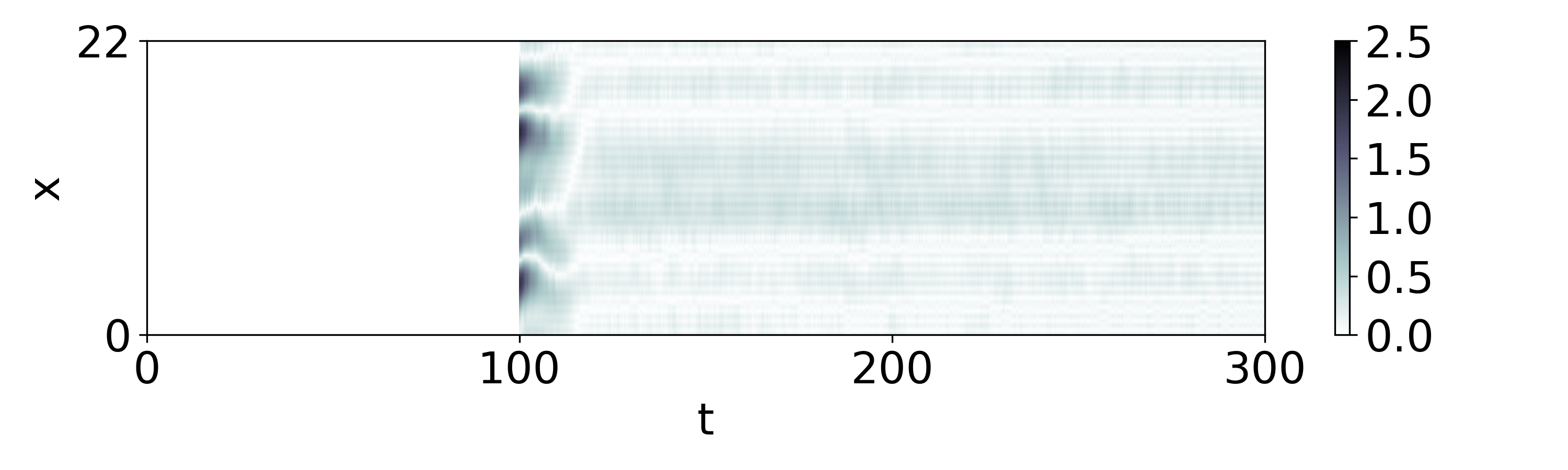}}
    \end{minipage}
    \caption{Controlled solutions for the SUNRISE case. Left column: ideal settings. Right column: noisy setting. From top to bottom: controlled solution, and absolute state-tracking error.}
    \label{fig:sunrise_clean_noise}
\end{figure*}

\begin{figure*}[h!]
    \centering
    \begin{minipage}{0.49\linewidth}
        \centering
        \subfloat{\includegraphics[width=\linewidth]{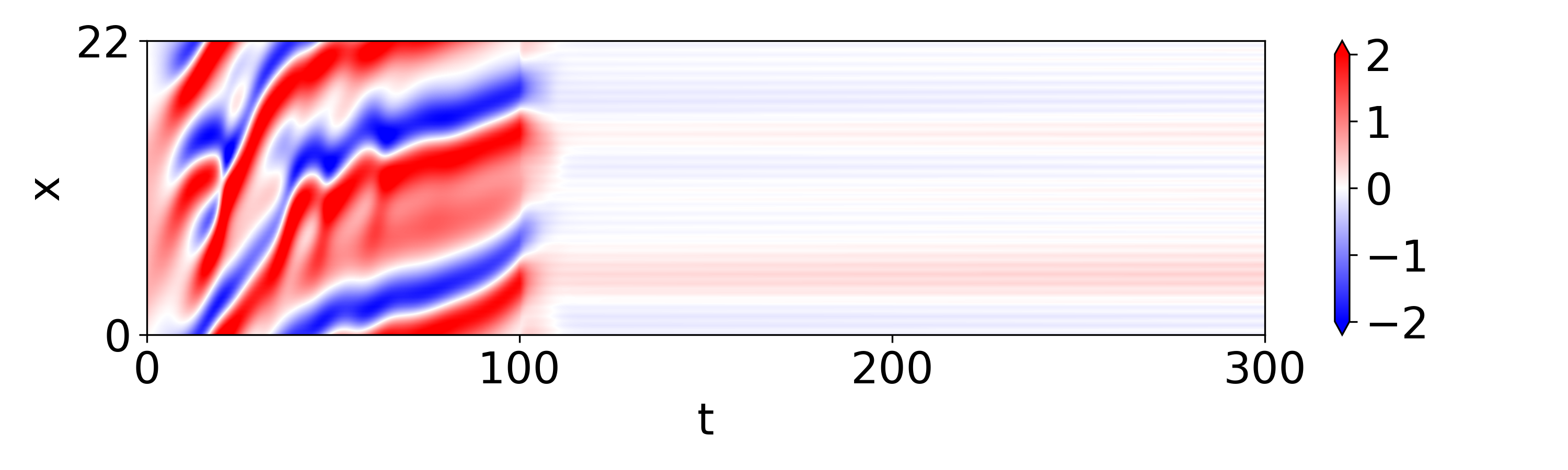}}\\[-0.1cm]
        \subfloat{\includegraphics[width=\linewidth]{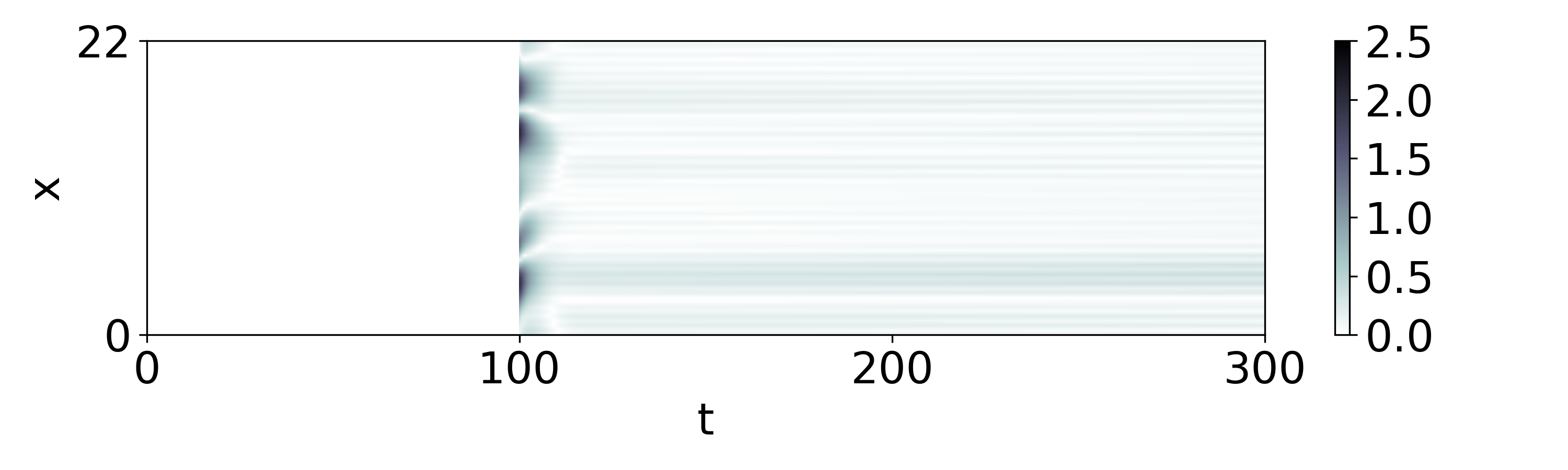}}
    \end{minipage}
    \hfill
    \begin{minipage}{0.49\linewidth}
        \centering
        \subfloat{\includegraphics[width=\linewidth]{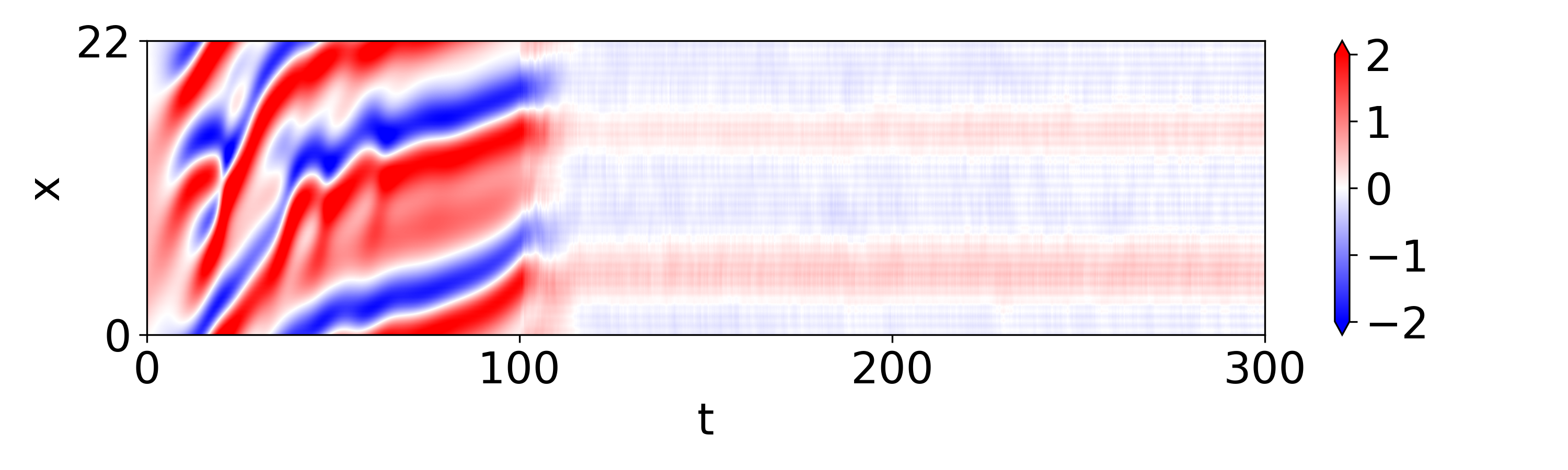}}\\[-0.1cm]
        \subfloat{\includegraphics[width=\linewidth]{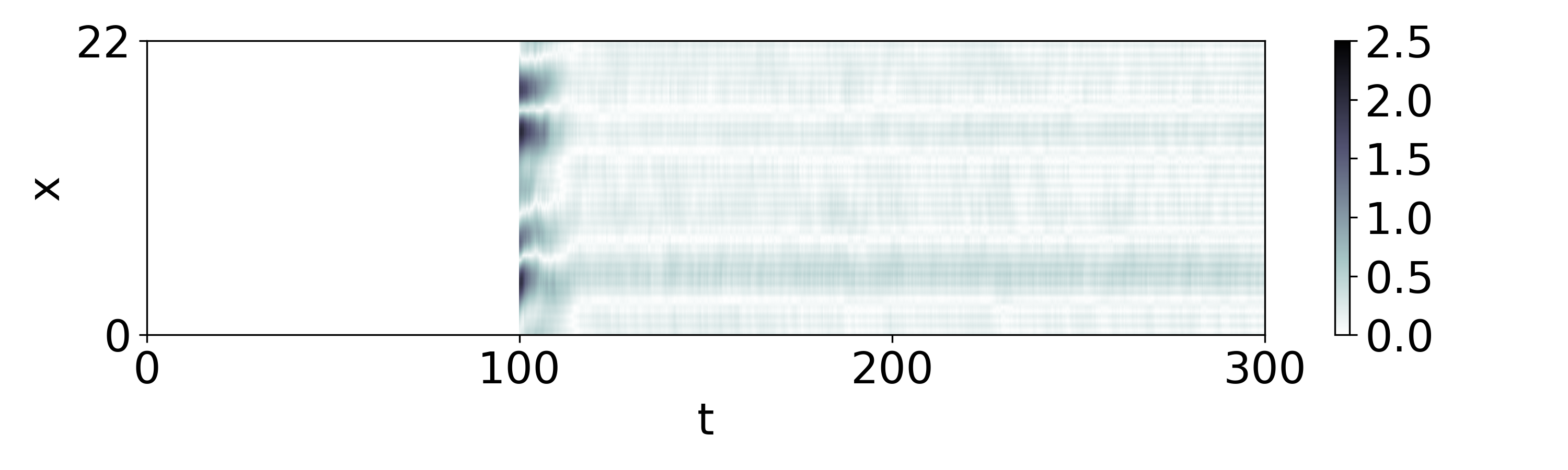}}
    \end{minipage}

    \caption{Controlled solutions for the HypeRL case. Left column: clean setting. Right column: noisy setting. From top to bottom: controlled solution, and absolute state-tracking error.}
    \label{fig:hypeRL_clean_noise}
\end{figure*}

In Figure~\ref{fig:hypEMBER_clean_noise}, \ref{fig:sunrise_clean_noise}, \ref{fig:hypeRL_clean_noise}, we show representative examples of controlled system trajectories obtained using HypEMBER, SUNRISE, and HypeRL. For the sake of a fair comparison, we evaluate the agents in ideal settings (training conditions) and with uncertainties (measurement noise with standard deviation equal to $20\%$ of the signal values and misspecified parameter knowledge with standard deviation equal to $30\%$), starting from the same initial condition and using the same model parameter $\mu = 0.175$. For each agent, from top to bottom we show the evolution of the controlled system state of the KS equation, the sensory measurements, the control actions, and the absolute value of state-tracking error. Compared to SUNRISE and HypeRL, HypEMBER is capable of best mitigating the effect of uncertainties on the control policy.


The results qualitatively highlight the different sensitivity of the different approaches to uncertainties. Moreover, we record in Table \ref{tab:agents_comparison} the following quantitative trends:
\begin{table}[b!]
\centering
\begin{tabular}{l|c|c|c}
\hline
Reward & HypEMBER & SUNRISE & HypeRL \\
\hline
\hline
Ideal settings & $-53.6$ & $-59.4$ & $\mathbf{-52.6}$ \\
\hline
Measurement noise & $\mathbf{-67.1}$ & $-70.8$ & $-68.1$ \\
\hline
Model misspecification & $\mathbf{-61.1}$ & $-73.7$ & $-66.7$ \\
\hline
Combination of the two & $\mathbf{-76.7}$ & $-84.7$ & $-89.1$ \\
\hline
\end{tabular}
\vspace{2pt}
\caption{Comparison of agent performance in ideal and uncertain settings for $\mu = 0.175$. We highlight in \textbf{bold} the best performing agent.}
\label{tab:agents_comparison}
\end{table}
While all the three agents (HypEMBER, SUNRISE, HypeRL) are able to successfully control the system in the ideal scenario with similar cumulative rewards (with HypeRL achieving the best performance), a noticeable performance degradation emerges when uncertainties on measurements and on the parameter $\mu$ come into play. In particular, we observe that the measurement noise introduces a degradation of $13.5, 11.4, 15.5$ absolute reward differences between ideal and perturbed settings for HypEMBER, SUNRISE and HypeRL, respectively (single episode), model misspecification instead introduces a degradation of $7.5 ,14.3, 14.1$, and with a combination of the two the degradation is even more pronounced and equal to $23.1, 25.3, 36.5$. 

\begin{figure*}[h!]
    \centering
    \begin{minipage}{0.49\linewidth}
        \centering
        \includegraphics[width=\linewidth]{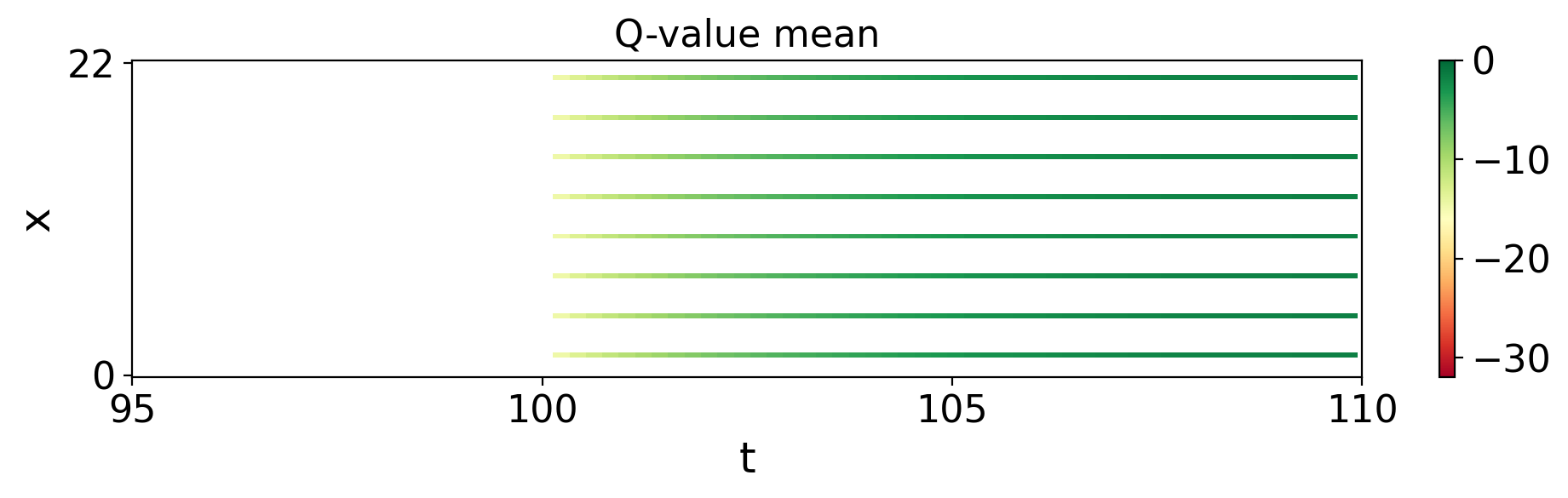}
    \end{minipage}
    \hfill
        \begin{minipage}{0.49\linewidth}
        \centering
        \includegraphics[width=\linewidth]{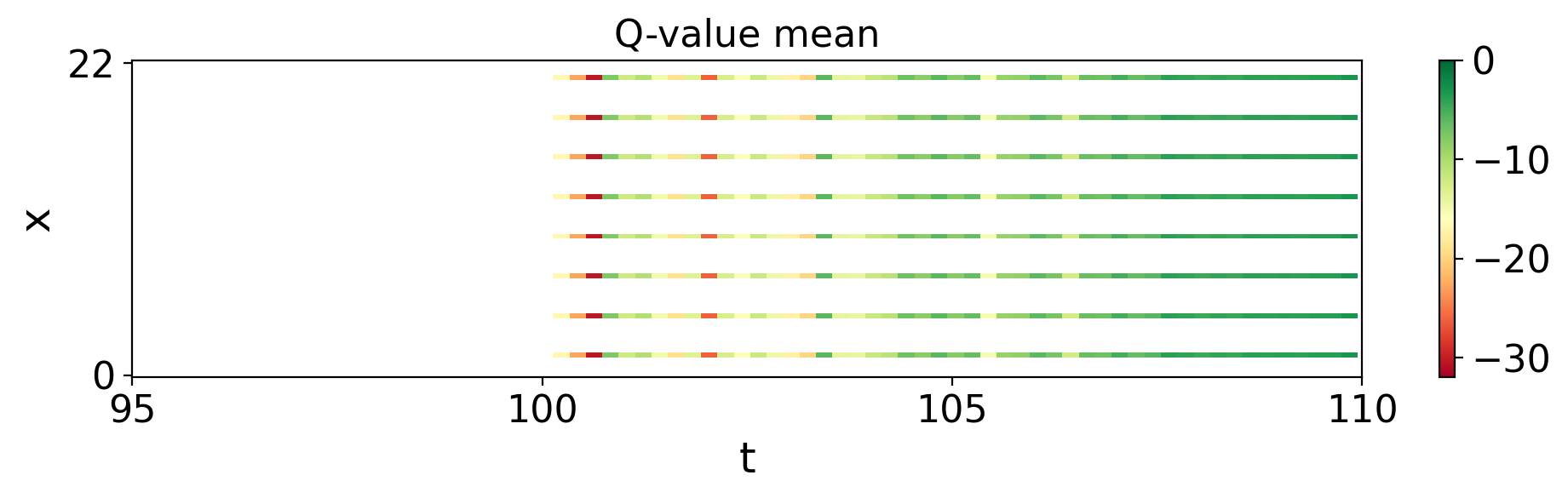}
    \end{minipage}
    \begin{minipage}{0.49\linewidth}
        \centering
        \includegraphics[width=\linewidth]{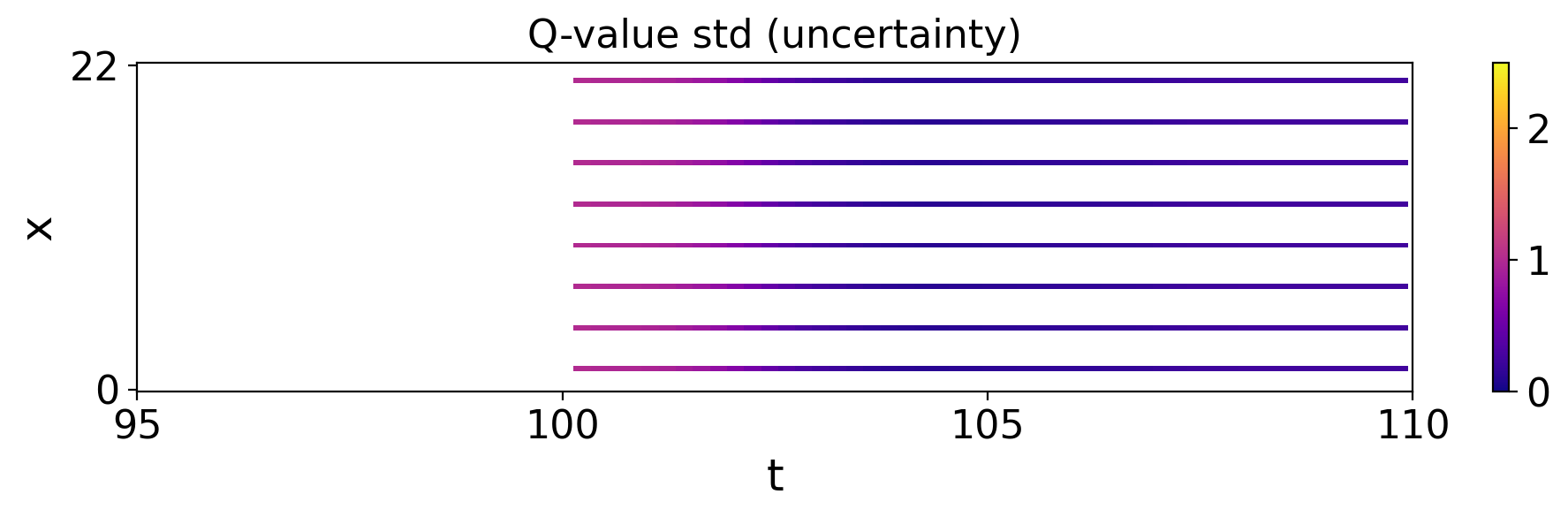}
    \end{minipage}
        \hfill
    \begin{minipage}{0.49\linewidth}
        \centering
        \includegraphics[width=\linewidth]{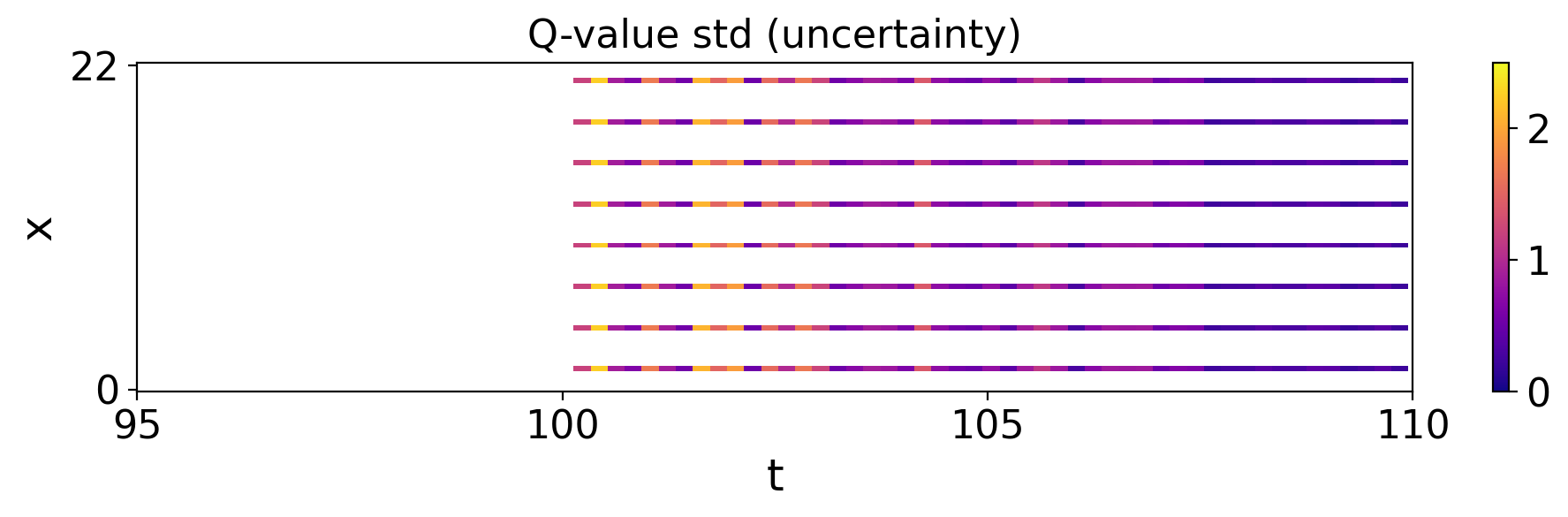}
    \end{minipage}
        \hfill
    \begin{minipage}{0.49\linewidth}
        \centering
        \includegraphics[width=\linewidth]{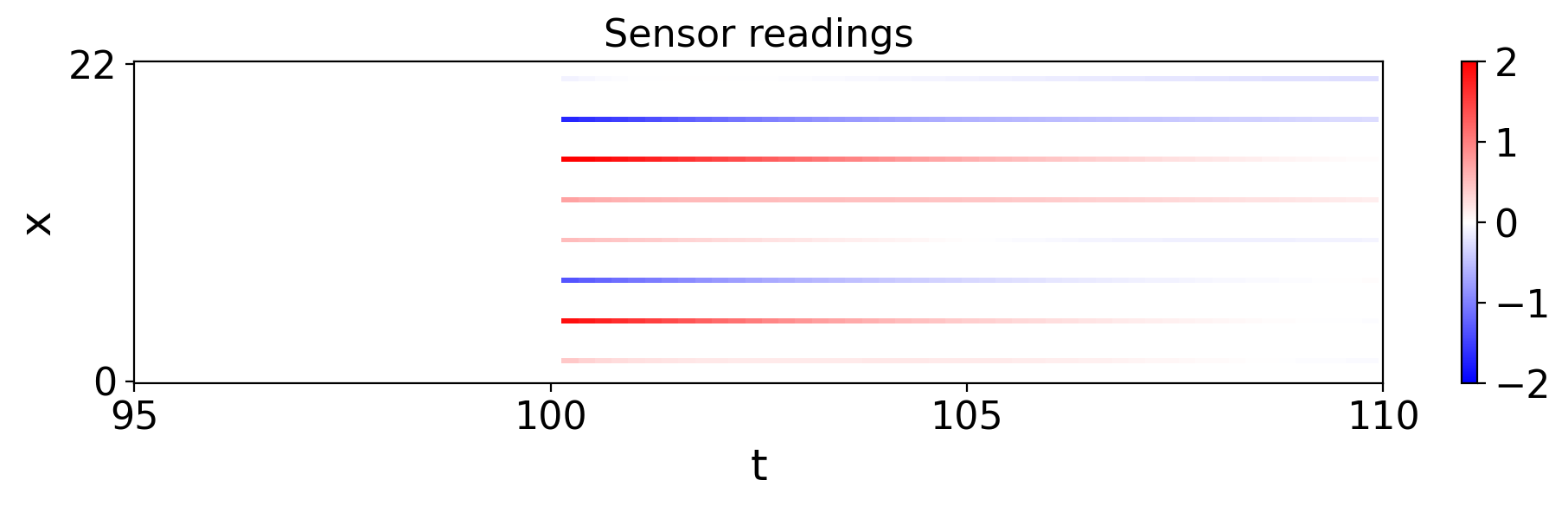}
    \end{minipage}
        \hfill
    \begin{minipage}{0.49\linewidth}
        \centering
        \includegraphics[width=\linewidth]{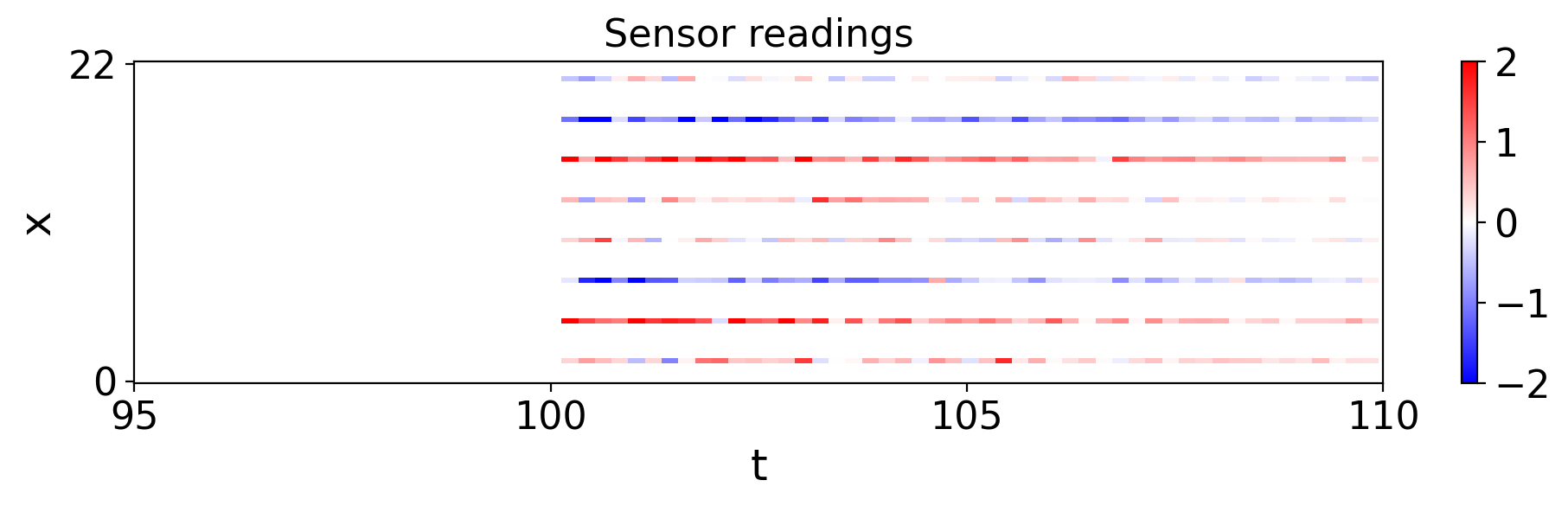}
    \end{minipage}
            \hfill
    \caption{Mean and standard deviation of the Q-values predicted by the ensemble for different levels of measurement noise, namely $0\%, 20\%$.}
    \label{fig:qvalue_ks_comparison}
\end{figure*}

Moreover, in Figure \ref{fig:qvalue_ks_comparison} we show mean and standard deviation of the Q-values predicted by the ensemble of critics of HypEMBER without and with measurement uncertainties. 
As expected, in the presence of uncertainties the predicted mean Q-value is lower, while its standard deviation is higher than in the uncertainty-free settings.

\newpage

\subsection{Particle Navigation in a Double-Gyre Flow}
\label{subsec:results_double_gyre}

The second test case considered in this work is the control of a particle in a time-dependent parametrized double-gyre flow. The controlled dynamics of the particle are defined on the spatial domain $(x,y)$ and are given by
\begin{equation}
\begin{cases}
\frac{dx(t)}{dt} &= \underbrace{-\pi A \sin\!\big(\pi f(x,t)\big)\cos(\pi y)}_{v_x} +  u_x(t), \\
\frac{dy(t)}{dt}&= \underbrace{\phantom{-}\pi A \cos\!\big(\pi f(x,t)\big)\sin(\pi y)\,\frac{\partial f(x,t)}{\partial x}}_{v_y} + u_y(t),
\end{cases}
\begin{aligned}
\end{aligned} \quad (x,y)\in (0,L_x) \times (0,L_y), t \in (0, T).
\label{eq:double_gyre_controlled}
\end{equation}
where $L_x=2$, $L_y=1$, $T=80$, $v_x, v_y$ denote the gyre flow velocities along $x$ and $y$, respectively, and $u(t) = (u_x(t), u_y(t)) \in (-0.4, 0.4)$ denotes the control input. The function $f(x,t)$ is defined as
\begin{equation*}
f(x,t) = a(t)\,x^2 + b(t)\,x,
\qquad
\frac{\partial f(x,t)}{\partial x} = 2a(t)\,x + b(t),
\end{equation*}
with time-dependent coefficients
\begin{equation*}
a(t) = A \sin(\omega t),
\qquad
b(t) = 1 - 2a(t).
\end{equation*}
To solve the state equation in \eqref{eq:double_gyre_controlled}, we use forward Euler with $\Delta t=0.1$, leading to 
\begin{equation*}
\begin{split}
    x_{t+1} &= x_t + \Delta t (v_{x,t}+u_{x,t})\, , \quad x(0) = x_0\, ,\\
    y_{t+1} &= y_t + \Delta t (v_{y,t}+u_{y,t})\, , \quad y(0) = y_0\, ,
\end{split}
\end{equation*}
with $x_0 \sim \mathcal{U}(0.1, 1.9)$ and $y_0 \sim \mathcal{U}(0.1, 0.9)$,randomly sampled from a uniform distribution $\mathcal{U}$ at the beginning of each training episode, and the agent is allowed to control the velocity of the particle $\bm{a}_t = [u_{x,t}, u_{y,t}]$. The parameters $A$ and $\omega$ represent the amplitude and frequency of the flow oscillations, respectively, and are collected in the parameter vector
$\boldsymbol{\mu} = [A, \omega]$. 

Differently from \cite{gunnarson2021learning,krishna2023finite}, where the agent is trained to reach a single target, we consider a more challenging control problem, namely the navigation to arbitrary target locations in a parametric gyre flow. To allow the agent to learn an advanced navigation strategy rather than a single path, we randomly sample the starting point $(x_0, y_0)$, the target location $\bm{s}_{\text{ref}} = (x_{\text{ref}}, y_{\text{ref}})$ with $x_{\text{ref}} \sim \mathcal{U}(0.1, 1.9), y_{\text{ref}} \sim \mathcal{U}(0.1, 0.9)$, and the parameters of the gyre $\boldsymbol{\mu} = (A, \omega)$ for each episode of training and testing. The gyre parameters are uniformly sampled in the intervals $A \sim \mathcal{U}(0.1,0.4)$ and $\omega \sim \mathcal{U}(0.5,2\pi/3)$, with  $A=0.1$.

Biological swimmers usually navigate fluids by sensing the local features of the underlying flow, like velocity, vorticity, or pressure. In particular, the work in~\cite{oteiza2017novel} showed that the lateral line of the zebrafish acts as a vorticity sensor. Following this intuition and leveraging the initial results presented in~\cite{gunnarson2021learning}, we solve the aforementioned problem with a bio-inspired approach. Each agent can observe its current relative position with respect to the target $(\Delta x_t, \Delta y_t) = (x_{\text{ref}}-x_t, y_{\text{ref}}-y_t)$ and the vorticity of the gyre flow field calculated around the agent position $\zeta_t = \zeta(x_t, y_t)$, leading to an observation vector $\bm{s}_t = [\Delta x_t, \Delta y_t, \zeta_t, A, \omega] \in \mathbb{R}^5$. The vorticity is defined as the curl of a given velocity field and in a 2-dimensional setting, its expression boils down to $\zeta_t = \partial v_{y,t}/\partial x_t - \partial v_{x,t}/\partial y_t$.

The control objective consists of steering the particle towards a prescribed target location, while minimizing the control effort as defined in Equation \eqref{eq:reward_function}. Each episode terminates either when the agent successfully reaches the target location up to predefined tolerance, i.e.,
\begin{equation*}
\left(x_t-x_{\mathrm{ref}}\right)^2 + \left(y_t-y_{\mathrm{ref}}\right)^2 < 0.005,
\end{equation*}
or when the maximum allowed episode duration is reached, i.e., $t\geq T$. The experimental setup adopted for this test case mirrors the one previously introduced for the KS equation, where all agents are trained and evaluated for the same number of episodes, number of independent runs, and modalities. 

In Table \ref{tab:training_rew_gyro}, we show the cumulative reward over training collected by the five different agents when trained on 5000 episodes. 
\begin{table}[b!]
\centering
\begin{tabular}{l|c|c|c|c|c}
\hline
Training reward & HypeRL & HypEMBER & SUNRISE & PolyL0-TD3 & TD3 \\
\hline
\hline
Mean $\pm$ Std 
& $-34.2 \pm 1.3$ 
& $\mathbf{-28.6 \pm 5.5}$ 
& $-49.3 \pm 7.4$ 
& $-69.0 \pm 2.2$ 
& $-76.5 \pm 5.5$ \\
\hline
\end{tabular}
\vspace{2pt}
\caption{Training rewards collected by the different agents. We highlight in \textbf{bold} the best performing agent.}
\label{tab:training_rew_gyro}
\end{table}
In this case, HypEMBER is able to achieve the highest training rewards, followed by HypeRL, SUNRISE, PolyL0-TD3, and TD3, which again achieves the worst performance.

\begin{figure}[t!]
    \centering
    \begin{minipage}{0.49\textwidth}
        \centering
        \includegraphics[width=\textwidth]{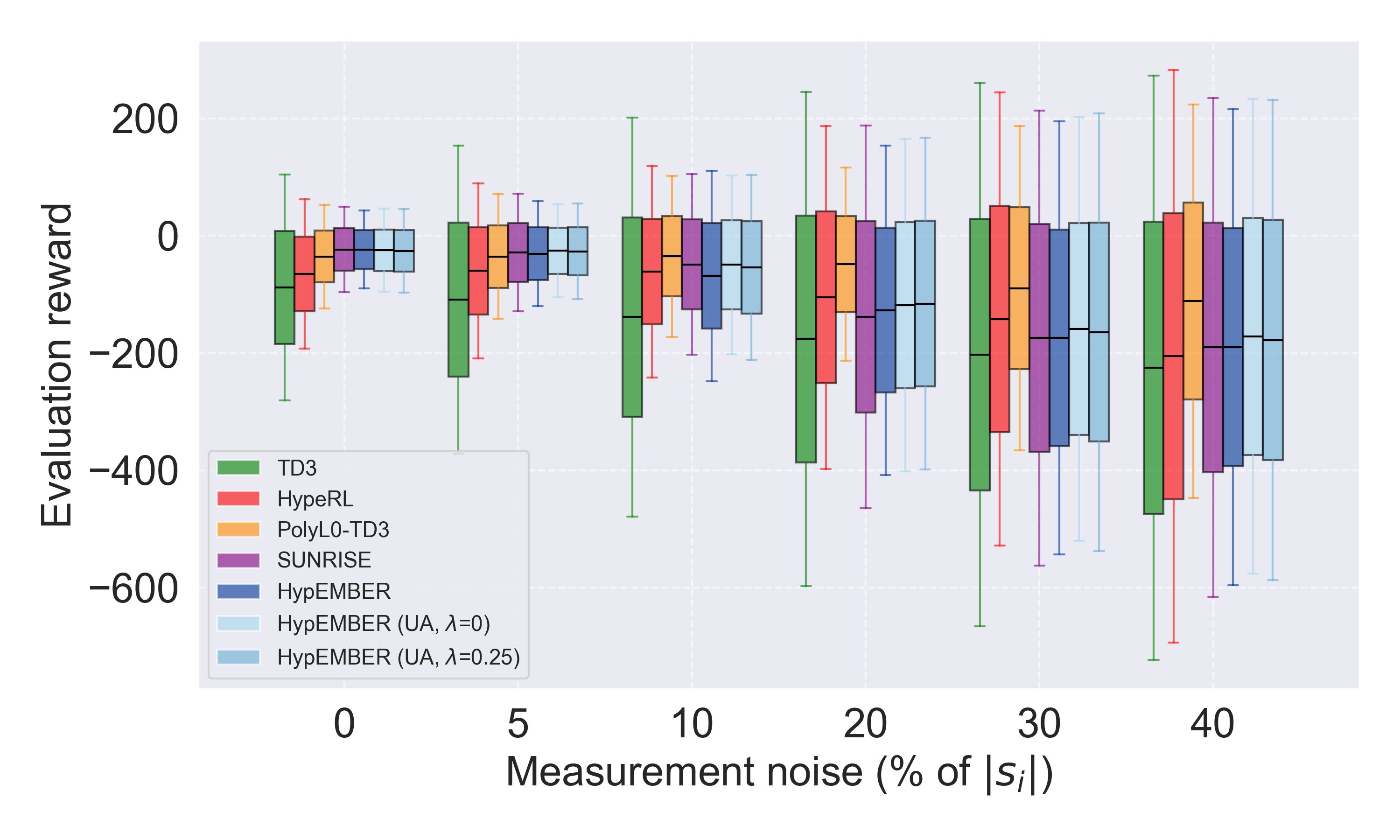}
        \caption{Performance of the different agents when evaluated with measurement noise.}
        \label{fig:measurment_noise_gyre}
    \end{minipage}
    \hfill
    \begin{minipage}{0.49\textwidth}
        \centering
        \includegraphics[width=\textwidth]{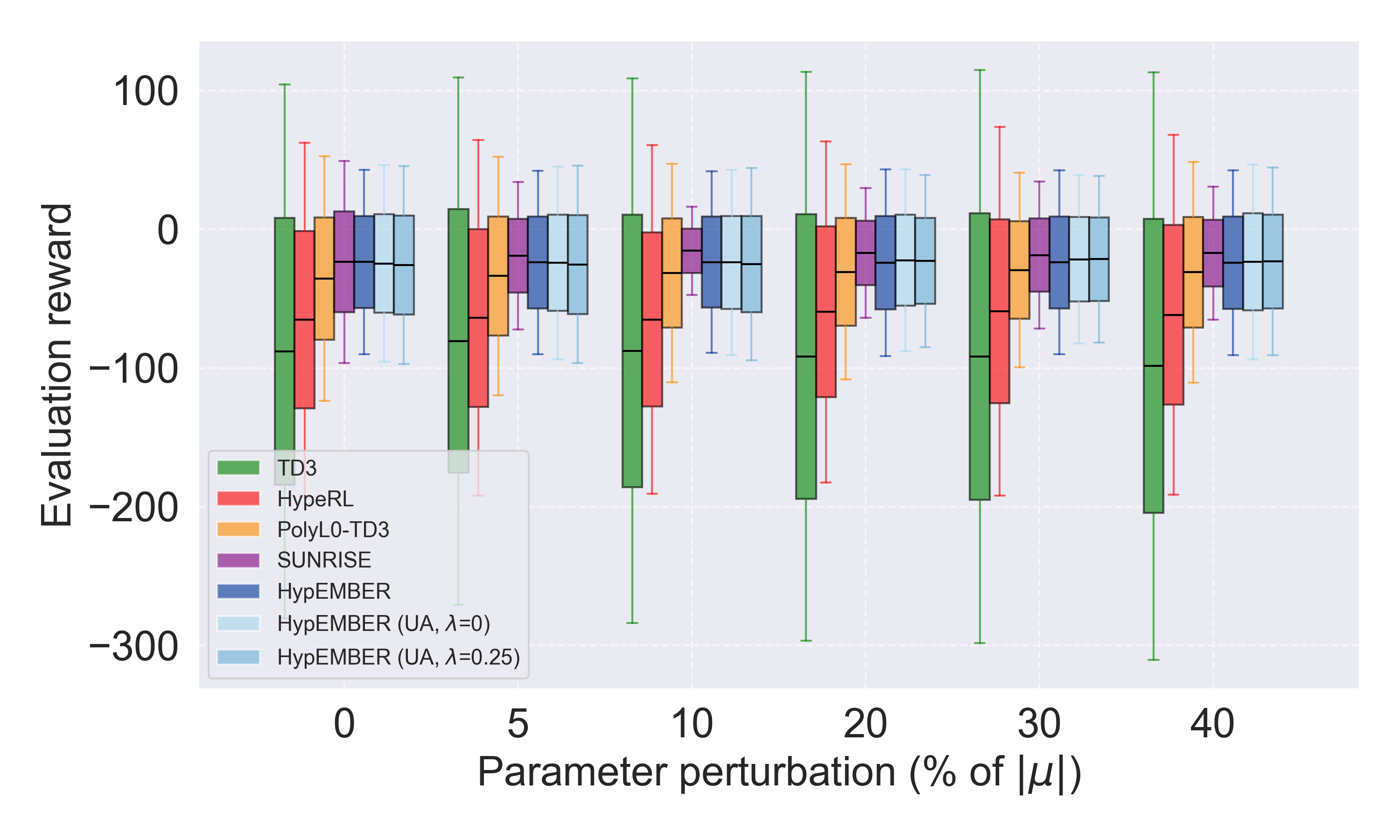}
        \caption{Performance of the different agents when evaluated with perturbation on the physical parameter.}
        \label{fig:misspecification_gyre}
    \end{minipage}
        \hfill
    \begin{minipage}{0.49\textwidth}
        \centering
        \includegraphics[width=\textwidth]{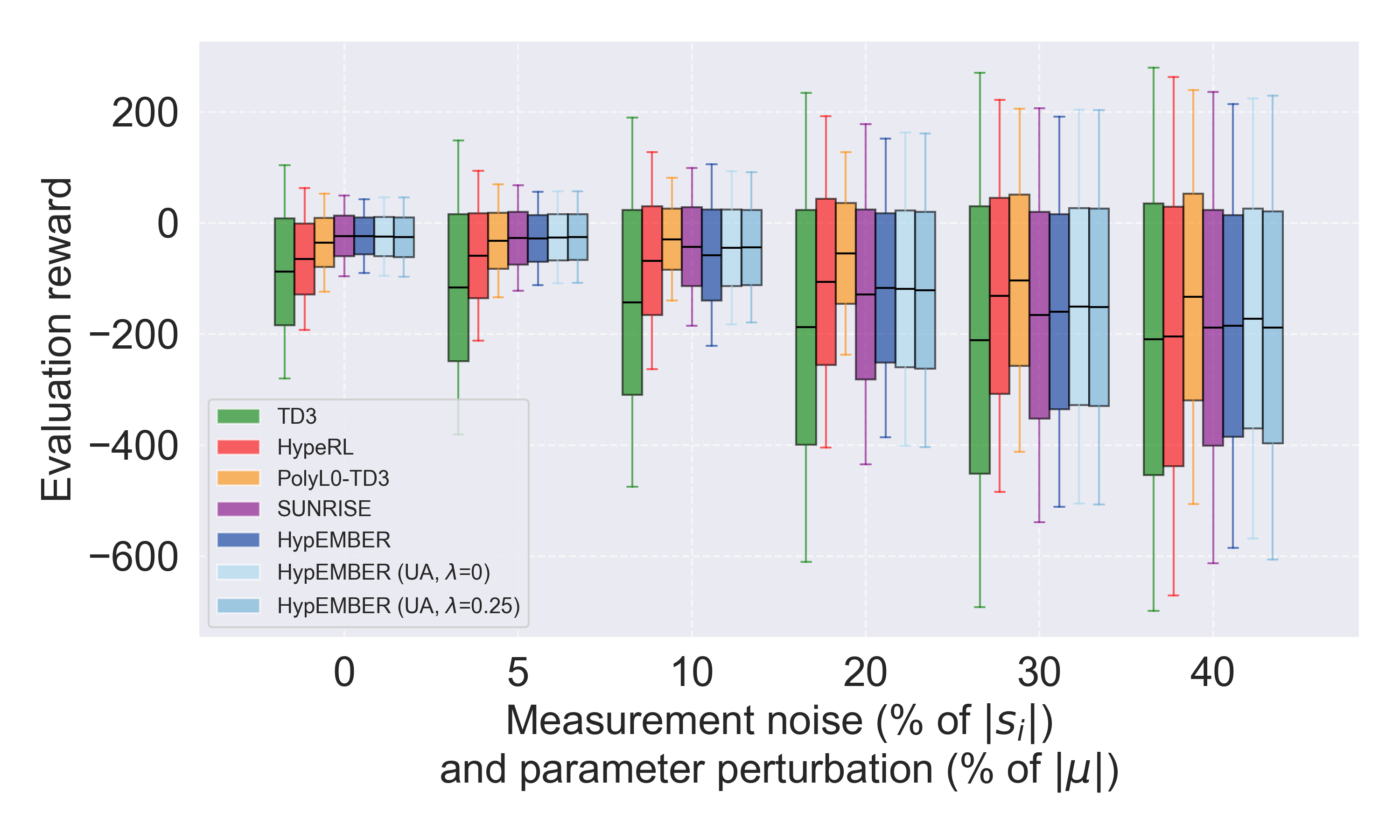}
        \caption{Performance of the different agents when evaluated with measurement noise and perturbation on the physical parameter.}
        \label{fig:combo_gyre}
    \end{minipage}
\end{figure}

Similarly to the KS case, for each algorithms and seed, we assess the performance of the agents over 20 independent evaluation episodes with a randomly sampled $\boldsymbol{\mu}$, initial position $(x_0, y_0)$, and target position $(x_{\text{ref}},y_{\text{ref}})$. The additive noise intensity -- namely $\boldsymbol{\sigma}_M$ and $\boldsymbol{\sigma}_P$ (see Equation \eqref{eq:measurement_noise} and \eqref{eq:model_misspecification}) -- is varied from $0\% $ (ideal testing scenario) to $40\%$ of the signals, allowing a systematic assessment of performance degradation as uncertainties increase. In Figures \ref{fig:measurment_noise_gyre}, \ref{fig:misspecification_gyre}, and \ref{fig:combo_gyre} we report the results (mean and standard deviation of the rewards) obtained from the robustness analysis.
As expected, the agents' performance degrades with the increment of the uncertainties and we can identify a large drop in the performance usually between $10\%-30\%$. In particular:
\begin{itemize}
    \item[-] TD3 is very sensitive to measurement noise and shows a consistent drop of performance from $5\%$ noise onward.
    \item[-] By learning a sparse (polynomial) policy, PolyL0-TD3 is capable of mitigating uncertainties of the sensory measurement up to $20\%$ with a slight decay of performance after.
    \item [-] HypeRL achieves good performance with small amounts of noise, with the results showing a non-negligible drop when the noise goes above $10\%$.
    \item[-] Even in this case, the two ensemble-based approaches, namely SUNRISE and HypEMBER, are capable of best coping with higher levels on noise and best mitigating the degradation due to uncertainties.
    \item[-]  The UA action selection still appears slightly beneficial in improving performance and reducing the variance of the rewards.
\end{itemize}

\begin{figure}[h!]
    \centering
    \begin{minipage}{0.49\textwidth}
        \centering
        \includegraphics[width=\textwidth]{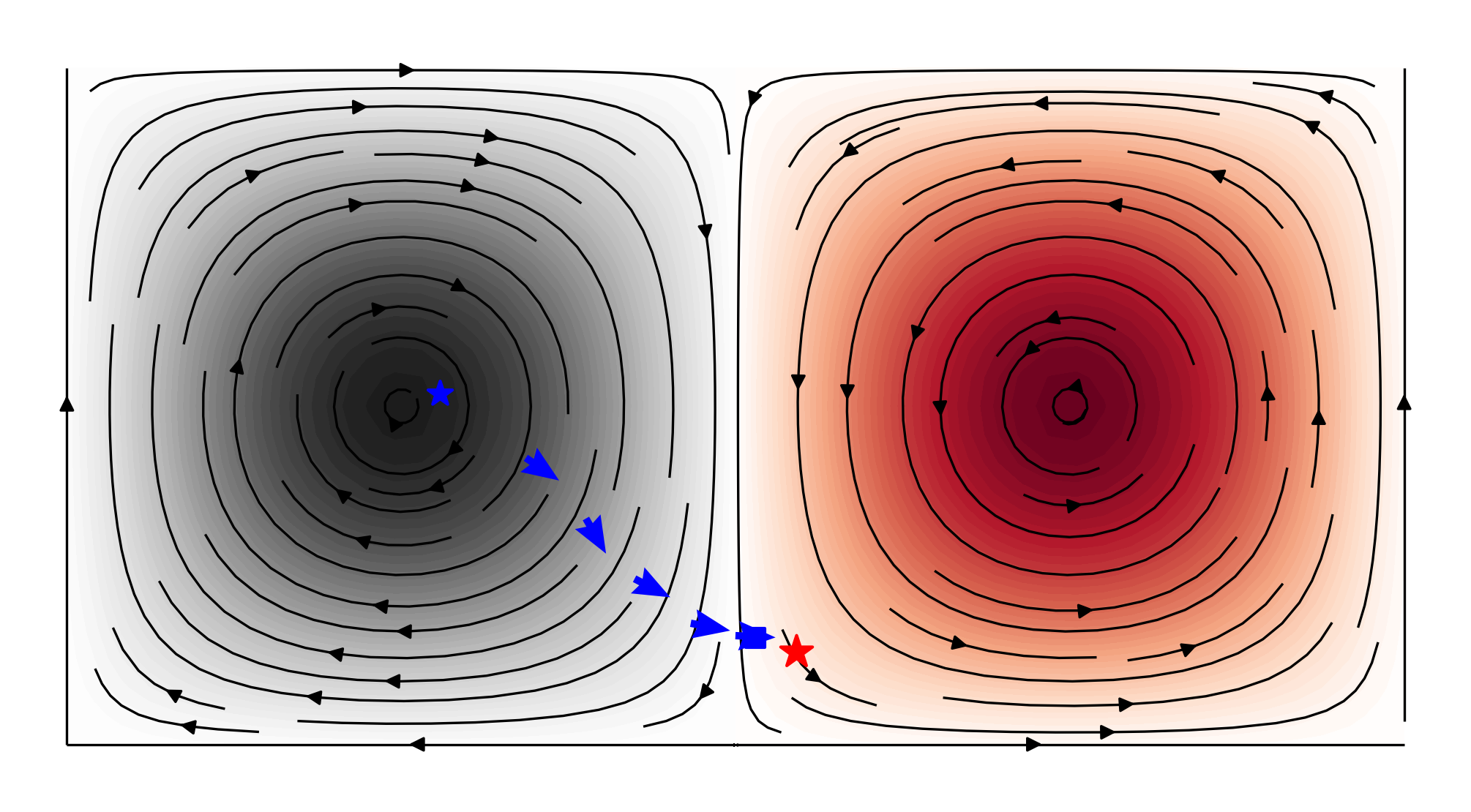}
        \caption*{HypEMBER.}
    \end{minipage}
    \hfill
    \begin{minipage}{0.49\textwidth}
        \centering
        \includegraphics[width=\textwidth]{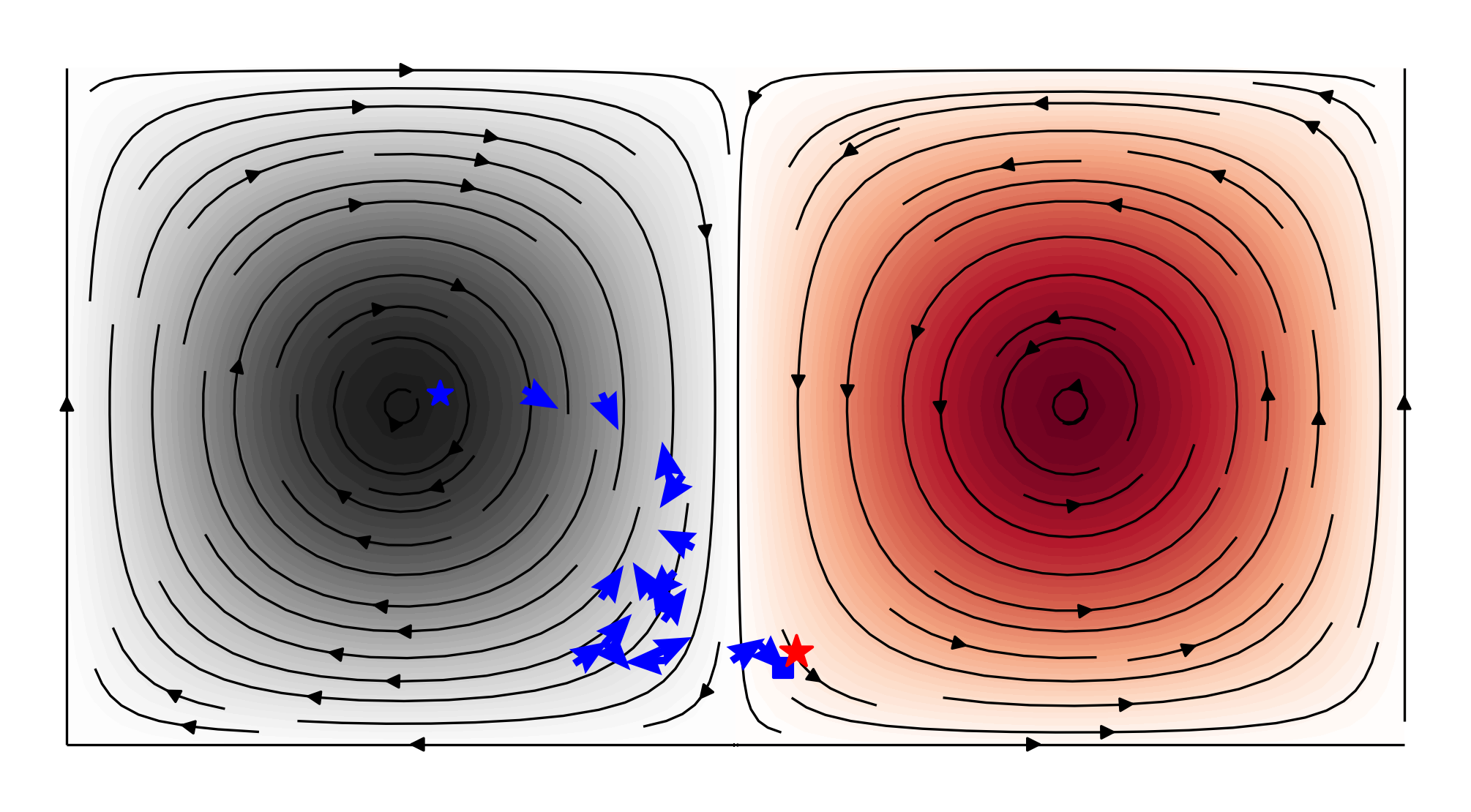}
        \caption*{HypEMBER (uncertainty-aware action selection with \hbox{$\lambda=0.25$}) with measurement noise.}
    \end{minipage}
        \hfill
    \begin{minipage}{0.49\textwidth}
        \centering
        \includegraphics[width=\textwidth]{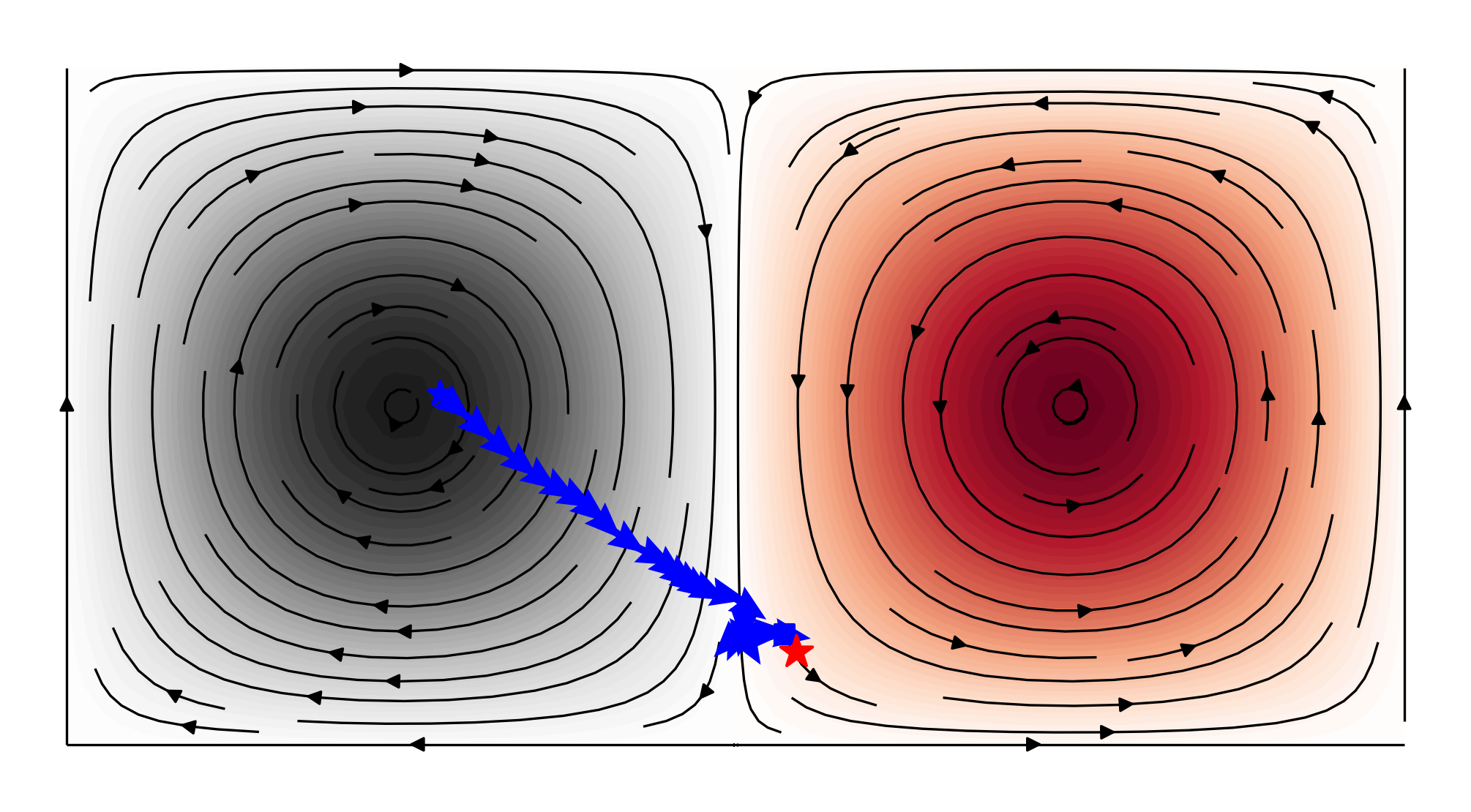}
        \caption*{SUNRISE.}
    \end{minipage}
        \begin{minipage}{0.49\textwidth}
        \centering
        \includegraphics[width=\textwidth]{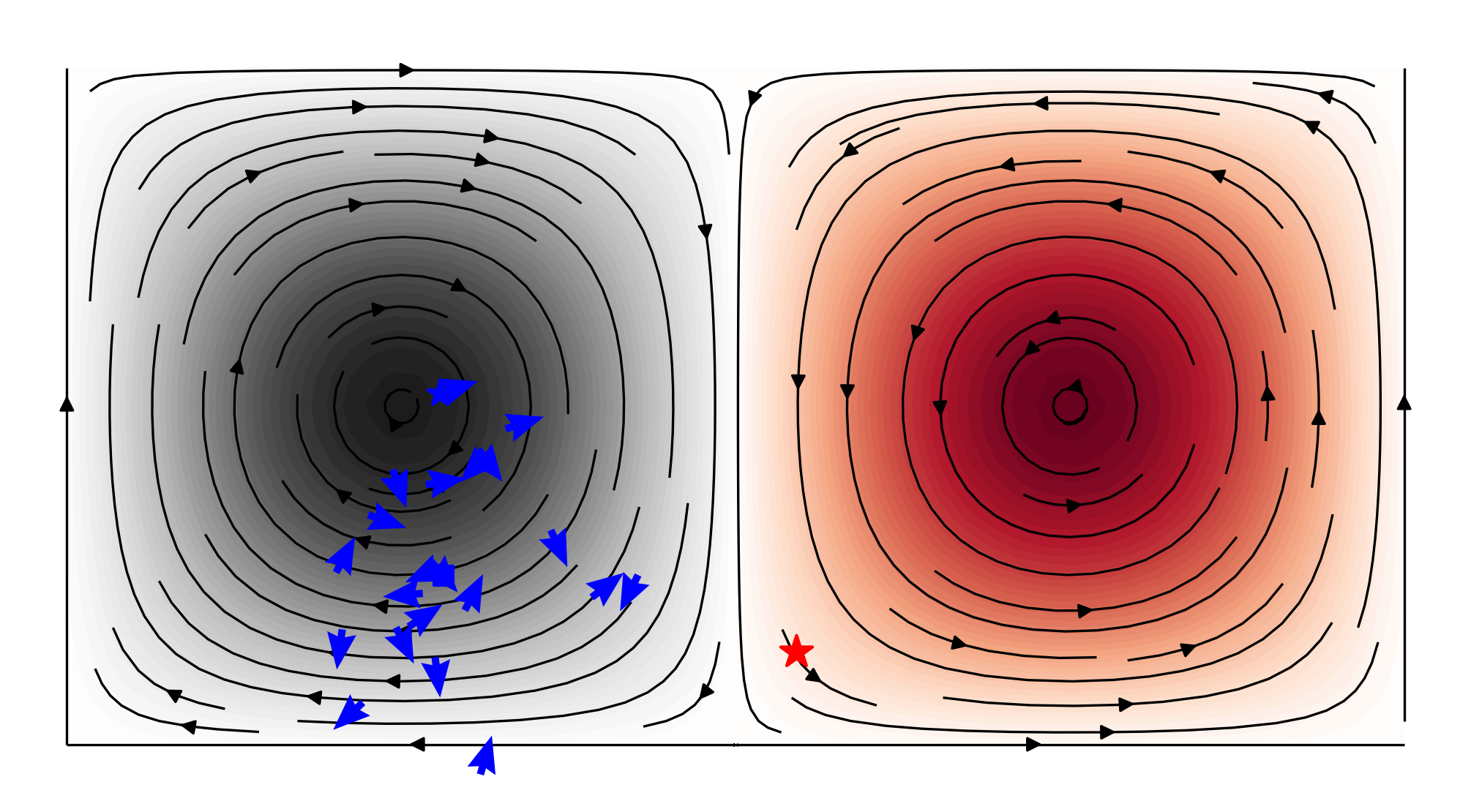}
        \caption*{SUNRISE with measurement noise.}
    \end{minipage}
    \hfill
    \begin{minipage}{0.49\textwidth}
        \centering
        \includegraphics[width=\textwidth]{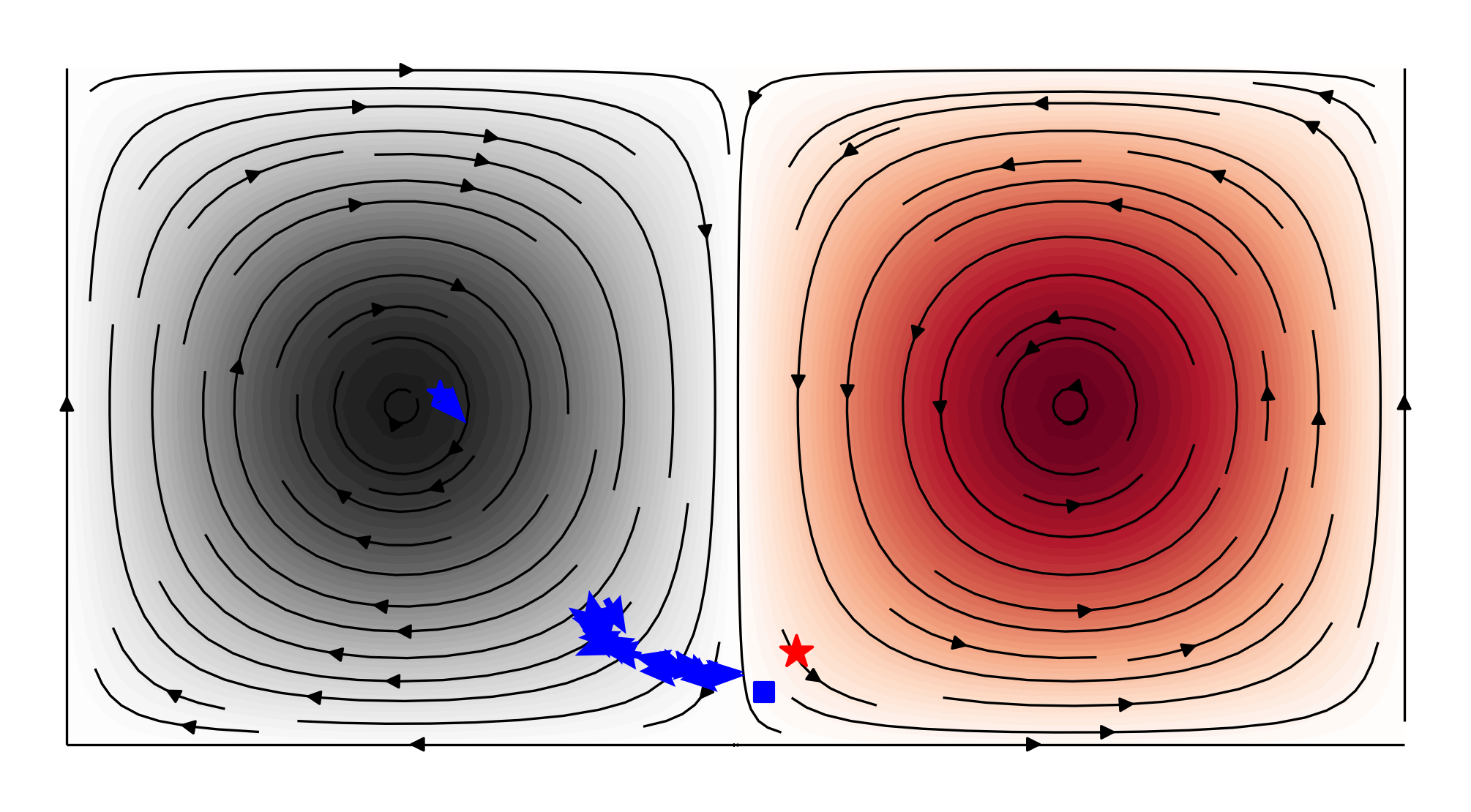}
        \caption*{HypeRL.}
    \end{minipage}
        \hfill
    \begin{minipage}{0.49\textwidth}
        \centering
        \includegraphics[width=\textwidth]{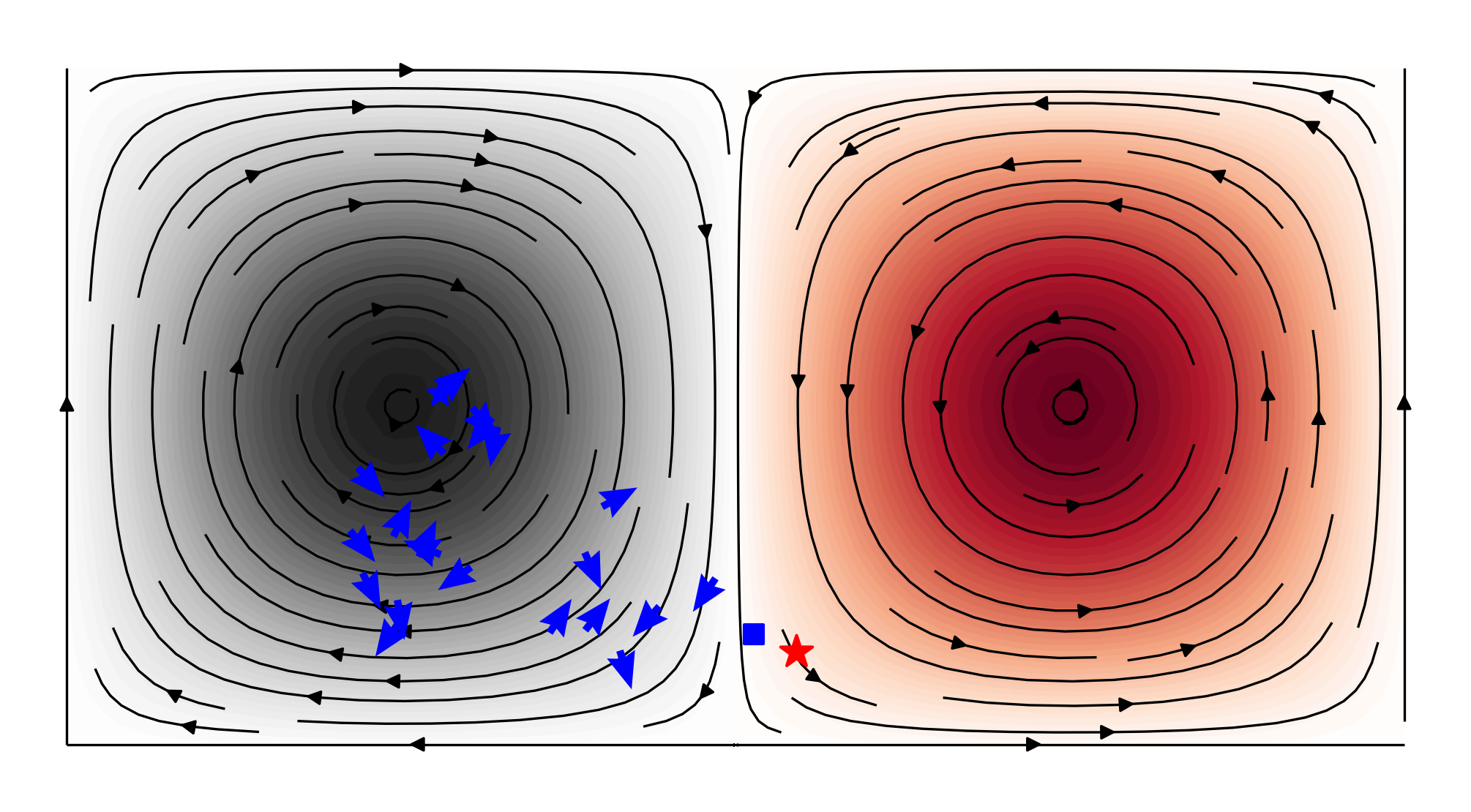}
        \caption*{HypeRL with measurement noise.}
    \end{minipage}
    \caption{Controlled trajectories obtained by  HypEMBER, SUNRISE, and HypeRL in the double gyre flow starting from
the same initial condition and evaluated under identical parameter realizations, in the presence and absence of sensor noise. The blue star denotes the initial position of the particle, the red star the target position, and the blue arrows the velocity of the agent at different timesteps.}
    \label{fig:controlled_solution_gyro}
\end{figure}

In Figure~\ref{fig:controlled_solution_gyro}, we show representative examples of controlled system trajectories obtained using HypEMBER, SUNRISE, and HypeRL. For the sake of a fair comparison, we evaluate the agents in ideal settings (training conditions) and with uncertainties (measurement noise with standard deviation equal to $40\%$ of the signal values and misspecified parameter knowledge with standard deviation equal to $40\%$), starting from the same initial condition $(x_0, y_0)=(0.55, 0.51)$, target position $(x_{\text{ref}},y_{\text{ref}})=(1.09, 0.13)$, and using the same model parameter $A = 0.19$, $\omega=0.85$. 

\begin{table}[h!]
\centering
\begin{tabular}{l|c|c|c}
\hline
Reward & HypEMBER & SUNRISE & HypeRL \\
\hline
\hline
Ideal settings & $-9.75$ & $\mathbf{-8.80}$ & $-44.65$ \\
\hline
Measurement noise & $\mathbf{-315.9}$ & $-629.8$ & $-376.59$ \\
\hline
Model misspecification & $-9.12$ & $\mathbf{-8.31}$ & $-155.2$ \\
\hline
Combination of the two & $\mathbf{-307.8}$ & $-631.40$ & $-368.09$ \\
\hline
\end{tabular}
\vspace{2pt}
\caption{Comparison of agent performance in ideal and uncertain settings for $(x_0, y_0)=(0.55, 0.51)$, $(x_{\text{ref}},y_{\text{ref}})=(1.09, 0.13)$, and model parameter $A = 0.19$, $\omega=0.85$. We highlight in \textbf{bold} the best performing agent.}
\label{tab:agents_comparison_gyro}
\end{table}

\begin{figure}[h!]
    \centering
    \begin{minipage}{0.49\textwidth}
        \centering
        \includegraphics[width=\textwidth]{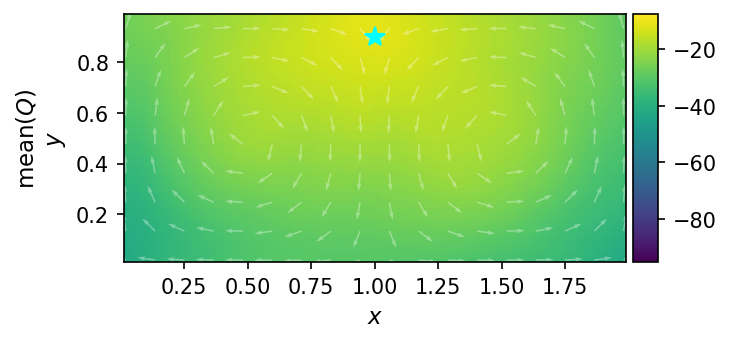}
        \caption*{Q-mean.}
    \end{minipage}
    \hfill
    \begin{minipage}{0.49\textwidth}
        \centering
        \includegraphics[width=\textwidth]{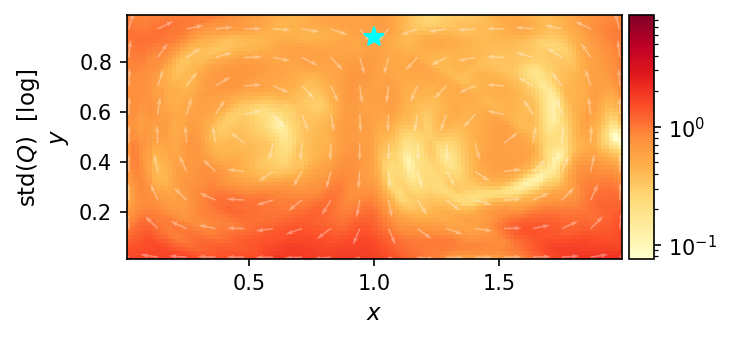}
        \caption*{Q-std.}
    \end{minipage}
        \hfill
    \begin{minipage}{0.49\textwidth}
        \centering
        \includegraphics[width=\textwidth]{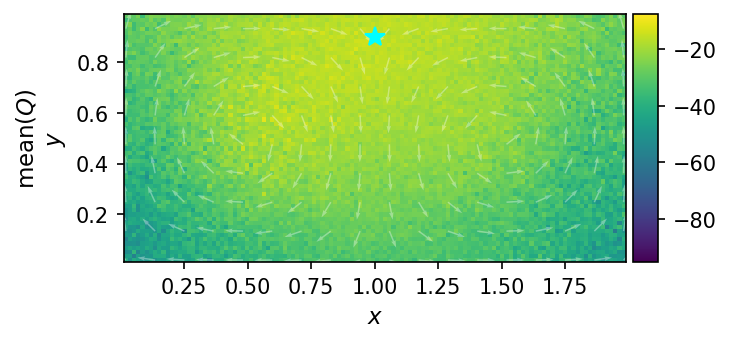}
        \caption*{Q-mean with $20\%$ measurement noise.}
    \end{minipage}
        \begin{minipage}{0.49\textwidth}
        \centering
        \includegraphics[width=\textwidth]{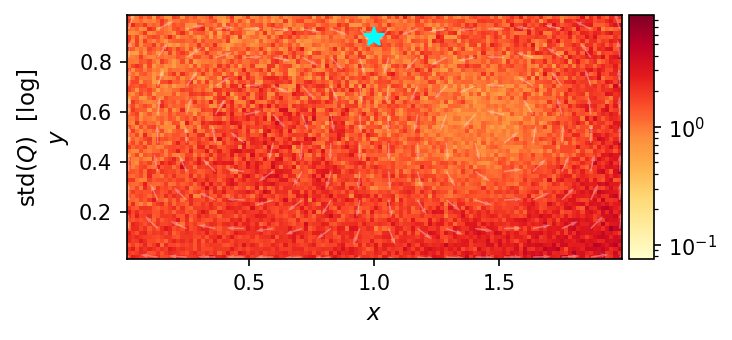}
        \caption*{Q-std with $20\%$ measurement noise.}
    \end{minipage}
    \hfill
    \begin{minipage}{0.49\textwidth}
        \centering
        \includegraphics[width=\textwidth]{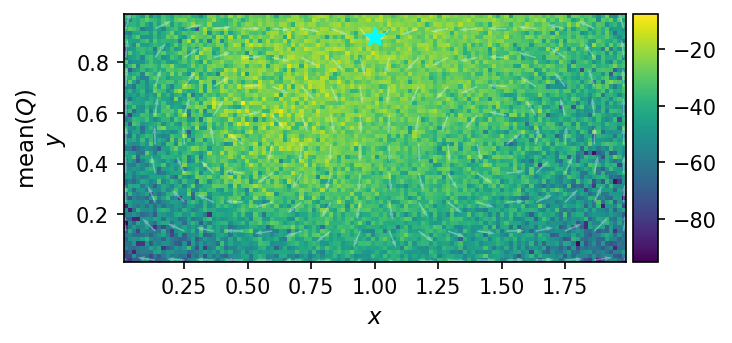}
        \caption*{Q-mean with $40\%$ measurement noise.}
    \end{minipage}
        \hfill
    \begin{minipage}{0.49\textwidth}
        \centering
        \includegraphics[width=\textwidth]{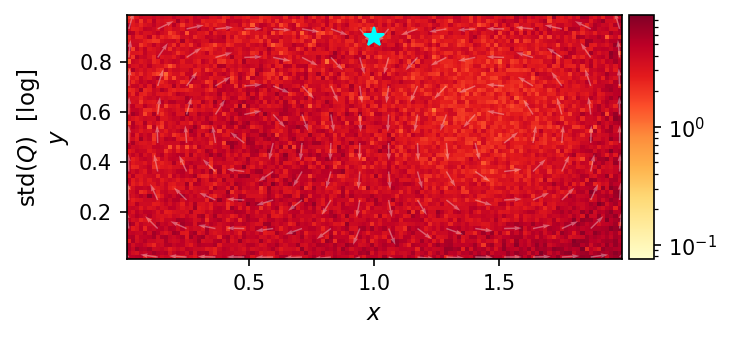}
        \caption*{Q-std with $40\%$ measurement noise.}
    \end{minipage}
    \caption{Mean and standard deviation of the Q-values predicted by the ensemble for different levels of measurement noise, namely $0\%, 20\%, 40\%$.}
    \label{fig:uncertainty_noise_comparison}
\end{figure}

Eventually, in Figure \ref{fig:uncertainty_noise_comparison} we show mean and standard deviation of the Q-values predicted by the ensemble of critics of HypEMBER without and with measurement uncertainties over the domain. 
As expected, in the presence of uncertainties the predicted mean Q-value is lower, while its standard deviation is higher than in the uncertainty-free settings. In addition, it is worth highlighting that even in noise-free settings, the ensemble is capable of identifying the region of higher uncertainties in the flow, namely the one characterized by higher field velocity and, therefore, the ones leading to highest variance of the trajectory if the wrong action is selected.

\section{Conclusion}\label{sec:conclusions}

In the paper, we investigated the application of RL to the control of parametrized dynamical systems in the presence of uncertainties. In particular, we focused on analyzing the robustness of state-of-the-art algorithms in the presence of measurement uncertainties and model misspecification unseen during the training phase of the agents as to simulate a potential simulation-to-reality gap.

To improve robustness and generalization across different dynamical regimes, we introduced HypEMBER, an ensemble-based actor-critic framework in which both policy and value function parameters are dynamically generated through hypernetworks conditioned on physical system parameters. HypEMBER combines uncertainty estimation from critic ensembles with parameter-dependent policy representations, allowing the agent to account for variability in both the system dynamics and the available measurements. In addition, we proposed an uncertainty-aware action selection strategy exploiting the standard deviation of the ensemble of critics to maximize reward and, at the same time, reduce uncertainties.

The experimental results obtained on two benchmark control problems, namely a parametrized Kuramoto-Sivashinsky equation and a particle-navigation task in a parametrized double-gyre flow, highlight the importance of uncertainty-aware action selection and parameter-conditioned policy representations. In particular, measurement uncertainties, model misspecification, in the form of variation of the system parameters from their nominal values, and combinations of the two can be mitigated through our approach effectively.

\section*{Code Availability}

The code used in this research can be found at \href{https://github.com/nicob15/Hypernetwork-Based-Reinforcement-Learning-for-Control-of-Parametrized-Dynamical-}{github.com/nicob15/Hypernetwork-based-Reinforcement-Learning}.

\section*{Acknowledgements}
NB, GP, and AM acknowledge the Project “Reduced Order Modeling and Deep Learning for the real-time approximation of PDEs (DREAM)” (Starting Grant No. FIS00003154), funded by the Italian Science Fund (FIS) - Ministero dell'Università e della Ricerca. AM also acknowledges the project “Dipartimento di Eccellenza” 2023-2027 funded by MUR.

\bibliographystyle{unsrt} 
\bibliography{references}

\appendix

\section{Pseudo-code HypEMBER}\label{sec:pseudocode_HypeRMBER}

\begin{algorithm}
\caption{HypEMBER}\label{alg:hypEMBER}
\begin{algorithmic}
\State Initialize actor hypernetworks
$\{g_{i}(\cdot; \boldsymbol{\phi}_{g_{i}})\}_{i=1}^N$ and actors $\{\pi(\cdot; \boldsymbol{\phi}_{i})\}_{i=1}^N$ \Comment{\texttt{Algorithm initialization}}
\State Initialize critic hypernetworks
$\{h_{i}(\cdot; \boldsymbol{\theta}_{h_{i}})\}_{i=1}^N$ and critics $\{Q(\cdot,  \cdot; \boldsymbol{\theta}_{i})\}_{i=1}^N$
\State Initialize target critic hypernetworks $\{\bar{\boldsymbol{\theta}}_{h_i} \leftarrow \boldsymbol{\theta}_{h_i}\}_{i=1}^N$
\State Initialize entropy coefficients $\{\alpha_i\}_{i=1}^N$
\State Initialize memory buffer $\mathcal{B}$
\State
\For{$e = 1 : E_{\max}$} \Comment{\texttt{Loop over the number of training episodes}}
\State Reset the environment and get initial measurement $\bm{z}_0 = [\bm{s}_0, \boldsymbol{\mu}]$
\For{$t = 0 : N_t$} \Comment{\texttt{Loop over the number of steps in each episode}}
\State Set $\bm{z}_t = [\bm{s}_t,\boldsymbol{\mu}]$
\For{$i = 1,\dots,N$} \Comment{\texttt{Loop over the ensemble size}}
\State Generate policy parameters $\boldsymbol{\phi}_{i} = g_{i}(\bm{z}_t;\boldsymbol{\phi}_{g_i})$
\State Sample action $\bm{a}_{t,i} \sim \pi(\bm{a}_t | \bm{z}_t;\boldsymbol{\phi}_{i})$            
\EndFor
\State  $ \bm{a}^*_t =
        \arg\max_{\bm{a}_{t}}
        \left(
        Q_{\mathrm{mean}}(\bm{z}_t,\bm{a}_{t})
        + \lambda\, Q_{\mathrm{std}}(\bm{z}_t,\bm{a}_{t})
        \right)$
         \Comment{\texttt{UCB action selection}}
\State Apply control $\bm{a}_t=\bm{a}^*_t$, observe reward $r_t$ and next measurement $\bm{s}_{t+1}$
\State Set $\bm{z}_{t+1}=[\bm{s}_{t+1}, \boldsymbol{\mu}]$
\State Store transition $(\bm{z}_t,\bm{a}_t,r_t,\bm{z}_{t+1})$ in $\mathcal{B}$
\EndFor
\If{Update agents}
\For{$k=1,\cdots,K$} \Comment{\texttt{Loop over the number of updates}} 
        \State Sample minibatches
        $(\bm{z}_t,\bm{a}_t,r_t,\bm{z}_{t+1}) \sim \mathcal{B}$
        \For{$i = 1,\dots,N$} \Comment{\texttt{Loop over the ensemble size}}
            \State Generate target policy parameters $\boldsymbol{\phi}_{i} = g_{i}(\bm{z}_{t+1};\boldsymbol{\phi}_{g_{i}})$ \Comment{\texttt{Critic update}}
            \State Sample next action
            $\bm{a}_{t+1} \sim {\pi}(\bm{a}_{t+1} | \bm{z}_{t+1};\boldsymbol{\phi}_i)$
            \State Compute i$^{\text{th}}$-critic target:
            \[
            y_t=
            r_t + \gamma
            \left(
            \bar{Q}(\bm{z}_{t+1},\bm{a}_{t+1};\bar{\boldsymbol{\theta}}_{h_{i}})
            - \alpha_i
            \log {\pi}(\bm{a}_{t+1} | \bm{z}_{t+1};\boldsymbol{\phi}_{i})
            \right)
            \]
            \State Update i$^{\text{th}}$-critic parameters by minimizing: 
            \[ \mathcal{L}_{wQ}(\boldsymbol{\theta}_{h_i}) 
= \mathbb{E}\Big[
w(\bm{z}_{t+1}, \bm{a}_{t+1}) \
\texttt{Huber}\left(
Q_{\boldsymbol{\theta}_i}(\bm{z}_t, \bm{a}_t)
-
y_t
\right)\Big] \]
\State Generate policy parameters $\boldsymbol{\phi}_{i} = h_{\pi_i}(\bm{z}_k;\boldsymbol{\phi}_{g_{i}})$ \Comment{\texttt{Actor update}}
\State Update i$^{\text{th}}$-policy parameters by minimizing:
\[ \mathcal{L}_{\pi}(\boldsymbol{\phi}_{g_i})
=
\mathbb{E}
\left[
\alpha_i \log \pi_{\boldsymbol{\phi}_i}(\bm{a}_t | \bm{z}_t)
-
Q_{\boldsymbol{\theta}_i}(\bm{z}_t,\bm{a}_t)
\right] \]
\If{Update target networks} \Comment{\texttt{Target update}}
\State Soft-update target critic 
\[ \bar{\boldsymbol{\theta}}_{h_{i}}
\leftarrow
\tau\,\boldsymbol{\theta}_{h_{i}} + (1-\tau)\,\bar{\boldsymbol{\theta}}_{h_{i}}\]
\EndIf
\EndFor
\EndFor
\EndIf
\EndFor
\end{algorithmic}
\end{algorithm}

\section{Interpretable Reinforcement Learning with L0-Sparse Polynomial Policies}\label{subsec:polyl0}
\label{polyl0}
The PolyL0 approach is based on the method introduced in \cite{botteghi2024parametric} and is built upon the TD3 actor-critic framework, in which the value function is approximated through standard deep neural networks while the policy representation is replaced by a sparse, structured mapping constructed from a predefined library of nonlinear basis functions.
 In particular, the control policy is expressed as a polynomial mapping between the observed system state and the control action.

Given the agent state
\begin{equation}
\bm{z}_t = [s_t,\mu],
\end{equation}
where $s_t$ denotes the vector of sensor measurements and $\mu$ represents the parameter vector characterizing the underlying parametrized dynamical system, the state is first lifted into a higher-dimensional feature space through a polynomial dictionary $\Theta(\cdot)$. This transformation generates a set of candidate nonlinear features of the form
\begin{equation}
\tilde{z}_t = \Theta(\bm{z}_t) =
[1,\,s_t,\,s_t^2,\,\dots,\,\mu,\,\mu^2,\dots],
\end{equation}
which includes both measurement-dependent terms and parameter-dependent interactions.

The control policy is then defined as a linear combination of these candidate features, namely
\begin{equation}
a_t = \pi(\tilde{z}_t;\xi) = \Theta(\bm{z}_t)\,\Xi,
\end{equation}
where the learnable coefficients $\Xi$ correspond to the weights of a single-layer neural network acting on the polynomial feature space.

To avoid over-parametrization and promote interpretability, sparsity is explicitly enforced on the coefficient matrix through a differentiable approximation of the $\ell_0$-norm. This is achieved by introducing a binary mask $Z(\Psi)$, whose entries determine whether a given polynomial feature is active or inactive in the policy representation. The resulting sparse policy takes the form
\begin{equation}
a_t = \Theta(\bm{z}_t)\,\Xi \odot Z(\Psi),
\end{equation}
where $\odot$ denotes the element-wise product between the polynomial coefficients and the learned sparsity mask.

The policy parameters are trained jointly with the critic networks within the TD3 actor-critic framework by minimizing a loss function that combines the standard deterministic policy-gradient objective with an additional regularization term,
\begin{equation}
L(\xi,\Psi)=
\mathbb{E}\big[- Q(\bm{z}_t,\pi(\tilde{z}_t))+\kappa L_0(\Psi)\big],
\end{equation}
where the coefficient $\kappa$ controls the trade-off between performance optimization and policy sparsity.

This formulation retains the expressive power of deep reinforcement learning for approximating the value function, while replacing the policy network with a compact polynomial structure containing only a limited number of active terms. As a result, the learned control law admits an explicit analytical representation, enabling interpretability and facilitating robustness or stability analyses that are typically not accessible when using dense neural-network policies.

\section{Hyperparameters of the Experiments}
\label{hyperparameters}

In Table \ref{table:hyperparameters}, we report the hyperparameters used in our numerical experiments.
\begin{table}[h!]

\centering
\caption{Training hyperparameters for the algorithms}
\begin{tabular}{lccccc}
\toprule
Hyperparameter & TD3 & HypeRL & PolyL0-TD3 & SUNRISE & HypEMBER \\
\midrule
Max episodes (KS/gyre) & 1000/5000 & 1000/5000 & 1000/5000 & 1000/5000 & 1000/5000\\
Batch size & 256 & 256 & 256 & 256 & 256\\
Updates per episode & 100 & 100 & 100 & 100 & 100\\
Actor learning rate & $3 \times 10^{-4}$ & $1 \times 10^{-6}$ & $1 \times 10^{-3}$ & $1 \times 10^{-4}$ & $1 \times 10^{-6}$ \\
Critic learning rate & $3 \times 10^{-4}$ & $5 \times 10^{-5}$ & $3 \times 10^{-4}$ & $1 \times 10^{-4}$ & $5 \times 10^{-6}$ \\
Critic loss & MSE & Huber & MSE & MSE & Huber\\
Actor type & deterministic & deterministic & deterministic & stochastic & stochastic \\ 
Actor hidden size & 256 & 256 (dynamic) & -- & 1024 & 256 (dynamic) \\
Critic hidden size & 256 & 256 (dynamic) & 256 & 1024 & 256 (dynamic) \\
Actor  hidden layers & 2 & 1 (dynamic) & -- & 2 & 1 (dynamic) \\
Critic hidden layers & 2 & 1 (dynamic) & 2 & 2 & 1 (dynamic)\\
Polynomial degree & -- & -- & 2 & -- & -- \\
L0 droprate & -- & -- & 0.2 & -- & -- \\
L0 weight decay & -- & -- & $1 \times 10^{-5}$ & -- & --\\
Entropy coefficient LR & -- & -- & -- & $1\times10^{-4}$ & $1\times10^{-4}$ \\
Initial entropy $\alpha$  & -- & -- & -- & 0.1 & 0.1 \\
Weighted Bellman temperature $T$ & -- & -- & --  & 20.0 & 20.0 \\
Target delay & 2 & 2 & 2 & 2 & 2 \\
Target update $\tau$ & 0.005 & 0.005 & 0.005 & 0.005 & 0.005 \\
Num. actors (ensemble) & -- & -- & -- & 5 & 5 \\
Num. Critics (ensemble) & -- & -- & -- & 5 & 5 \\
\bottomrule
\end{tabular}
\label{table:hyperparameters}
\end{table}

\section{Additional Results}\label{app:additional_results_lambda}
\begin{figure}[h!]
    \centering
    \begin{minipage}{0.49\textwidth}
        \centering
        \includegraphics[width=\textwidth]{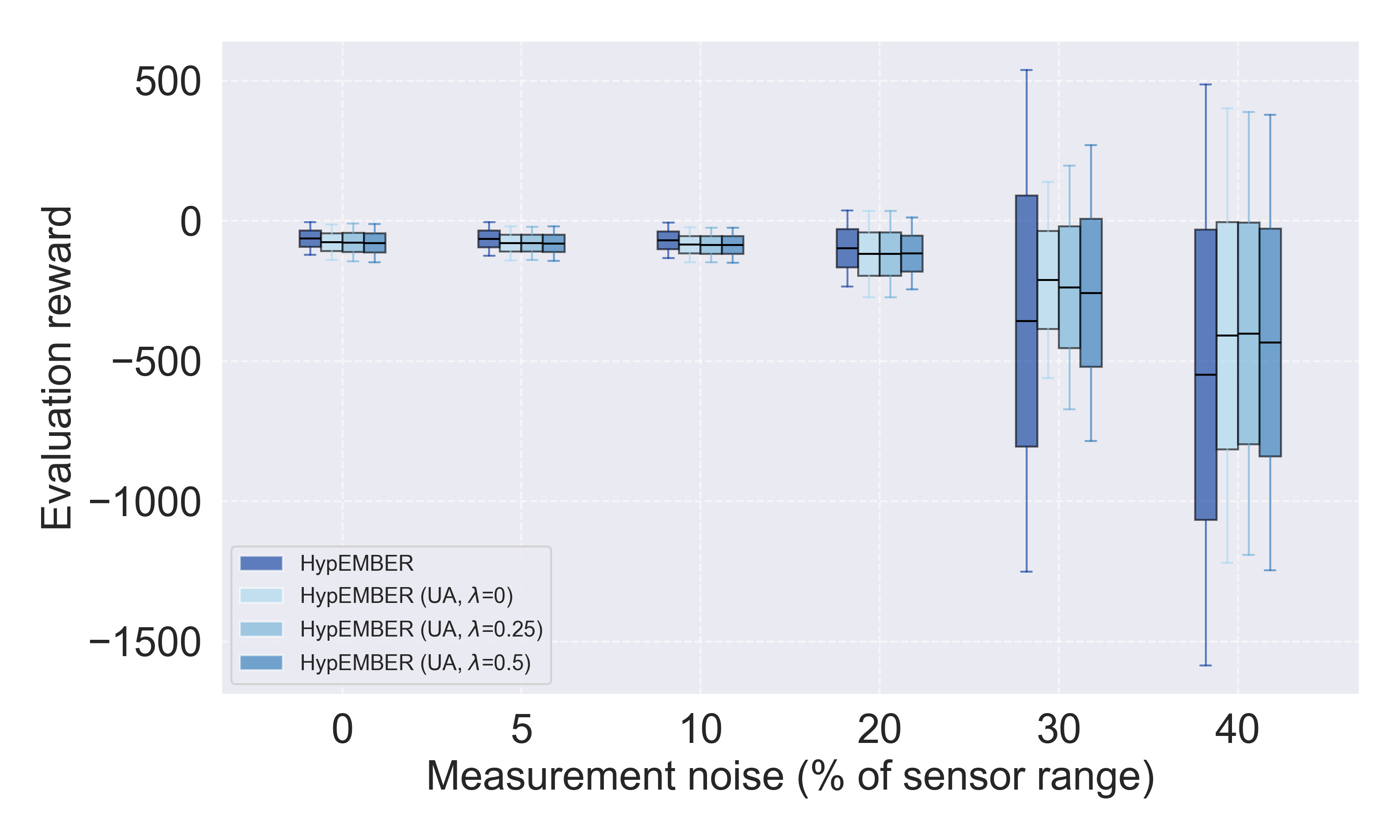}
        \caption{Performance of HypEMBER with different action-selection strategies when evaluated with measurement noise.}
        \label{fig:measurment_noise_ks_app}
    \end{minipage}
    \hfill
    \begin{minipage}{0.49\textwidth}
        \centering
        \includegraphics[width=\textwidth]{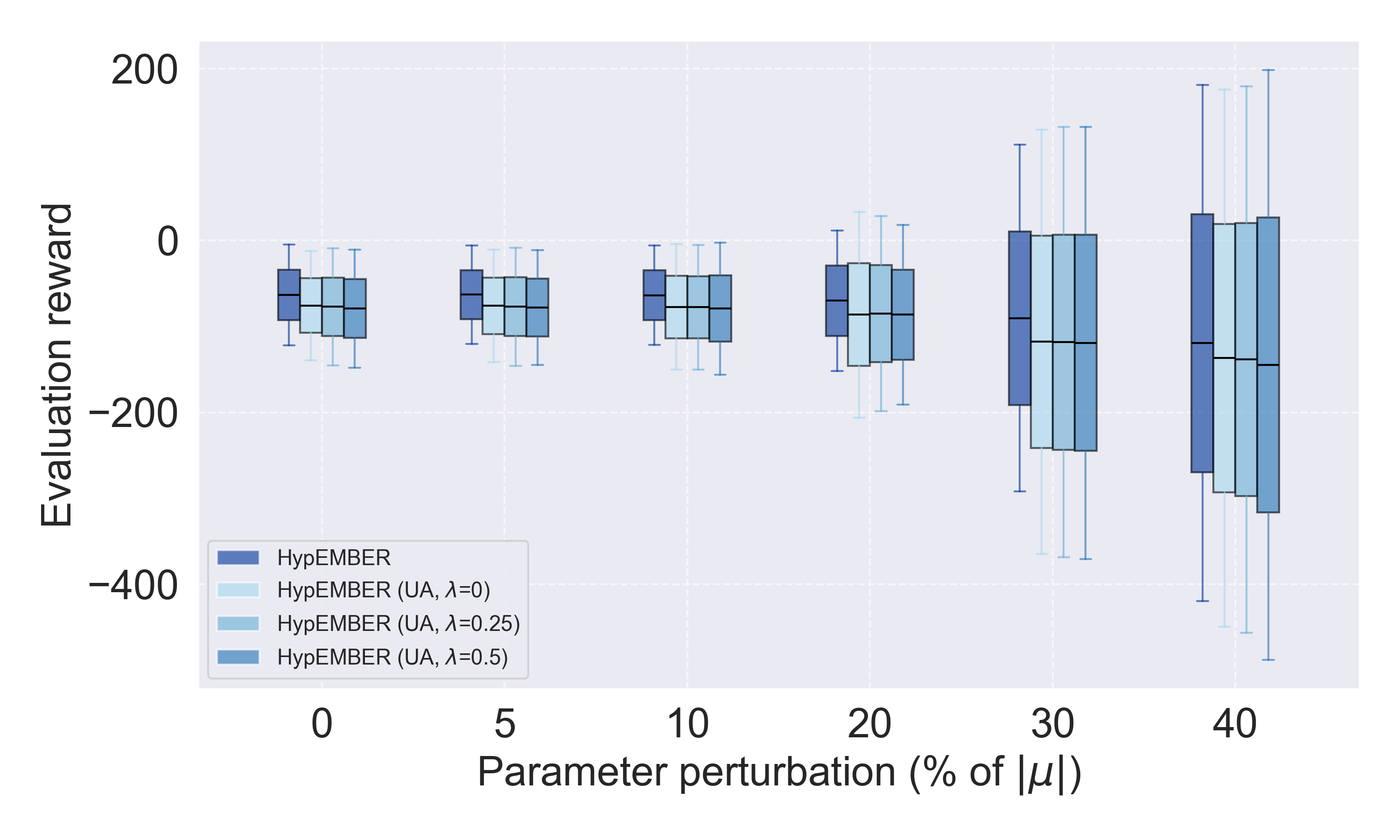}
        \caption{Performance of HypEMBER with different action-selection strategies when evaluated with perturbation on the physical parameter.}
        \label{fig:misspecification_ks_app}
    \end{minipage}
        \hfill
    \begin{minipage}{0.49\textwidth}
        \centering
        \includegraphics[width=\textwidth]{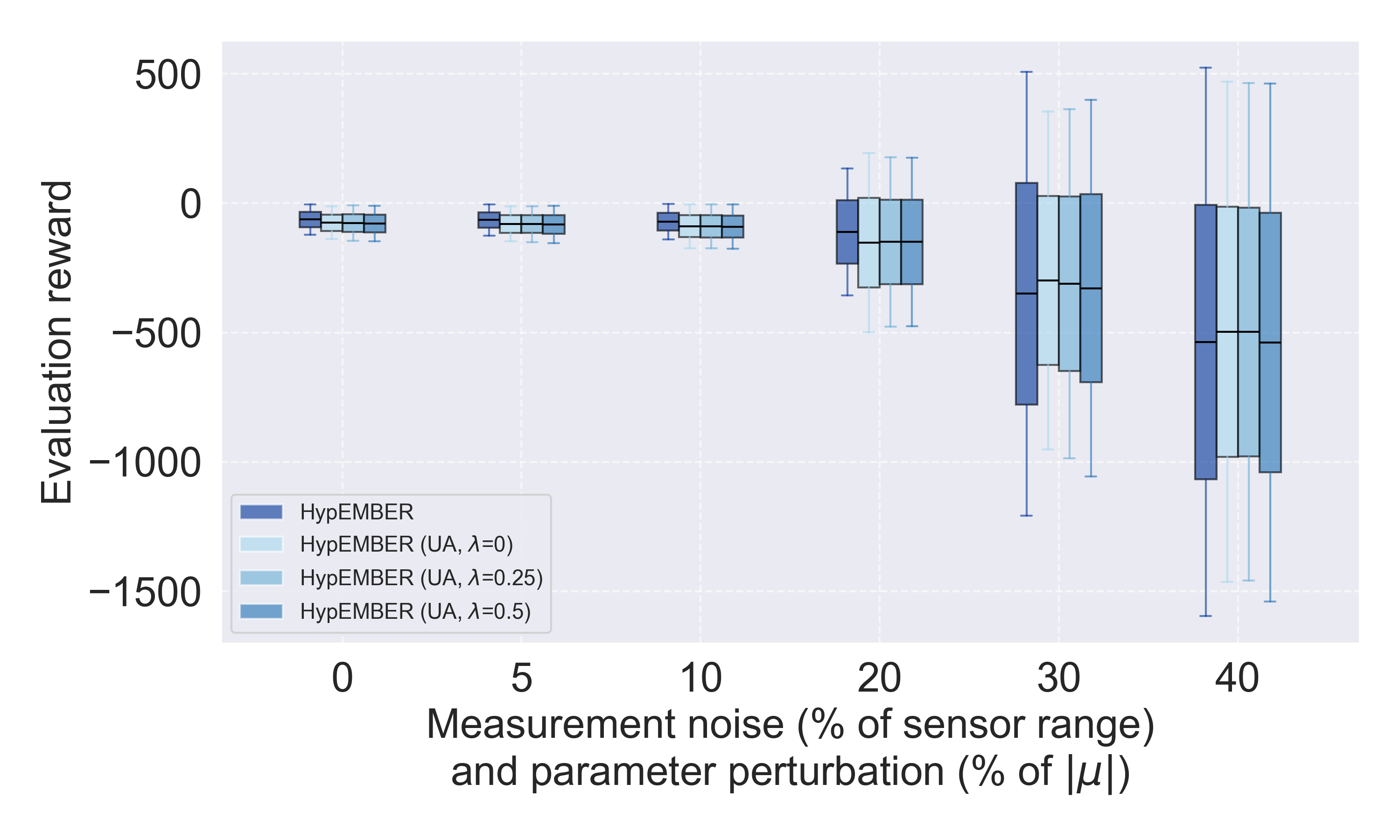}
        \caption{Performance of HypEMBER with different action-selection strategies when evaluated with measurement noise and perturbation on the physical parameter.}
        \label{fig:combo_ks_app}
    \end{minipage}
\end{figure}

\begin{figure}[h!]
    \centering
    \begin{minipage}{0.49\textwidth}
        \centering
        \includegraphics[width=\textwidth]{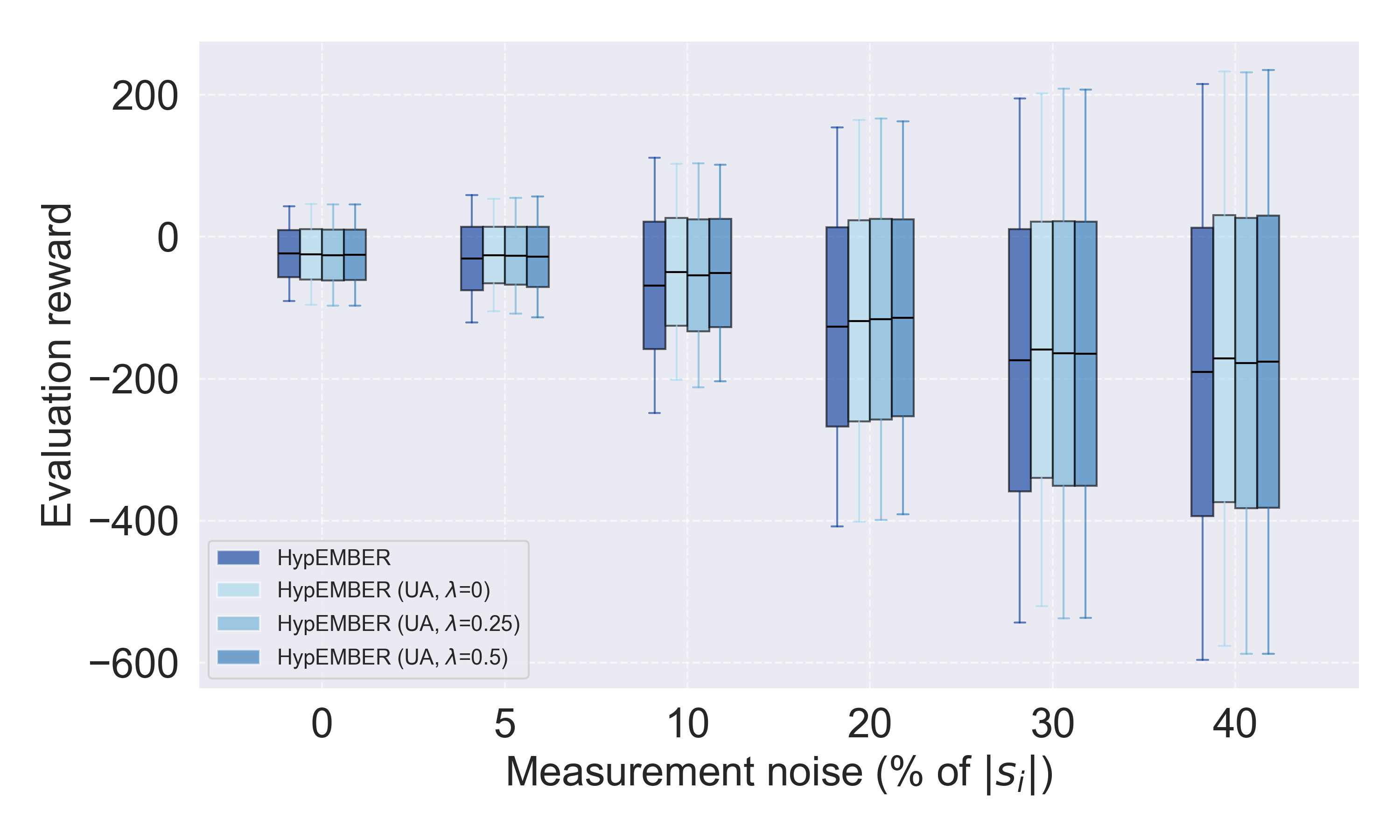}
        \caption{Performance of HypEMBER with different action-selection strategies when evaluated with measurement noise.}
        \label{fig:measurment_noise_gyro_app}
    \end{minipage}
    \hfill
    \begin{minipage}{0.49\textwidth}
        \centering
        \includegraphics[width=\textwidth]{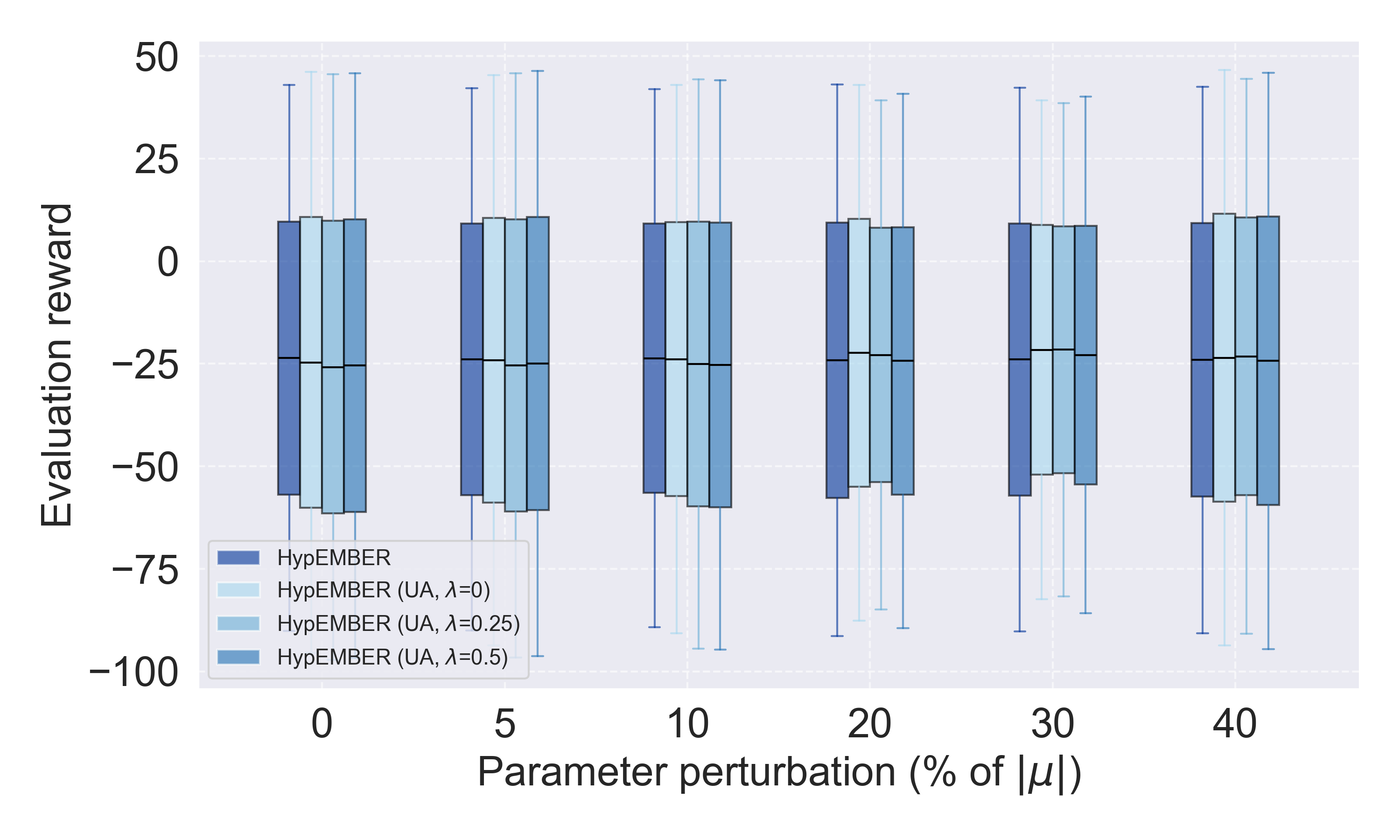}
        \caption{Performance of HypEMBER with different action-selection strategies when evaluated with perturbation on the physical parameter.}
        \label{fig:misspecification_gyro_app}
    \end{minipage}
        \hfill
    \begin{minipage}{0.49\textwidth}
        \centering
        \includegraphics[width=\textwidth]{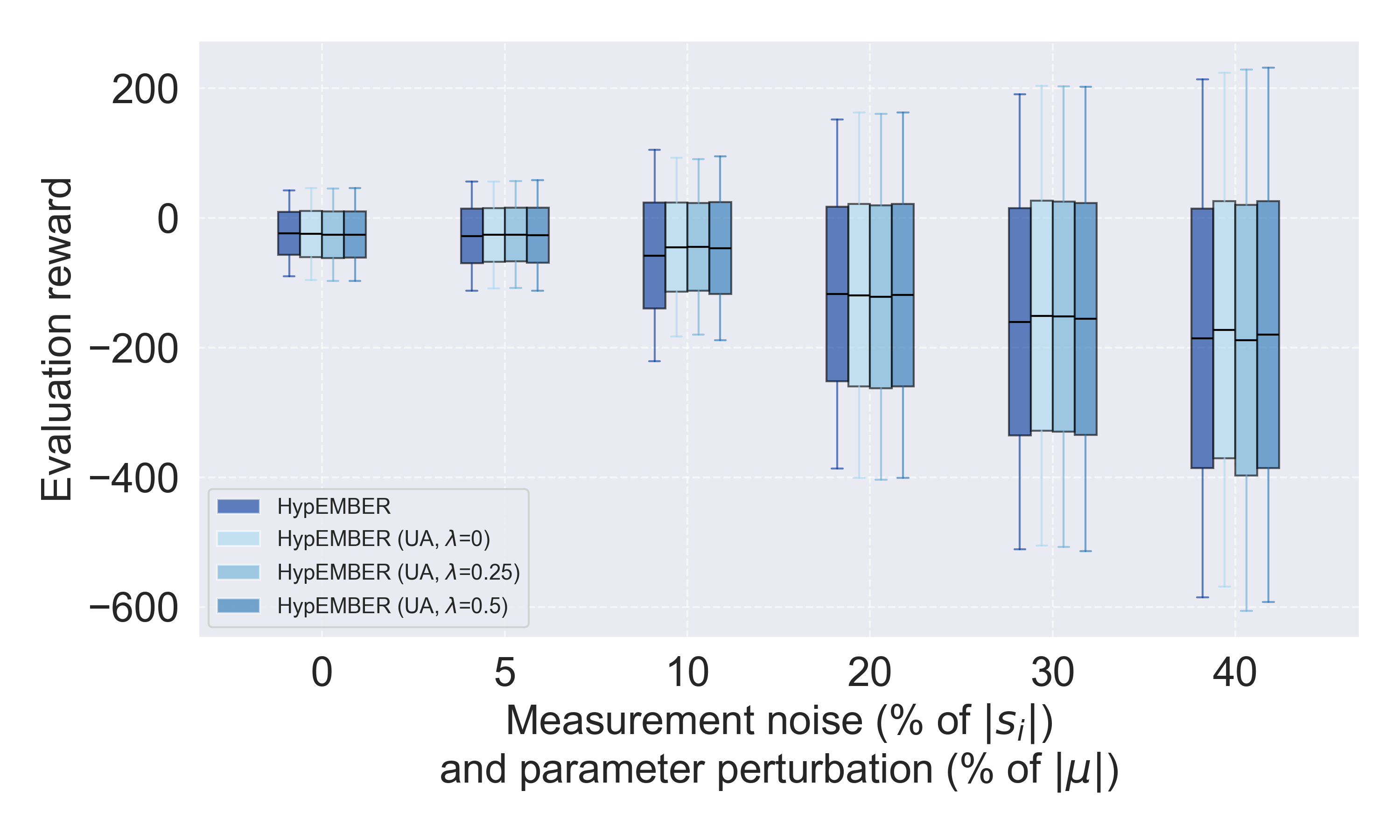}
        \caption{Performance of HypEMBER with different action-selection strategies when evaluated with measurement noise and perturbation on the physical parameter.}
        \label{fig:combo_gyro_app}
    \end{minipage}
\end{figure}

In Figures \ref{fig:measurment_noise_ks_app}-\ref{fig:combo_ks_app}, we show the effect of the different action-selection strategies when controlling the KS equation, while in Figures \ref{fig:measurment_noise_gyro_app}-\ref{fig:combo_gyro_app} the same experiment is repeated for the double-gyre flow problem. Especially for high uncertainties, $\lambda=0.25$ seems to improve the rewards, while reducing their variance.

\end{document}